\documentclass{comnet}%%%%where comnet is the template name
\usepackage{amsmath}
\usepackage{natbib}
\usepackage{xcolor}
\usepackage{graphicx}
\usepackage{xurl}
\usepackage{booktabs}
\usepackage{array}
\usepackage{float}
\usepackage{tcolorbox}
\usepackage{enumitem}
\usepackage{amssymb}   % for \Box, \checkmark
\usepackage{placeins}
\usepackage{tabularx}
\usepackage[hidelinks]{hyperref}

\begin{document}
\raggedbottom

\title{A Generative Model for Higher-Order Network Dynamics}

%A Generative Model and Validation Framework for Higher-Order Network Dynamics

%When Are Higher-Order Representations Necessary? A Generative Benchmark for Network Dynamics

%A generative model for benchmarking higher-order interaction detection in dynamical networks

%a generative validation framework for higher-order interaction detection in network dynamics

%a  generative framework for evaluating higher-order representations on network dynamics.

%HOND: Higher-Order Neural Dynamics

%Modelling higher-order interactions in network dynamical systems with applications to neural population dynamics
%HO-DCM: Modelling Higher-Order Interactions in Neural Dynamics

%A generative model for higuer-order interactions in network dynamics

%HO-DCM: Modelling Higher-Order Interactions in Network Dynamics

%A generative model for higher-order interactions in network dynamics

% A Benchmark for Higher-Order Interaction Recovery in Network Dynamics

\shorttitle{A Generative Model for Higher-Order Network Dynamics} %%%for recto running head
\shortauthorlist{S\'anchez et al.} %%% for verso running head

\author{%%%% First author details
\name{Valentina S\'anchez$^{1,*}$,
      Theodore Papamarkou$^{2}$,
      Gonzalo N\'apoles$^{1}$,
      Çiçek G\"uven$^{1,\dagger}$ and
      Koen V. Haak$^{1,3,\dagger}$\\
      }
\address{$^{1}$Department of Intelligent Systems, Research Center for Cognitive Science and Artificial Intelligence, Tilburg School for Humanities and Digital Sciences, Tilburg University, Tilburg, The Netherlands\\
$^{2}$School of Applied Mathematical and Physical Sciences,
National Technical University of Athens, Athens, Greece\\
$^{3}$Donders Institute for Brain Cognition and Behaviour, Centre for Cognitive Neuroimaging,
Radboud University, Nijmegen, The Netherlands\\
$^{\dagger}$These authors contributed equally to this work and are joint last authors\\
\email{$^{*}$Corresponding author: v.sanchezmelchor@tilburguniversity.edu}}}

\maketitle

\begin{abstract}
{Conventional network representations inferred from multivariate time series describe statistical relationships between pairs of components, even when the underlying dynamics involve group interactions that cannot be represented by pairwise interactions alone. Several multivariate information-theoretic measures have been proposed to characterise dependence beyond pairs. However, due to a lack of systems with known generating mechanisms, these measures have not been carefully validated or compared for empirical use. To address this, we here introduce a generative model of stochastic network dynamics with pairwise and higher-order interactions specified by the user. The higher-order interactions are represented by directed hyperedges, and the model provides a controlled benchmark for evaluating information-theoretic measures. Interaction strengths and topology can be varied, and alternative higher-order coupling functions can be compared. Removing only the higher-order term gives a pairwise-only reference. The experiments test whether a measure detects changes induced by the higher-order term relative to this reference and recovers the participating nodes and their source-target roles. A separate experiment tests discrimination between a product coupling and a specified nonlinear pairwise control, with equal root-mean-square coupling magnitude under the pairwise-only reference. Several measures detected sufficiently strong higher-order interactions across the tested network topologies and coupling functions. In the separate discrimination experiment, the measures also separated the product coupling from the specified nonlinear pairwise control. Recovery, however, became harder as the number of candidate interactions grew. The framework therefore provides a testbed where the link between higher-order signatures and mechanisms can be tested, rather than assumed, before measures are applied to empirical data. The model also provides a reusable generator of synthetic time series with known pairwise and higher-order interactions for evaluating other models and statistical measures.}
{network dynamics; higher-order networks; multivariate information theory; generative models}
\end{abstract}

\section{Introduction}

% P1 --- the pairwise limit, the gap, and how we address it
The analysis of complex systems from multivariate time series has long relied on graph representations, in which nodes correspond to system components and edges summarise statistical dependence between pairs of observed time series
\cite{newman2010networks,estrada2012structure}. Higher-order interactions extend beyond pairwise relationships, allowing multiple source nodes to influence a target jointly through a coupling that cannot be expressed as a sum of separate source contributions \cite{battiston2020networks,bick2023higher}. At the level of observed data, studies of brain recordings and financial and epidemiological time series have identified statistical structure beyond that captured by pairwise summaries \cite{santoro2023higher,luppi2022synergy,varley2023sinfo}.

Multivariate information-theoretic measures are increasingly used to characterise statistical dependence among groups of variables in time series \cite{rosas2019oinfo,varley2023sinfo,luppi2024information}. Yet a higher-order statistical pattern can arise even when the underlying
interactions are purely pairwise \cite{rosas2022disentangling}. Assessing what these measures reveal about higher-order mechanisms requires evaluation on systems with known generating dynamics. We address this gap with a generative model of network dynamics in
which pairwise and higher-order interactions are specified in advance and can
be varied or switched off independently. We use this model for three tasks. For detection experiments, we test whether a measure detects changes induced by the higher-order term relative to a pairwise-only reference. For recovery experiments, the identity of the planted interaction
is withheld, and we test whether a measure can identify its participating
nodes and direction among all candidate interactions. For discrimination experiments, we test whether a measure separates the non-decomposable product coupling from a specified nonlinear pairwise control, with equal root-mean-square coupling magnitude under the pairwise-only reference. The central contribution is a generative benchmark for evaluating how statistical measures respond to known higher-order mechanisms and how reliably they recover the associated network structure.

% P2 --- existing methods and why their agreement is not enough
A growing body of methods characterises higher-order structure in empirical
data \cite{battiston2021physics,bianconi2021higher}. Topological approaches
describe the geometry and organisation of data-derived structures
\cite{otter2017roadmap,santoro2023higher}, whereas information-theoretic measures characterise the amount and organisation of statistical dependence among groups of variables \cite{rosas2019oinfo,varley2023sinfo}. These
traditions have largely developed in parallel, but Varley et al.~\cite{varley2025topology}
recently compared them directly using synthetic point clouds with known
topology and multivariate functional magnetic resonance imaging (fMRI) time series. They associated enclosed voids in three-dimensional point clouds with information-theoretic synergy. In the fMRI data, greater synergy dominance was associated with more numerous cavities and greater average persistence. The two
approaches can therefore converge on related structure in observed data. As
the authors note, however, both describe the statistical behaviour of a
system rather than the mechanism that generated it, so their agreement does not establish that the
shared structure arose from a higher-order mechanism. A generative model with known interaction mechanisms allows this relationship to be tested under controlled conditions.

% P3 --- testing measures against known mechanisms
This approach is relevant to neuroscience, where network representations are often inferred from statistical dependencies between regional time series instead of observations of interactions among brain regions \cite{bullmore2009complex}. Recent work has also compared information-theoretic and topological descriptors of higher-order dependence in fMRI data \cite{santoro2026charting}. For pairwise connectivity, this problem motivated
the work of Smith et al.~\cite{smith2011network}, which simulated fMRI data from
networks with known connectivity and evaluated whether network modelling
methods recovered the planted structure. For higher-order mechanisms, Robiglio et al.~\cite{robiglio2025synergistic} examined synergistic signatures and reconstruction of group interactions in simplicial Ising and social contagion models. Antonacci et al.~\cite{antonacci2026structural} assessed predictive improvements over additive models using analytical examples and stochastic autoregressive simulations. We extend the simulation approach of Smith et al.~\cite{smith2011network} to stochastic network dynamics with higher-order interactions. Our benchmark evaluates detection and recovery of participating nodes and source-target roles across coupling functions and network topologies, using pairwise-only references for comparison.

% P4 --- the model, the experiments, and what we find
The model builds on the neural state equation of dynamic causal modelling (DCM)
\cite{friston2003dcm}, which Smith et al.~\cite{smith2011network} used to generate
synthetic fMRI data with known connectivity. A separate higher-order term can be added or removed while retaining the other terms and their parameter values. The experiments assess how detection and recovery performance varies across coupling functions and network structures. A separate experiment tests whether the measures can separate the product coupling from a specified nonlinear pairwise control.

% P5 --- how the paper is organised
The remainder of the paper is organised as follows. Section~\ref{sec:background} briefly defines higher-order interactions,
their directed representation, and the distinction between interaction
mechanisms and statistical dependence, and introduces the measures evaluated.
Further background is provided in Appendix~\ref{app:background}.
Section~3 specifies the generative model and the experimental design. Section~4 presents the base-case results for detection, recovery and discrimination, followed by further tests of detection and recovery across network topologies and coupling functions. Section~5 concludes with the
implications and limitations of the framework and outlines directions for
future work.
\section{Background}
\label{sec:background}

We define higher-order interactions by the form of the coupling in the generating dynamics \cite{battiston2020networks,bick2023higher}. For two different source nodes $j$ and $k$ acting on a target $i \notin \{j,k\}$, let $\phi_{ijk}(z_j,z_k)$ denote their contribution as a function of their states $z_j,z_k\in\mathbb R$. We call this coupling higher-order if there are no univariate functions $g_{ij}$ and $g_{ik}$ such that
\begin{equation}
\label{eq:decomposability}
\phi_{ijk}(z_j,z_k)=g_{ij}(z_j)+g_{ik}(z_k)
\end{equation}
for all $(z_j,z_k)\in\mathbb R^2$. This criterion identifies an interaction through non-additivity in the source states \cite{lim2015interactions}. The product coupling $\phi_{ijk}(z_j,z_k)=z_jz_k$ satisfies this criterion, whereas a sum of separate source contributions remains pairwise even when those contributions are non-linear. We apply the criterion to coupling functions expressed in the original node states.

We represent each such interaction by a directed hyperedge $e=(i,\{j,k\})$, where $i$ is the target and $\{j,k\}$ is the unordered pair of source nodes \cite{gallo2022directed}. The hyperedge identifies the participating nodes and their roles, while $\phi_{ijk}$ specifies how the source states jointly affect the target. Exchanging $j$ and $k$ leaves the hyperedge unchanged, whereas changing the target within the triplet defines a different directed interaction.

The coupling functions describe interaction mechanisms, whereas information-theoretic measures describe statistical dependence in the resulting time series. This dependence reflects the combined effects of the pairwise and higher-order couplings, external inputs, and process noise. Pairwise mechanisms can produce statistical patterns beyond those captured by pairwise summaries \cite{rosas2022disentangling}. A multivariate statistical signature therefore does not establish that a higher-order mechanism generated it, and a weak response from a measure does not establish that a higher-order interaction is absent.

We use pairwise mutual information as a reference for dependence between nodes. O- and S-information \cite{rosas2019oinfo} and Streitberg information \cite{liu2024streitberg} describe dependence among groups of variables without assigning source and target roles. For recovery of these roles, we use the synergy term of a partial information decomposition under the minimum mutual information convention \cite{barrett2015exploration}, together with joint-source transfer entropy \cite{schreiber2000,novelli2019}. The former quantifies information supplied jointly about a target beyond that provided by the more informative source, while the latter quantifies directed predictive dependence. We also use a combined S/O ranking score for candidate node sets, defined in Section~\ref{sec:triplet-recovery}. Definitions and interpretations
of the measures are provided in Appendix~\ref{app:information-measures},
and estimator settings in Section~\ref{sec:estimators-implementation}. Further background on the DCM neural state equation and higher-order
mechanisms is provided in Appendices~\ref{app:dcm}
and~\ref{app:higher-order}.

\section{Methods}

This section defines the generative model and the experiments used to evaluate information-theoretic measures. In the four-node base case, we test whether the measures detect changes induced by a known higher-order interaction relative to a pairwise-only reference. We then assess recovery of its participating nodes and source-target roles with the interaction's identity withheld. A separate experiment tests discrimination between the product coupling and a nonlinear additive control. Further experiments assess how detection and recovery vary across alternative
coupling functions and more complex topologies.
Sections~\ref{sec:methods-model} and~\ref{sec:methods-coupling} define the model and coupling functions. Section~\ref{sec:experimental-setup} describes the base case, validation and experimental procedures. Section~\ref{sec:methods-topology-generalisation} describes the topology experiments and any changes to the base-case settings.

\subsection{Generative model}
\label{sec:methods-model}

We define the generative model by extending the DCM neural state equation
(Equation~\ref{eq:dcm}) so that it carries explicit higher-order interactions.
Rather than replacing that structure, we retain its decay, pairwise connectivity,
and external-input terms, and add a source-pair interaction term together with
additive process noise. Most experiments in this work concern directed
triplets, in which an unordered pair of source nodes acts jointly on a distinct
target node. The state of node $i$ evolves as
\begin{equation}
\label{eq:generative-general}
  dz_i(t)
  =
  \Bigg[
    -\lambda z_i(t)
    +\sum_{j\neq i}A_{ij}z_j(t)
    +\sum_{\substack{j<k\\j,k\neq i}}
      H_{ijk}\beta_{ijk}\,
      \phi_{ijk}\bigl(z_j(t),z_k(t)\bigr)
    +(Cu(t))_i
  \Bigg]dt
  +\sigma_z\,dW_{z,i}(t).
\end{equation}

Here, $z\in\mathbb R^N$ is the vector of node states and
$\lambda>0$ is the self-decay rate, shared by all nodes. The matrix $A\in\mathbb R^{N\times N}$ is the weighted
pairwise connectivity matrix, where $A_{ij}$ gives the strength and sign of
the directed influence of source $j$ on target $i$; self-decay is represented
separately, so $A_{ii}=0$. The binary tensor
$H\in\{0,1\}^{N\times N\times N}$ encodes the directed-triplet topology:
$H_{ijk}=1$ indicates that the unordered source pair $\{j,k\}$ jointly acts on
target $i$. The restriction $j<k$ ensures that each source pair is counted
only once. The coefficient $\beta_{ijk}\in\mathbb R$ is a scalar that sets the strength and
sign of the interaction, and
$\phi_{ijk}\colon\mathbb R^2\rightarrow\mathbb R$ is its coupling function.
Finally, $u(t)\in\mathbb R^M$ is the vector of external inputs,
$C\in\mathbb R^{N\times M}$ is a matrix that determines how
those inputs enter each node; the $W_{z,i}$ are independent standard Wiener
processes, and $\sigma_z$ is the process-noise intensity. The input process $u$ is generated independently of the Wiener processes $W_{z,i}$.

Unless otherwise stated, the experiments use the product
coupling function
$\phi_{ijk}(z_j,z_k)=z_jz_k$. Substituting this function into
Equation~\ref{eq:generative-general} gives a joint contribution
$H_{ijk}\beta_{ijk}z_jz_k$ to the drift of target $i$. The base case contains
the single directed interaction $(1,2)\rightarrow0$, specified under the
$j<k$ convention by $H_{012}=1$ and $\beta_{012}=\beta$, with all other
entries of $H$ equal to zero. Section~\ref{sec:methods-coupling} explains the
choice of the product coupling and its extension to larger source sets. The
dedicated coupling-function sensitivity analysis replaces the product with
the alternative functions defined in that section while leaving all other
model components unchanged.

The drift in Equation~\ref{eq:generative-general} decomposes
the dynamics into four interpretable components.
The decay term $-\lambda z_i$ provides self-regulation, pulling each node back
toward baseline and opposing unbounded activity in the absence of sustained
input. The pairwise term $\sum_{j\neq i}A_{ij}z_j$
represents pairwise connectivity, with each source
contributing additively to the target according to the signed weight
$A_{ij}$. The higher-order term
$\sum_{j<k}H_{ijk}\beta_{ijk}
\phi_{ijk}(z_j,z_k)$ captures joint source-pair effects that cannot be
represented as a sum of separate pairwise influences. Finally, the input term $(Cu)_i$ models experimentally imposed or exogenous
drive, with $C$ controlling which nodes receive each input channel. The terms inside the square brackets form the drift, or
systematic part of the state dynamics. The Wiener term
$\sigma_z\,dW_{z,i}(t)$ instead adds random fluctuations directly to the
evolution of the node states. It is separate from the external inputs
$u(t)$, and no additional observation noise is included in the primary
model.

The decay and pairwise terms form the linear backbone of the model,
$\dot z=(A-\lambda I)z+Cu$. Setting $u\equiv0$,
$H_{ijk}=0$ for all $i,j,k$, and $\sigma_z=0$ gives
$\dot z=(A-\lambda I)z$, whose solution is the matrix exponential
$z(t)=e^{(A-\lambda I)t}z(0)$, so stability is governed by the eigenvalues of
$A-\lambda I$. The backbone is asymptotically stable when
$\alpha(A)<\lambda$, where $\alpha(A)$ is the spectral abscissa of $A$;
intuitively, self-decay must dominate recurrent pairwise amplification. The
product higher-order term can produce nonlinear amplification, so large
$\beta_{ijk}$, strong inputs, or large initial conditions
can produce unbounded trajectories even when the linear backbone is stable.
Linear stability does not guarantee that solutions of the nonlinear stochastic model remain finite for all times. We assess numerical boundedness over the simulation interval and examine timestep convergence in selected experimental conditions.

We integrate the stochastic state equation using Euler--Maruyama,
\begin{equation}
\label{eq:euler-maruyama}
  z_{n+1}
  =
  z_n+f(z_n,u_n)\Delta t
  +\sigma_z\sqrt{\Delta t}\,\xi_n,
  \qquad
  \xi_n\overset{\mathrm{i.i.d.}}{\sim}\mathcal N(0,I_N),
\end{equation}
where $f$ denotes the drift in
Equation~\ref{eq:generative-general}.

\subsection{Coupling function}
\label{sec:methods-coupling}

The tensor $H$ determines which source-pair interactions are present, whereas
the coupling function $\phi_{ijk}$ determines how the states of each source
pair are combined before entering the drift of target $i$. We require every
$\phi_{ijk}$ to satisfy two local conditions at the origin,
\begin{equation}
\label{eq:coupling-local-conditions}
  \phi_{ijk}(0,0)=0,
  \qquad
  \nabla\phi_{ijk}(0,0)=\mathbf 0.
\end{equation}
The first condition ensures that the higher-order term contributes nothing at
baseline, leaving $z=0$ as an equilibrium of the unforced deterministic
drift. The second ensures that the interaction does not enter the Jacobian at
the origin, which remains $A-\lambda I$. The higher-order term is therefore
absent from the first-order Taylor linearisation of the drift at the origin.
These are local conditions and do not imply that the interaction is invisible
to linear or pairwise statistics computed away from the origin. We additionally require $\phi_{ijk}$ to be non-decomposable in the sense of
Equation~\ref{eq:decomposability}, so that the joint source-pair contribution
cannot be absorbed into separate pairwise terms.

The main experiments use the product coupling
$\phi_{ijk}(z_j,z_k)=z_jz_k$. Its contribution to target $i$ takes the sign of
$\beta_{ijk}$ multiplied by the product of the source states, so for positive
$\beta_{ijk}$ it is positive when the sources have the same sign and negative
when their signs differ. This choice is motivated on three grounds.  First, the two-source product has precedent in the DCM framework, where
Stephan et al.~\cite{stephan2008nonlinear} introduce a term in which the activity of one
region modulates the influence of another. We use the same bilinear product
but treat the sources as an unordered pair because $z_jz_k=z_kz_j$. Second,
it is the lowest-degree polynomial term that depends jointly on both sources.
Such products also form the first additional coordinates beyond the original
node states when polynomial dynamics are represented in an enlarged state
space \citep{lacasa2026}. Third, the product satisfies both conditions in
Equation~\ref{eq:coupling-local-conditions} and the non-decomposability
criterion by construction.

For experiments involving a larger source set $S$, the product coupling is
extended as
\[
  \phi_{iS}(z_S)
  =
  \prod_{j\in S}z_j.
\]
This extension satisfies the corresponding local conditions and is
non-decomposable into additive functions of proper subsets of $S$. Its
$|S|$-fold mixed partial derivative is one, whereas any term omitting one of
the sources has a zero derivative of that order.

The coefficient $\beta_{ijk}$ sets the strength and sign of a source-pair
interaction. It is not directly comparable with a pairwise coefficient
$A_{ij}$ because the realised scale of $z_jz_k$ depends on the amplitudes of
both source states. Scaling every source state by the same factor $c>0$ multiplies the product coupling $\prod_{j\in S}z_j$ by $c^{|S|}$. We therefore
report realised component-wise scales and assess boundedness for each
experimental condition.

To assess whether the conclusions depend on the bilinear product, we compare it
with three alternative couplings described in
Section~\ref{sec:methods-coupling-sensitivity}; their results are reported in
Sections~\ref{sec:results-coupling-sensitivity}
and~\ref{sec:coupling-interaction-order}. For any source set $S$, these are defined
by the factorised family
\begin{equation}
  \phi_{m,iS}(z_S)
  =
  a_{m,S}
  \prod_{j\in S}
  g_m\!\left(\frac{z_j}{s_{j,S}}\right),
  \label{eq:alternative-coupling-family}
\end{equation}
where $m$ labels the coupling and $g_m$ its source-response function,
$s_{j,S}$ rescales each source to an order-one input, and $a_{m,S}$ rescales
the resulting coupling output. The two-source case is obtained by setting
$S=\{j,k\}$.

The source-wise transformations, applied separately to each source before
forming their joint product, are
\begin{equation}
\begin{aligned}
  g_{\tanh}(x) &= \tanh(x),\\
  g_{\sin}(x) &= \sin(x),\\
  g_{\mathrm{root}}(x)
  &= \frac{x}{(x^2+\varepsilon^2)^{1/4}},
  \qquad \varepsilon=0.25.
\end{aligned}
\label{eq:alternative-source-responses}
\end{equation}

The regularised denominator makes
$g_{\mathrm{root}}$ smooth at zero, while its large-amplitude behaviour is
approximately $\operatorname{sign}(x)\sqrt{|x|}$. It therefore provides a
smooth signed-root response without the non-differentiability of the exact
signed square root. All three source-response functions satisfy
$g_m(0)=0$ and $g_m'(0)\neq0$. For $|S|\geq2$, their factorised couplings satisfy the corresponding local conditions in Equation~\ref{eq:coupling-local-conditions}. For non-zero $a_{m,S}$ and positive $s_{j,S}$, the mixed partial derivative involving all $|S|$ source variables at the origin is $a_{m,S}\prod_{j\in S}\frac{g_m'(0)}{s_{j,S}}\neq0$. These couplings therefore cannot be expressed as sums of functions of proper subsets of $S$.

The four mechanisms represent different source responses.
The product is bilinear and unbounded. Tanh is monotone and saturating, sine
is bounded and oscillatory, and the smooth signed root is monotone and
sublinear. These alternatives test sensitivity to coupling function.

\subsection{Experimental setup \& Validation}
\label{sec:experimental-setup}

The model of
Sections~\ref{sec:methods-model}
and~\ref{sec:methods-coupling}
is defined up to its topology, coupling function, weights, input process, and noise intensities.
This section fixes those choices in a single four-node configuration, reports
the numerical and dynamical validation of that configuration, and specifies the
null-calibration procedures used for detection and recovery experiments.
This configuration is also the reference point for the topology experiments
in Section~\ref{sec:methods-topology-generalisation}.
The experiments in  Section~\ref{sec:methods-topology-generalisation} retain its dynamics and coupling convention while varying the network topology, reusing initial states and stochastic paths for nodes shared across layouts.

\newpage

\subsubsection{Base case}
\label{sec:base-case}

\begin{figure}[H]
  \centering
  \includegraphics[width=0.55\linewidth]{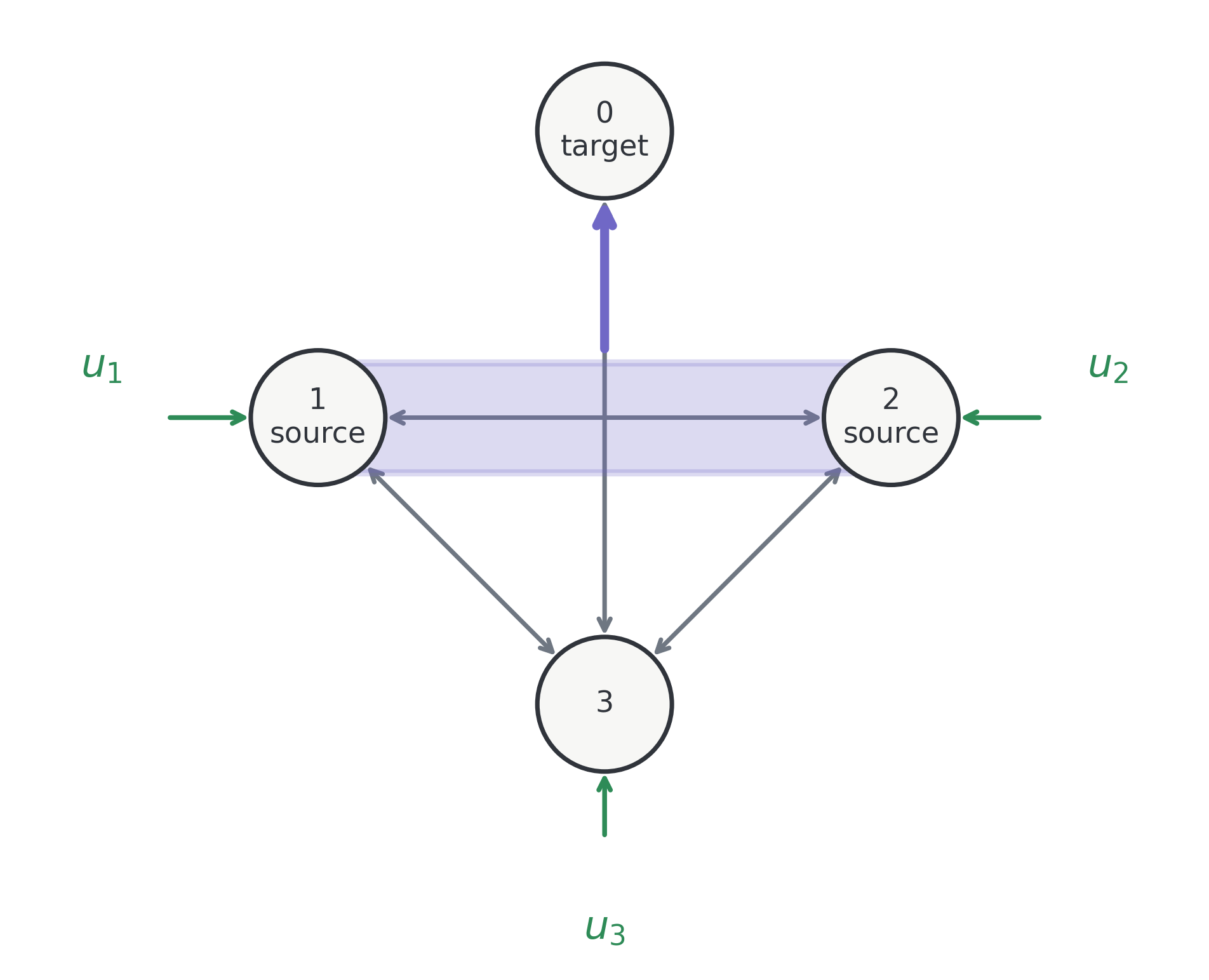}
  \caption{Four-node base system. Grey arrow pairs represent
reciprocal directed edges whose weights may differ. The coloured band and
arrow represent the planted higher-order interaction
$(1,2)\rightarrow0$. Green arrows denote independent external inputs;
node~0 receives no direct input.}
  \label{fig:base_topology}
  \par\smallskip
{\small\raggedright
\textbf{Alt text:} Four-node network with reciprocal pairwise links
and a joint interaction from nodes one and two to node zero.
External inputs enter nodes one, two and three.\par}
\end{figure}

Figure~\ref{fig:base_topology} shows the base case, and
Table~\ref{tab:base-configuration} lists its full configuration.
This configuration is used first to assess whether the
planted higher-order interaction can be detected and then whether its
participating nodes and direction can be recovered. The planted interaction
corresponds to the unordered node set $\{0,1,2\}$ and the directed hyperedge
$(1,2)\rightarrow0$, where nodes~1 and~2 are the sources and node~0 is the
target. With the convention
that $A_{ij}$ is the influence of source $j$ on target $i$, in node order
$(0,1,2,3)$, the weighted pairwise template and its spectrally normalized
adjacency matrix are
\[
T=\begin{pmatrix}
0&0&0&0.06\\
0&0&0.30&0.16\\
0&0.30&0&0.10\\
0.05&0.08&0.12&0
\end{pmatrix},
\qquad
A=\frac{r_A\lambda}{\alpha(T)}T.
\]

\begin{table}[!ht]
  \centering
  \caption{Configuration of the four-node base case.}
  \label{tab:base-configuration}
  \begin{tabular}{p{0.20\linewidth}p{0.65\linewidth}}
    \hline
    \textbf{Component} & \textbf{Chosen value} \\
    \hline
    \noalign{\vskip 0.5em}
    Nodes & $N=4$ \\
    Decay & $\lambda=1$ \\
    Pairwise topology & $\{0\leftrightarrow3,\ 1\leftrightarrow2,\
      1\leftrightarrow3,\ 2\leftrightarrow3\}$ \\
    Pairwise weights & $A=\dfrac{r_A\lambda}{\alpha(T)}T,\quad
      r_A=0.35,\quad \alpha(A)=0.35\lambda$ \\
    Higher-order topology & $(1,2)\rightarrow0$ \\
    Higher-order weight & $\beta\in\{0,0.2,0.5,1,2,5,10,20,30\}$ \\
    Input contribution & $(Cu)_i=g\,u_i$ for $i\in\{1,2,3\}$, $(Cu)_0=0$,
      $g=0.02285$ \\
    Input signals & $u_i\sim\mathrm{OU}(\tau_u,\sigma^2{=}1)$, independent,
      $\tau_u=0.2756$ ($\rho=0.93$ at $\Delta t=0.02$) \\
    Process noise & Independent isotropic Wiener noise,
      $\sigma_z=4.24\times10^{-4}$; at $\Delta t=0.02$, the per-step
      increment standard deviation is
      $\sigma_z\sqrt{\Delta t}=6.00\times10^{-5}$ \\
    Initial state & $z_i(0)\overset{\mathrm{i.i.d.}}{\sim}
      \mathcal N(0,0.1^2)$ \\
    Integration & Euler--Maruyama, $\Delta t=0.02$, $t\in[0,100]$
      ($5001$ samples); discard $t<10$, retaining $4501$ samples \\
    Randomness & Distinct recorded seeds across replicates; initial states,
      OU paths, and Wiener paths reused across $\beta$ within each replicate;
      observation noise $=0$ \\
    \hline
  \end{tabular}
\end{table}

The use of heterogeneous rather than identical connection strengths is
motivated by the network simulations of Smith et
al.~\cite{smith2011network}, in which non-zero off-diagonal strengths were
drawn with mean $0.4$ and standard deviation $0.1$ and restricted to
$[0.2,0.6]$. Their numerical values are not transferred directly because the
decay convention, topology, and overall dynamical scaling differ from ours.
Instead, we use a fixed heterogeneous template whose non-zero entries range
from $0.05$ to $0.30$. After spectral normalization, the corresponding
non-zero entries of $A$ range from $0.0471$ to $0.2828$. Fixing the relative
weights across replicates ensures that comparisons across $\beta$ isolate the
higher-order term, while retaining the connection heterogeneity used in
realistic network simulations.

The overall pairwise strength is controlled by a single spectral scale rather
than by tuning individual edges. Four nodes are the smallest system containing
the planted triplet together with an additional node that does not belong to
the planted interaction. Node~3 lies on every pairwise path from either source to the target, so source-target statistical dependence can arise even without direct source-target edges. The linear backbone $\dot z=(A-\lambda I)z$ is asymptotically stable while
the spectral abscissa $\alpha(A)$ remains below the decay rate $\lambda$. We set
$\alpha(A)=r_A\lambda$ with $r_A=0.35$, leaving a stability margin of
$(1-r_A)\lambda=0.65\lambda$.

External inputs enter as $Cu(t)$, the product of the input matrix $C$ and the
signal $u(t)$. Each component of $u$ is an independent stationary
Ornstein--Uhlenbeck process,
\[
du_i(t)=-\frac{1}{\tau_u}u_i(t)\,dt
+\sqrt{\frac{2}{\tau_u}}\,dW_i^{(u)}(t),
\]
with zero mean, unit variance, and independent Wiener processes
$W_i^{(u)}$. We integrate it exactly at each step via
\[
u_{i,k+1}
=\rho u_{i,k}
+\sqrt{1-\rho^2}\,\varepsilon_{i,k},
\qquad
\rho=\exp(-\Delta t/\tau_u),
\]
where $\varepsilon_{i,k}\sim\mathcal N(0,1)$. We draw
$u_i(0)\sim\mathcal N(0,1)$, so that each input process is stationary from the
first simulated sample. We set $\tau_u=0.2756$, corresponding to $\rho=0.93$ at
$\Delta t=0.02$. Sources~1 and~2 and node~3 each receive a separate channel
through a sparse $C$ with one common gain $g=0.02285$, avoiding node-specific
input tuning. The target receives no direct input. After warmup, its activity
is explained by decay, the node-3-mediated pairwise route, process noise, and,
when present, the higher-order term. Process noise represents endogenous fluctuations not captured by the
specified inputs or coupling terms.

Initial states, OU paths, and process-noise paths vary across independent
replicates but are drawn from fixed, recorded seeds.
Whenever conditions are compared across the higher-order
coupling coefficient $\beta$, they reuse the same initial state, OU path, and
process-noise path within each replicate. Duration, input correlation time,
input gain, warmup, and sampling are also held fixed. This design
isolates the effect of changing the higher-order coupling strength. We use OU inputs to represent temporally correlated drive alongside the white process noise. Because
neighbouring observations are therefore not independent,
Table~\ref{tab:base-configuration} reports the simulation duration and
$\tau_u$ alongside the sample count. Null calibration uses independent
simulations with the same OU parameters and sampling procedure as the
evaluation simulations. We do not permute individual time points, because
permutation would remove their temporal dependence.

\subsubsection{Model validation}
\label{sec:model-validation}

The linear stability margin established above applies to the pairwise
backbone near baseline, but does not establish numerical accuracy or
boundedness of the complete nonlinear model. We therefore assessed timestep convergence at
$\Delta t\in\{0.02,0.01,0.005\}$ over 12 replicates for each
$\beta\in\{0,20\}$. Within each replicate, the same initial state and underlying
OU and Wiener paths were used across timesteps. For each replicate, we express the target-state RMSE as a percentage of the RMS of its reference trajectory at $\Delta t=0.005$, evaluating both trajectories at common sampling times. Relative to the reference step $\Delta t=0.005$, the median
target-state RMSE at $\Delta t=0.02$ was $1.89\%$ for $\beta=0$ and
$3.03\%$ for $\beta=20$. Halving the step to $\Delta t=0.01$ reduced these
values to $0.79\%$ and $1.19\%$, but doubled the number of integration steps.
We therefore retained $\Delta t=0.02$ as the computational operating point for
the full benchmark. Complete results are reported in
Figure~\ref{fig:validation-convergence} and
Table~\ref{tab:validation-convergence} in the Appendix.

One-factor-at-a-time sensitivity tests used
$\beta\in\{0,20\}$, the null and recovery operating points; the full $\beta$
sweep is evaluated in Section~\ref{sec:c-sensitivity}. We varied the pairwise
spectral fraction over $\{0.20,0.35,0.50\}$ and the input correlation time,
input gain, and process-noise intensity over $\{0.5,1,2\}$,
$\{0.5,1,1.5\}$, and $\{0,0.5,1,2\}$ times their base values, respectively.
Initial states and stochastic paths were generated using the same
seeds across the tested parameter values. For every
setting and each $\beta\in\{0,20\}$, all 30 simulations remained finite and
satisfied $\max_{i,t}|z_i(t)|<10^{6}$. This is a finite-time check for
numerical blow-up rather than a proof of global stability. At $\beta=20$, the
median target-node higher-order-to-decay drift RMS ratio remained at least
$1.09$, and the higher-order-to-pairwise ratio remained at least $2.41$,
confirming that the planted term remained dynamically substantial across the
tested conditions.

Finally, although the two sources receive independent external inputs, their
pairwise connections induce dependence between their states. Their median
zero-lag Pearson correlation ranged from $0.219$ to $0.517$, showing that the
product interaction was evaluated against moderate network-induced source
dependence rather than synchronisation imposed by a shared input.
Condition-wise results are reported in
Figure~\ref{fig:validation-sensitivity} and
Table~\ref{tab:validation-sensitivity} in the Appendix.

\subsubsection{Higher-order interaction detection}
\label{sec:c-sensitivity}

Detection first validates the generative model by confirming that the planted
higher-order term leaves a measurable signature before recovery is attempted.
It is evaluated over the coupling strengths in Table~\ref{tab:base-configuration} using an adaptation of
the c-sensitivity concept of Smith et al.~\cite{smith2011network}, which measures
how often each estimator distinguishes the higher-order condition from the
pairwise-only generative null.

Every estimator is applied to the post-warm-up time series
of the three participating nodes, $(z_0,z_1,z_2)$. We analyse the node states
directly, without first removing the effects predicted by the pairwise
dynamics. Before estimation, each series is centred and scaled to unit
variance.

O-information, S-information, and Streitberg information are computed jointly
on the three variables. To test what an analysis restricted to individual
target--source relationships can detect, we use
\[
  M_{\mathrm{MI}}
  =
  \max\!\left\{I(z_0;z_1),I(z_0;z_2)\right\}.
\]
Taking the maximum gives the pairwise baseline its strongest score, while
avoiding the mixture of unique, redundant, and synergistic contributions
contained in joint-source mutual information.

The calibration procedure is applied separately to each metric. An independent
ensemble of $N_0=200$ simulations with $\beta=0$ provides the calibration
scores $B_1,\ldots,B_{N_0}$. These simulations are not reused as detection
replicates. Let $\widehat{\mu}_0$ and $\widehat{\sigma}_0$ denote the empirical
mean and standard deviation of the calibration scores. We define the absolute
standardized departure of a score $x$ as
\[
  D(x)
  =
  \left|
    \frac{x-\widehat{\mu}_0}{\widehat{\sigma}_0}
  \right|.
\]
The detection threshold is
\[
  q
  =
  Q_{0.95}\!\left(\{D(B_j)\}_{j=1}^{N_0}\right),
\]
the empirical 95th percentile of the standardized departures in the
calibration ensemble. Once estimated, $\widehat{\mu}_0$,
$\widehat{\sigma}_0$, and $q$ are held fixed throughout the detection experiment. The threshold targets a $5\%$ false-positive rate for each measure. The false-positive probability may differ from $0.05$ because the threshold is estimated from a finite calibration ensemble.

For each value of $\beta$, we generate $N_\beta=40$ separate detection
replicates. Within each replicate, the initial state, OU input path, and
Wiener process-noise path are held fixed across $\beta$, so that the
higher-order coefficient is the only manipulation. The $\beta=0$ detection
condition is independent of the calibration ensemble and provides a held-out
check of the false-positive rate.

If $M_r(\beta)$ is the score from detection replicate $r$, the
null-calibrated c-sensitivity of metric $m$ is
\begin{equation}
\label{eq:c-sensitivity}
  \widehat{c}_m(\beta)
  =
  \frac{1}{N_\beta}
  \sum_{r=1}^{N_\beta}
  \mathbf{1}\!\left\{
    D\bigl(M_r(\beta)\bigr)>q
  \right\}.
\end{equation}
This is the fraction of detection replicates whose absolute standardized deviations from the calibration mean exceed $q$. For each measure and value of $\beta$, we report pointwise Wilson $95\%$ confidence intervals for the detection probability conditional on $\widehat{\mu}_0$, $\widehat{\sigma}_0$, and $q$. These intervals quantify uncertainty from the detection replicates without including uncertainty from estimating the calibration parameters. The detection rule is two-sided because the included
information measures do not share a common signed alternative: evidence may
appear as either an increase or a decrease relative to the null distribution.

This definition is inspired by, but differs from, the original c-sensitivity
of Smith et al.~\cite{smith2011network}, which compared scores for true network edges with
the within-network distribution of scores for absent edges. Our adaptation
instead estimates how often a metric detects departure from a pairwise-only
generative null under the specified model and sampling design. It therefore
evaluates the estimator against known generative conditions, using independent
calibration and detection ensembles and a pairwise-only
counterfactual. The resulting c-sensitivity is a repeated-simulation detection
rate, not the probability that an observed dataset was generated by a
higher-order mechanism.

\subsubsection{Triplet recovery}
\label{sec:triplet-recovery}

Following detection, the recovery benchmark withholds the planted identity and
asks whether it can be inferred from the observations, more closely reflecting
analysis of empirical data. Unordered-set recovery identifies the three
participating nodes, whereas directed recovery also distinguishes the sources
from their target. Both use the same simulation ensembles but rank different
candidate spaces. The chance probabilities assume uniform random ranking and can differ from selection probabilities under the $\beta=0$ model.

Recovery is evaluated at $\beta=20$, the smallest tested coupling strength at
which KSG S-information, our primary detector, and the KSG mutual-information
reference both attained a c-sensitivity of $1$ across all 40 detection
replicates (Section~\ref{sec:results-base-detectability}). This value is fixed
across all recovery measures and evaluated using simulations separate from the
detection experiment. We use $N_0=100$ independent $\beta=0$ simulations for calibration
and 60 test replicates, each using the same seeds for the
$\beta=0$ and $\beta=20$ conditions. After warmup, estimation time points are spaced by 14 integration
steps, or approximately one OU correlation time.

In the unordered-set experiment, the symmetric measures rank the four possible
three-node sets using states recorded at the same retained time points. We use
KSG S-information, KSG O-information, the combined S/O score, and an
entropy-kNN estimate of O-information from HOI. S-information uses the upper null tail because larger values
indicate stronger collective dependence, whereas both O-information estimates
use the lower tail because more negative values indicate a more
synergy-dominated departure. The chance probability of ranking the planted set $\{0,1,2\}$ first is
$1/4$. Because these measures are symmetric, they identify the participating
nodes but not their source and target roles.

In the directed experiment, the target-specific measures rank all
$4\times\binom{3}{2}=12$ assignments of one target and two sources. For
candidate $(j,k)\rightarrow i$, the sources are evaluated as $z_j(t)$ and
$z_k(t)$ and the target as $z_i(t+\Delta t)$, with $\Delta t=0.02$.
Conditional variants additionally condition on $z_i(t)$. The target-specific measures use their upper null tails
because larger values indicate stronger target-specific synergy or predictive
dependence. The inferred direction represents temporal prediction
and does not by itself establish causality. We report the correct target, the
correct source pair among candidates with target~0, and the exact candidate
$(1,2)\rightarrow0$, with chance probabilities $1/4$, $1/3$, and $1/12$,
respectively.

Each measure and candidate is calibrated against its own distribution from the
independent $\beta=0$ ensemble. Candidate-specific calibration accounts for
differences in baseline scores caused by network position and source--target
assignment. For an upper-tail score $v$,
\[
p_{\mathrm{upper}}(v)
=
\frac{1+\sum_{r=1}^{N_0}\mathbf 1[v_r^{(0)}\geq v]}{N_0+1},
\]
with the inequality reversed for a lower-tail score. The combined score is
$Q=-\log p_{S,\mathrm{upper}}-\log p_{O,\mathrm{lower}}$. For each candidate,
$Q$ is calibrated against leave-one-out values from its null ensemble, thereby
accounting for dependence between S-information and O-information.

Candidates are ranked by increasing empirical probability of equally or more extreme scores under the pairwise-only model. Ties are broken by decreasing $(v-\widehat{\mu}_0)/\widehat{\sigma}_0$ when larger scores are favoured and decreasing $(\widehat{\mu}_0-v)/\widehat{\sigma}_0$ when smaller scores are favoured. For each measure and candidate, $\widehat{\mu}_0$ and $\widehat{\sigma}_0$ are the mean and standard deviation of its calibration scores. These probabilities provide a
common ranking scale rather than multiplicity-adjusted significance tests. The
planted identity is withheld until the rankings are complete. The
$\beta=0$ condition provides the negative control, and the $\beta=20$
condition evaluates recovery when the interaction is present. We report pointwise Wilson $95\%$ confidence intervals for the top-rank selection probabilities, conditional on the calibration ensemble.

\subsubsection{Coupling-function sensitivity analysis}
\label{sec:methods-coupling-sensitivity}

The coupling-function sensitivity analysis included two
complementary experiments. The first compared coupling functions in the
two-source base case (Section~\ref{sec:results-coupling-sensitivity}). At each
value of $\beta$, the coupling function was the only model component varied;
the topology, remaining model parameters, initial states, external-input
paths, and Wiener process-noise paths were held fixed. The second repeated
this controlled comparison across interaction sizes
(Section~\ref{sec:coupling-interaction-order}). Within each fixed layout and
interaction size, only the coupling function was varied.

In the base case, the product coupling was compared with the sourcewise tanh,
sine, and smooth signed-root functions defined in
Section~\ref{sec:methods-coupling}. Their normalisation constants were
estimated from 200 independent $\beta=0$ simulations. This scaling ensemble
was separate from the detector-null and test ensembles. For each source, we
selected $s_j$ such that
\[
  Q_{0.95}\!\left(\left|z_j/s_j\right|\right)=1.5.
\]
This placed the source states on a comparable input scale. The amplitude
constant $a_m$ was then chosen so that each alternative function had the same
RMS magnitude as the product $z_jz_k$ on the scaling ensemble. All constants were estimated once from the scaling ensemble and then held
fixed across all evaluation simulations.
After this rescaling, the remaining difference from the product was
measured by
\[
  \mathrm{NRMSE}_m
  =
  \frac{\mathrm{RMS}(\phi_m-\phi_{\mathrm{prod}})}
       {\mathrm{RMS}(\phi_{\mathrm{prod}})}.
\]
Detector performance was evaluated in 40 test replicates at
$\beta\in\{0,10,20,30\}$. At $\beta=20$, we also compared the RMS of the higher-order state increments $\beta\phi_m(z_1,z_2)\Delta t$ at target node~0 across coupling functions.

For the interaction-order analysis, the same coupling family was applied in the $N=16$ network to source sets containing two, three, or four nodes, corresponding to total interaction sizes of three, four, or five nodes after including the target.
Each layout used its own set of 200 independent $\beta=0$
scaling simulations. Its source-normalisation constants $s_{j,S}$ were
determined by applying the source-scaling rule defined above for the
two-source, four-node coupling comparison.

The product coupling used in the main analyses remained unnormalised. For this
sensitivity experiment only, we added the normalised-product control
\[
  \phi_{\mathrm{norm},iS}(z_S)
  =
  a_{\mathrm{norm},S}
  \prod_{j\in S}\frac{z_j}{s_{j,S}}.
\]
This control tests whether differences across interaction sizes arise from
multiplying an increasing number of small source states. The normalised product
and the three alternative functions were scaled to the RMS magnitude
of the size-three product coupling. Their constants were then fixed and used
in 40 test replicates at $\beta=20$ for each interaction size.

Each layout used its own independent set of 200 $\beta=0$
detector-null simulations, separate from the scaling and test ensembles. The metric-specific detection thresholds derived from these null simulations
were fixed before detector performance was evaluated. Unordered-set recovery used a separate ensemble of 100 $\beta=0$
calibration simulations and 60 test replicates per layout, with
initial states, OU input paths, and Wiener process-noise paths
reused across coupling functions. Candidate-specific calibration and ranking followed Section~\ref{sec:triplet-recovery}. All conditions were assessed using the finite-time boundedness check in Section~\ref{sec:model-validation}. Timestep comparisons for the four coupling functions in the base case at $\beta=20$ are reported in Section~\ref{sec:results-coupling-sensitivity}.

\subsubsection{Discrimination from a decomposable control}
\label{sec:discrimination}

Detection asks whether the planted term changes the system, while
recovery asks whether its participating nodes and roles can be
identified. We additionally tested whether the measures distinguish
the non-decomposable product coupling from a specified decomposable alternative.
In the four-node base case, we compared
\[
\phi_{\mathrm{prod}}(z_1,z_2)=z_1z_2,
\qquad
\phi_{\mathrm{add}}(z_1,z_2)
=a\bigl(z_1|z_1|+z_2|z_2|\bigr).
\]
The additive control can be incorporated into separate nonlinear
pairwise contributions to the target equation, whereas the product
cannot be expressed as a sum of single-source functions.
This distinction is algebraic. The experiment tests whether it
produces distinguishable statistical signatures in the observed
dynamics. Both functions vanish with zero gradient at the
origin, satisfying Equation~\eqref{eq:coupling-local-conditions}.
The baseline is the $\beta=0$ condition, in which neither additional
coupling contributes and the original pairwise dynamics and
stochastic inputs remain unchanged.

Using 200 independent baseline scaling simulations, we fitted
$a\approx0.3994$ so that
\[
\operatorname{RMS}_{0}(\phi_{\mathrm{add}})
=
\operatorname{RMS}_{0}(\phi_{\mathrm{prod}}).
\]
The coefficient was held fixed across $\beta$. This gives the coupling outputs the same RMS on the baseline
ensemble, without equating their means, temporal dependence or realised effects on the target. Both couplings entered the same target equation with coefficient
$\beta$, with all other dynamical parameters unchanged.
Initial states, OU inputs and Wiener paths were reused across
conditions and strengths within each replicate.

Simulations ran to $T=5000$ with $\Delta t=0.02$ and a 10-unit
warmup. Retaining every 250th sample yielded 999 observations
spaced by 5 time units. This sampling scheme was selected using independent
linear-Gaussian baseline checks to reduce the distortion observed
in densely sampled KSG estimates (see Appendix~\ref{app:discrimination-validation}).
Metric definitions and estimator settings were unchanged.

At each $\beta\in\{0,5,10,20,30\}$, 200 independent additive simulations
calibrated the two-sided c-sensitivity rule of
Equation~\eqref{eq:c-sensitivity}, replacing the baseline reference
with the additive reference at the corresponding strength. Discrimination c-sensitivity was measured as the fraction of 200 held-out product
simulations whose scores fell outside the calibrated additive-reference
interval. Applying the same criterion to 200 held-out additive
simulations provided the false-positive rate.

An independent validation at $\beta=20$ used 1000 additive calibration
simulations and 400 held-out simulations per coupling form.
The score-distribution figure uses this validation ensemble,
including 400 baseline simulations; the across-$\beta$ curves use
the 200-replicate test ensembles.
Scaling, calibration and test ensembles were independent.
Wilson 95\% intervals are pointwise and conditional on the fitted
thresholds. This comparison tests discrimination from the
signed-square additive control; it does not establish discrimination
from other choices of decomposable coupling.

\subsubsection{Estimators and implementation}
\label{sec:estimators-implementation}

Before estimation, each node-state series is centred and scaled to unit
variance. Mutual information and conditional mutual information are computed with the
IDTxl toolbox \citep{wollstadt2019}, which wraps the JIDT estimators
\citep{lizier2014}, and O- and S-information are computed directly with JIDT. The Gaussian estimates evaluate the corresponding information measures for a multivariate Gaussian distribution using the estimated covariance matrix. All KSG estimates use $k=4$, the documented default in both toolkits, as a compromise between
local resolution and sampling variability. This value was fixed before
analysis across all conditions, null ensembles and candidates, and no noise
or dynamic-correlation exclusion was added. The HOI entropy-kNN estimate of
O-information~\citep{neri2024} also uses $k=4$. Streitberg information uses
the reference Tsallis-$\alpha$ kNN implementation of
Liu et al.~\cite{liu2024streitberg}, with $\alpha=0.5$ and $k=30$, and is therefore not
labelled as either Gaussian or KSG. Package-level permutation tests are not used; detection and recovery
are instead calibrated against independently simulated $\beta=0$
ensembles, as described in Sections~3.3.3 and~3.3.4,
while discrimination is calibrated against independent additive-control
simulations at the corresponding coupling strength, as described in
Section~3.3.6. 

\subsection{Generalisation across topology regimes}
\label{sec:methods-topology-generalisation}

\begin{figure}[!ht]
  \centering
  \includegraphics[width=\linewidth]{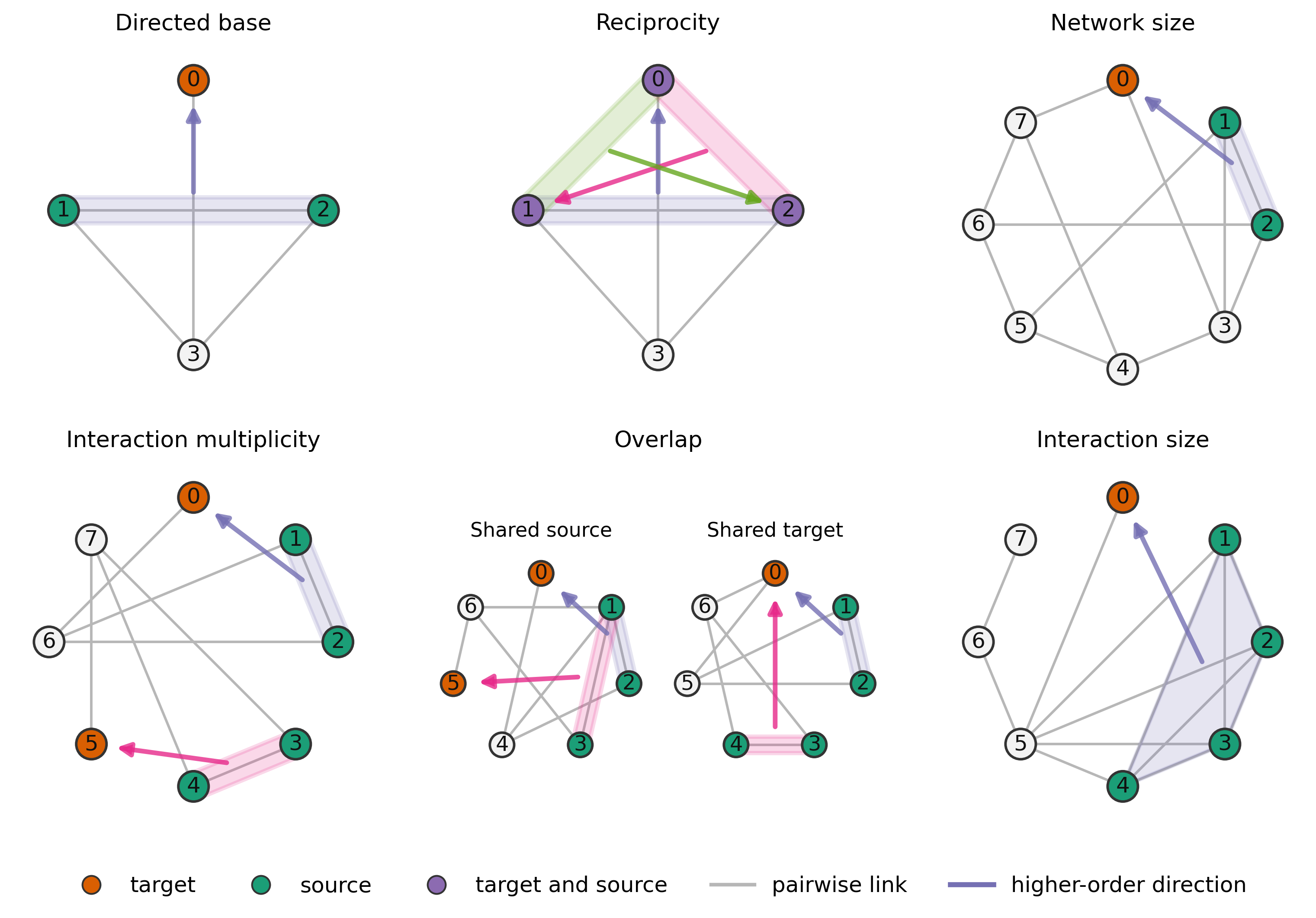}
  \caption{Topology regimes used in the generalisation experiments. Grey lines denote pairwise links, coloured bands connect higher-order sources, and arrows indicate their targets. The higher-order interactions and their coefficients are listed in Table~\ref{tab:topology-regimes}.}
  \label{fig:topology-regimes}
  \par\smallskip
{\small\raggedright
\textbf{Alt text:} Network diagrams compare the directed base case,
reciprocal interactions, a larger network, multiple interactions,
shared sources or targets, and an interaction with more than two sources
and one target.\par}
\end{figure}

Figure~\ref{fig:topology-regimes} shows the topology regimes used to test
whether detectability and recovery generalise beyond the four-node base case.
Across thirteen fixed deterministic layouts
(Table~\ref{tab:topology-regimes}), we vary network size over
$N\in\{4,8,16\}$, interaction multiplicity over $K\in\{1,2,4\}$ with shared or
separate targets, interaction overlap, and total interaction size over
$|S|+1\in\{3,4,5\}$.
Each layout embeds the pairwise motif of the base case around every planted
interaction and is spectrally normalised to $\alpha(A)=0.35\lambda$; all
remaining dynamical, stochastic, and estimator settings are inherited from
Section~\ref{sec:experimental-setup}. External-input allocation follows the planted target structure. In
non-reciprocal layouts, every node except the planted targets receives an
independent external input. Reciprocal layouts retain the base-case
allocation, in which node~0 is undriven and nodes~1, 2, and~3 receive
independent inputs. Comparisons between shared-target and separate-target
layouts therefore include the corresponding change in which nodes receive
external drive.
To approximate an undirected three-node hyperedge, the
reciprocal condition includes the three directed terms
$(1,2)\rightarrow0$, $(0,2)\rightarrow1$, and $(0,1)\rightarrow2$, each with
coefficient $\beta$. The normalised condition instead assigns
$\beta/\sqrt{3}$ to each term, giving the coefficient vector norm $\beta$.

Detection and recovery are calibrated separately for each layout against its own independently simulated $\beta=0$ ensemble, using the same calibration and test sizes as the base case. For interactions with more than two sources, mutual information remains
the maximum target--source pairwise reference, while O-information,
S-information, and Streitberg information are evaluated on the complete planted
node set. Because $\beta$ is not rescaled with source-set
size, the realised higher-order component is reported separately. For $K$ planted interactions, scores are retained separately
and recovery is evaluated using recall at rank $K$, mean average precision,
and the probability that all planted interactions occupy the first $K$ ranks.
For $K=1$, mean average precision reduces to mean reciprocal rank.
Directed MMI-PID and transfer-entropy recovery are restricted to three-node
interactions, so the size-four and size-five conditions contribute only to
detectability and unordered-set recovery.
The reciprocal configuration has one true unordered set,
$\{0,1,2\}$, but three true directed assignments; hence $K=3$ for directed
recovery.

\begin{table}[h]
  \centering
  \caption{Topology layouts used in the generalisation
  experiments. $N$ and $K$ denote the numbers of nodes and planted interactions,
  respectively. Interactions are written as an unordered source set followed
  by its target. Unless varied explicitly, all layouts inherit the base-case
  model settings.}
  \label{tab:topology-regimes}

  \begingroup
  \small
  \setlength{\tabcolsep}{4pt}
  \renewcommand{\arraystretch}{1.12}

  \begin{tabularx}{\linewidth}{
    @{}
    >{\raggedright\arraybackslash}p{0.22\linewidth}
    cc
    >{\raggedright\arraybackslash}X
    >{\centering\arraybackslash}p{0.11\linewidth}
    @{}
  }
    \toprule
    Layout & $N$ & $K$ & Planted directed interactions & Coefficient \\
    \midrule

    \multicolumn{5}{@{}l}{\textit{Base case and network size}}\\
    Directed base
      & 4 & 1
      & $(1,2)\!\to\!0$
      & $\beta$ \\
    Network size $N=8$
      & 8 & 1
      & $(1,2)\!\to\!0$
      & $\beta$ \\
    Network size $N=16$
      & 16 & 1
      & $(1,2)\!\to\!0$
      & $\beta$ \\

    \addlinespace
    \multicolumn{5}{@{}l}{\textit{Interaction multiplicity}}\\
    $K=2$, shared target
      & 16 & 2
      & $(1,2)\!\to\!0$, $(3,4)\!\to\!0$
      & $\beta$ \\
    $K=2$, disjoint
      & 16 & 2
      & $(1,2)\!\to\!0$, $(3,4)\!\to\!5$
      & $\beta$ \\
    $K=4$, shared target
      & 16 & 4
      & $(1,2)\!\to\!0$, $(3,4)\!\to\!0$,
        $(5,6)\!\to\!0$, $(7,8)\!\to\!0$
      & $\beta$ \\
    $K=4$, separate targets
      & 16 & 4
      & $(1,2)\!\to\!0$, $(3,4)\!\to\!5$,
        $(6,7)\!\to\!10$, $(8,9)\!\to\!11$
      & $\beta$ \\

    \addlinespace
    \multicolumn{5}{@{}l}{\textit{Overlap}}\\
    Shared source
      & 16 & 2
      & $(1,2)\!\to\!0$, $(1,3)\!\to\!5$
      & $\beta$ \\
    Shared source and target
      & 16 & 2
      & $(1,2)\!\to\!0$, $(1,3)\!\to\!0$
      & $\beta$ \\

    \addlinespace
    \multicolumn{5}{@{}l}{\textit{Interaction size}}\\
    Size four
      & 16 & 1
      & $(1,2,3)\!\to\!0$
      & $\beta$ \\
    Size five
      & 16 & 1
      & $(1,2,3,4)\!\to\!0$
      & $\beta$ \\

    \addlinespace
    \multicolumn{5}{@{}l}{\textit{Reciprocity}}\\
    Reciprocal cell
      & 4 & 3
      & $(1,2)\!\to\!0$, $(0,2)\!\to\!1$,
        $(0,1)\!\to\!2$
      & $\beta$ \\
    Reciprocal, normalised
      & 4 & 3
      & $(1,2)\!\to\!0$, $(0,2)\!\to\!1$,
        $(0,1)\!\to\!2$
      & $\beta/\sqrt{3}$ \\

    \bottomrule
  \end{tabularx}
  \endgroup
\end{table}
\FloatBarrier

Within each layout, the initial states, OU input paths, and
Wiener paths are reused across $\beta$. When a node appears in more than one
layout, it retains the same node-specific initial condition and stochastic
paths, allowing comparisons across topologies.

Reciprocal product interactions can produce positive nonlinear feedback.
Trajectories that fail the finite-time numerical-validity criterion are neither
clipped nor replaced. For each condition, we report the fraction of requested trajectories that remained valid and calculate detection and recovery scores using only those trajectories. These scores therefore describe performance conditional on numerical validity. For reciprocal layouts, we focus on $\beta=10$, where most trajectories remained valid, and report results at stronger coupling together with their validity counts.

\section{Results}
\label{sec:results}

This section reports what the information-theoretic measures recover from the
generative model. In the four-node base case, we first assess how often each measure detects changes induced by the higher-order term relative to the $\beta=0$ reference as coupling strength increases. We then test whether the participating nodes and their source-target roles can be identified when the interaction's identity is withheld. A further experiment tests whether the measures distinguish the
product coupling from a specified nonlinear pairwise control
scaled to have the same RMS as the product. We then ask how far the detection and
recovery results extend, applying the same measures and null-calibration
procedure across the topology regimes of
Section~\ref{sec:methods-topology-generalisation}. All detection and recovery scores reported below are calibrated against an independently simulated ensemble with
the higher-order term removed, so these results are always expressed
relative to what the same measure produces on the same system without a
higher-order interaction. Discrimination scores instead use the nonlinear pairwise
control at the corresponding coupling strength as their calibration reference.

\subsection{Base-case benchmark}
\label{sec:results-base-case}

The four-node base case separates validation of the generative model from
structural inference. Detection evaluates the known planted triplet to confirm
that activating the higher-order term produces a measurable effect relative to
the pairwise-only condition. Recovery then withholds this structural
information and tests whether the measures can identify the participating
nodes and their source--target roles, more closely reflecting their use with
empirical data. Discrimination then tests whether the statistical signatures
of the product coupling can be distinguished from those of a specified
decomposable nonlinear pairwise alternative.

\subsubsection{Higher-order interaction detectability}
\label{sec:results-base-detectability}

We varied the higher-order coupling strength over
$\beta\in\{0,0.2,0.5,1,2,5,10,20,30\}$. For each metric, the detection
threshold was estimated from 200 independent $\beta=0$ calibration simulations.
C-sensitivity was then evaluated across 40 test replicates; within each
replicate, the initial state, external Ornstein--Uhlenbeck input paths, and
Wiener process noise paths were held fixed across $\beta$.

The main result was that the planted higher-order interaction was detectable
and became more readily distinguishable from the pairwise-only null as
$\beta$ increased. This shows that the model produces a controllable
higher-order effect, detected most clearly by the non-parametric estimators. Detection remained close to its
independent $\beta=0$ level for $\beta\leq1$ and increased consistently from
$\beta=2$ onwards. At $\beta=10$, KSG mutual information and KSG
S-information detected the planted interaction in $39/40$ and $33/40$
replicates, respectively. At $\beta=20$, both detected all $40$ replicates
(c-sensitivity $1.000$; $95\%$ Wilson CI $[0.912,1.000]$). KSG
O-information and Streitberg information showed more moderate responses,
reaching c-sensitivities of $0.575$ and $0.625$ at $\beta=20$, whereas the
Gaussian estimators remained at or below $0.275$ throughout the sweep
(Figure~\ref{fig:base-detectability}). Exact metric-by-$\beta$ values are
provided in Appendix Figure~\ref{fig:app-csensitivity-heatmap}.

\begin{figure}[!htbp]
    \centering
    \includegraphics[width=0.79\linewidth]
        {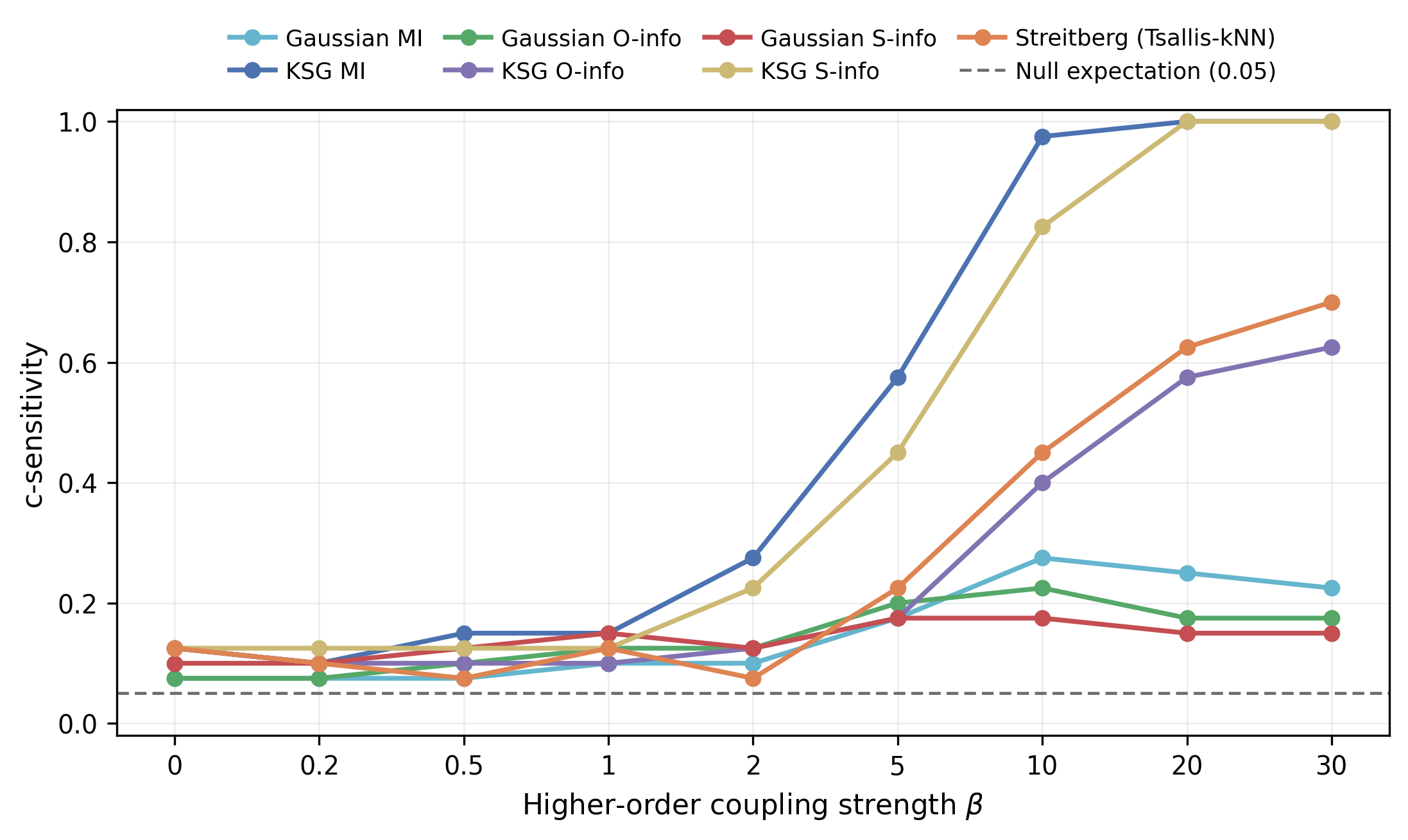}
    \caption{
Detection of the higher-order interaction as $\beta$ increases.
C-sensitivity is the fraction of 40 test replicates exceeding
metric-specific thresholds calibrated using 200 independent $\beta=0$
simulations. The dashed line marks the $5\%$ false-positive rate targeted by the calibration threshold.
}
    \label{fig:base-detectability}
    \par\smallskip
{\small\raggedright
\textbf{Alt text:} Detection rates versus coupling strength.
KSG mutual information and S-information rise to full detection;
KSG O-information and Streitberg information increase more moderately,
while Gaussian estimates remain low.\par}
\end{figure}
\FloatBarrier

Because the $\beta=0$ test replicates contain no higher-order term, any
detection among them is a false positive, and the calibration threshold is
designed to keep this rate at $0.05$. The observed rate was higher, at
$0.125$ ($5/40$; 95\% Wilson CI $[0.055,0.261]$) for the KSG and Streitberg
estimators and $0.075$--$0.100$ for the Gaussian estimators. This elevation did not recur in the independently simulated $N=4$ topology ensemble, where the corresponding rates were $0.000$-$0.075$. Both the calibration and test ensembles differed between these experiments, so this comparison does not isolate the source of the difference in false-positive rates.
Detection rates at larger $\beta$ should therefore be read relative to this
observed baseline rather than the nominal $0.05$.

The pairwise KSG mutual-information baseline responded most strongly, showing that changes induced by the higher-order term were detectable using a score computed from target-source pairs. This result provides a sensitivity benchmark, but does not establish
irreducibly multivariate dependence. Among the measures evaluated on the full triplet, KSG S-information was the
most sensitive: it separated from the null earlier than O-information and
Streitberg information and detected the interaction in all 40 replicates at
$\beta=20$. Because pairwise dynamics can also produce non-zero
S-information, the evidence for the planted mechanism comes from this
null-calibrated increase rather than from the S-information value alone.

KSG O-information had lower detection rates than KSG S-information. Mean KSG O-information decreased from $-0.631$ at $\beta=0$ to $-0.753$ at $\beta=20$ and became more negative in $38/40$ corresponding replicates (Appendix Table~\ref{tab:app-oinformation}). These results describe changes in estimated O-information relative to the pairwise-only reference. Interpreting their signs and magnitudes as population properties requires accounting for estimation bias, including the sampling effects examined in Appendix~\ref{app:discrimination-validation}. Streitberg information also
departed from the null, but neither this response nor the O-information change
rules out a lower-order statistical explanation.

The KSG estimators had higher detection rates than their Gaussian counterparts under the sampling scheme used. For jointly zero-mean Gaussian sources, $\operatorname{Cov}(z_j,z_jz_k) = \mathbb{E}[z_j^2z_k]=0$, including when the sources are correlated. Differences in estimation bias and variability may also contribute to the observed differences in detection rates.

\begin{figure}[!htbp]
    \centering
    \includegraphics[width=0.78\linewidth]
        {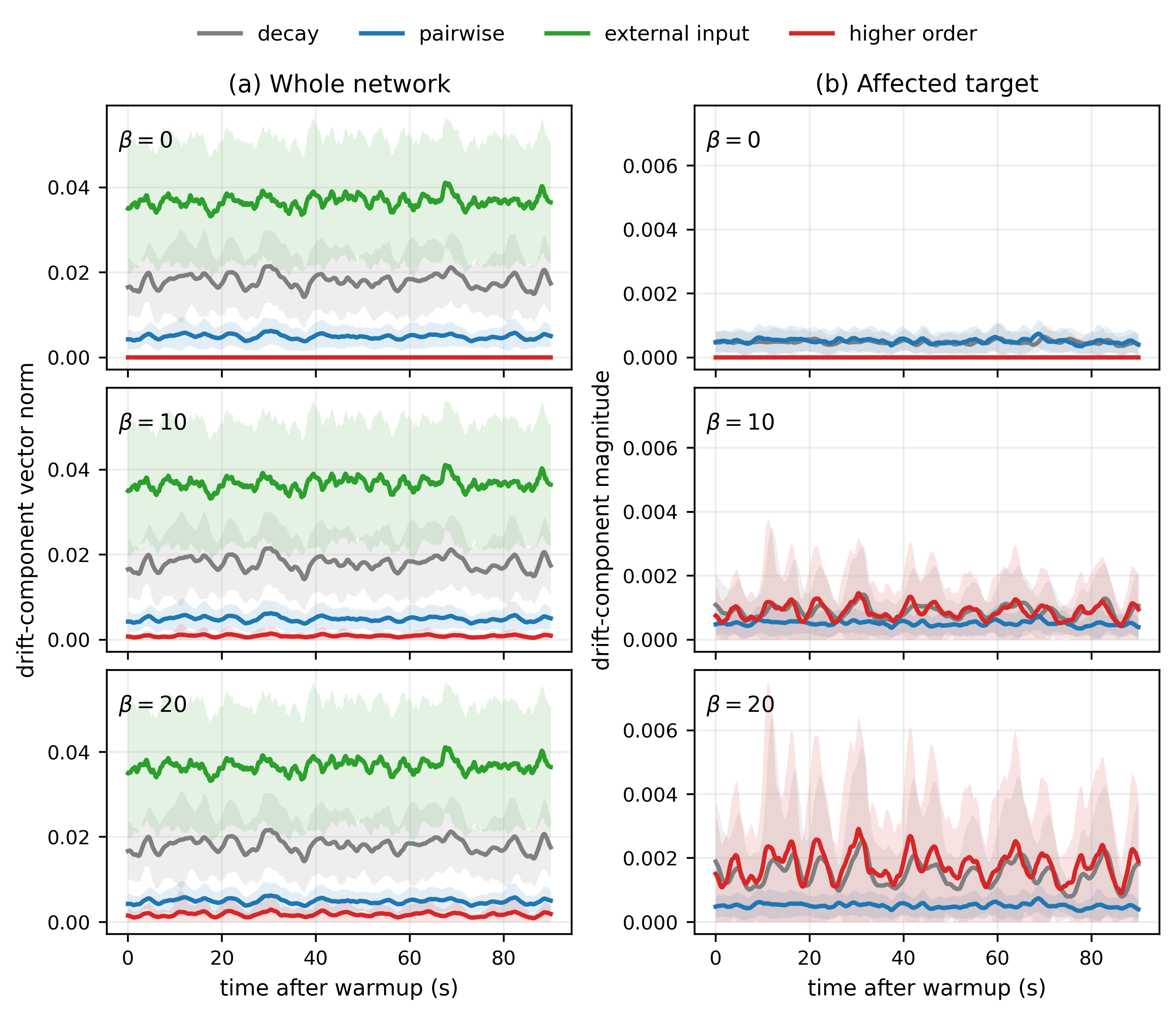}
    \caption{
Drift-component magnitudes across the network (a) and at the planted target
(b) for $\beta=0$, $10$, and $20$. Lines and shading show the mean
$\pm1$ standard deviation across 20 replicates, smoothed over
two seconds. The target receives no direct input, so panel~(b) omits the
external-input curve. The panels use different vertical scales; process noise
is excluded because it is not a drift term.
}
    \label{fig:component-norms}
    \par\smallskip
{\small\raggedright
\textbf{Alt text:} Drift magnitudes fluctuate without sustained growth.
External input dominates across the network. At the target,
higher-order and decay contributions increase with coupling strength,
while the pairwise contribution remains comparatively small.\par}
\end{figure}

Finally, the detection pattern occurred in bounded, sustained dynamics. All 40 replicates remained numerically valid at each tested coupling strength,
with no trajectories approaching the numerical-validity threshold. At $\beta=10$ and $\beta=20$, the ensemble-mean higher-order contribution
remained sustained throughout the post-warm-up interval. Its magnitude
increased from $\beta=10$ to $\beta=20$, as expected for stronger coupling,
but did not continue to grow over time within either condition
(Figure~\ref{fig:component-norms}). This temporal stability indicates a
persistent coupling effect rather than runaway dynamics. At $\beta=20$, its whole-network norm was
$0.36$ times the pairwise norm, whereas at the affected target its mean RMS
magnitude was $4.61$ times the pairwise contribution. This difference arises
because the higher-order term acts only on node~0, while pairwise contributions
act throughout the network.

Although process noise is excluded from the drift-component figure, its scale
remains relevant when interpreting the operating regime. Across the detection
replicates at $\beta=20$, the median target-node process-noise increment RMS
was $6.00\times10^{-5}$ per integration step, comparable to the median
higher-order increment RMS of $5.74\times10^{-5}$. The planted interaction is
therefore detectable despite operating in the presence of stochastic
fluctuations of comparable instantaneous scale.

Appendix Figure~\ref{fig:base-node-timeseries} shows the node trajectories
for one replicate. Because both conditions share the same initial
state, inputs, and noise, the non-target trajectories largely overlap, and
the clearest difference is the larger excursions of the target, node~0, where
the higher-order term enters directly.

\subsubsection{Recovery of the planted triplet}
\label{sec:results-base-recovery}

The detectability results showed that, when the planted triplet was supplied,
the higher-order term produced a detectable departure from the $\beta=0$
reference. Having validated that the model produces a detectable higher-order effect, we next considered a task closer to empirical
analysis. We withheld the interaction's identity and asked whether the measures
could recover its participating nodes and source--target orientation from all
possible candidates.

\begin{figure}[H]
    \centering
    \includegraphics[width=0.8\linewidth]
        {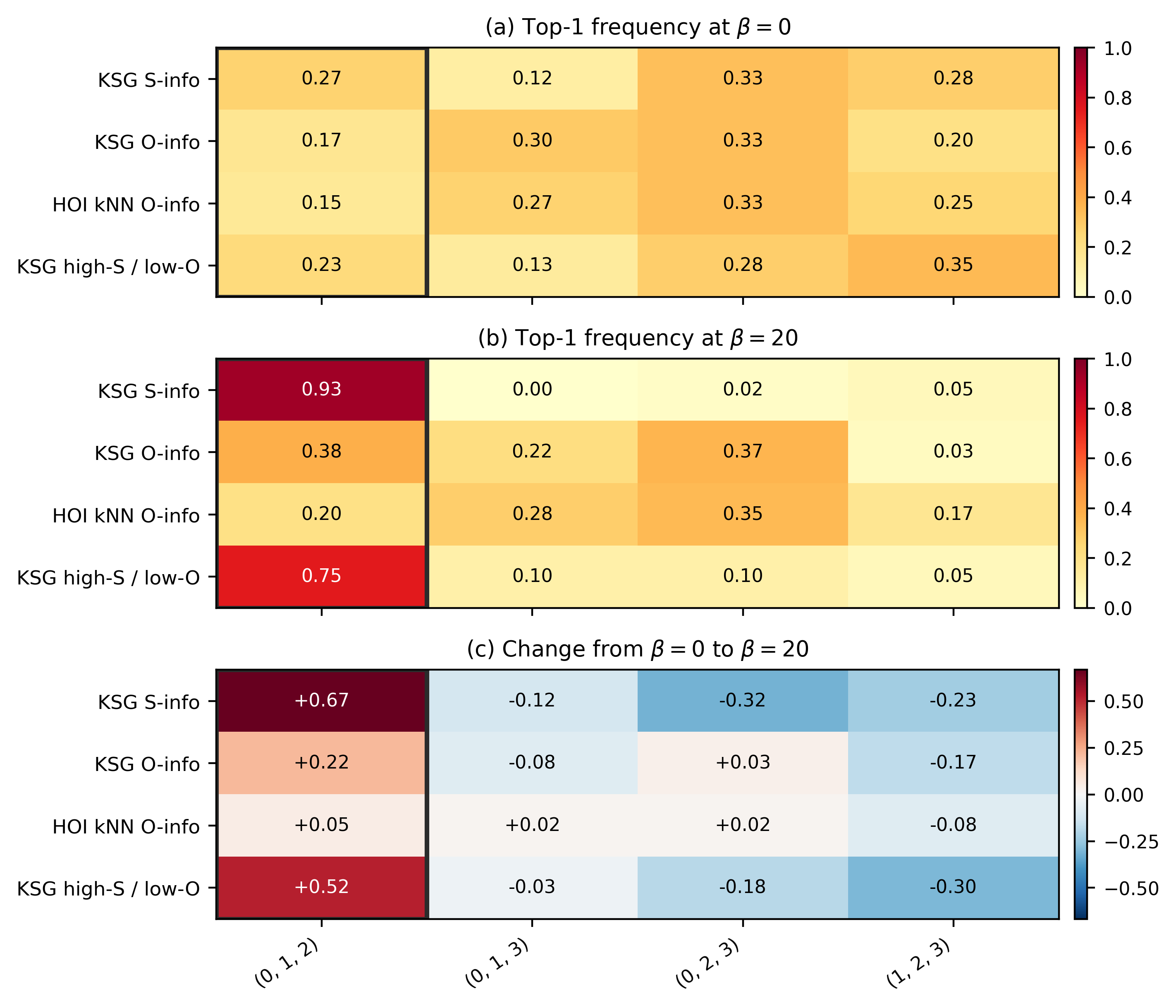}
    \caption{
Unordered-set recovery across 60 replicates. Cells show top-1
selection frequencies at $\beta=0$ (a), $\beta=20$ (b), and their difference
(c). The black outline marks the planted set $\{0,1,2\}$.
}
    \label{fig:base-unordered-recovery}
    \par\smallskip
{\small\raggedright
\textbf{Alt text:} Three heatmaps compare triplet recovery without
and with higher-order coupling and the resulting changes.
KSG S-information recovers the planted triplet most strongly,
followed by the combined high-S, low-O score.\par}
\end{figure}

Recovery was evaluated at $\beta=20$, the fixed operating point specified in
Section~\ref{sec:triplet-recovery}. Candidate-specific null distributions were estimated from
100 independent $\beta=0$ calibration simulations, followed by
60 test replicates, each using the same seeds for the
$\beta=0$ and $\beta=20$ conditions. Symmetric
measures ranked the four admissible unordered three-node sets, whereas
target-specific measures ranked all 12 directed source-pair--target
assignments. Each candidate was calibrated against its own null distribution to
account for differences in baseline dependence across the network.

The principal result was that the participating nodes could be recovered more
reliably than their source--target orientation. In the $\beta=0$ control, selection of the planted unordered set
remained close to the chance probability of $1/4$. At $\beta=20$, KSG
S-information ranked the planted set $\{0,1,2\}$ first in $0.933$ of
replicates ($95\%$ CI $[0.841,0.974]$). The combined high-S/low-O score reached
$0.750$, while neither O-information estimator exceeded $0.383$
(Figure~\ref{fig:base-unordered-recovery}). KSG S-information therefore
provided the clearest identification of the participating nodes. Adding
O-information to the combined score reduced performance because
O-information was less selective among the candidate sets.

\begin{figure}[H]
    \centering
    \includegraphics[width=0.85\linewidth]
        {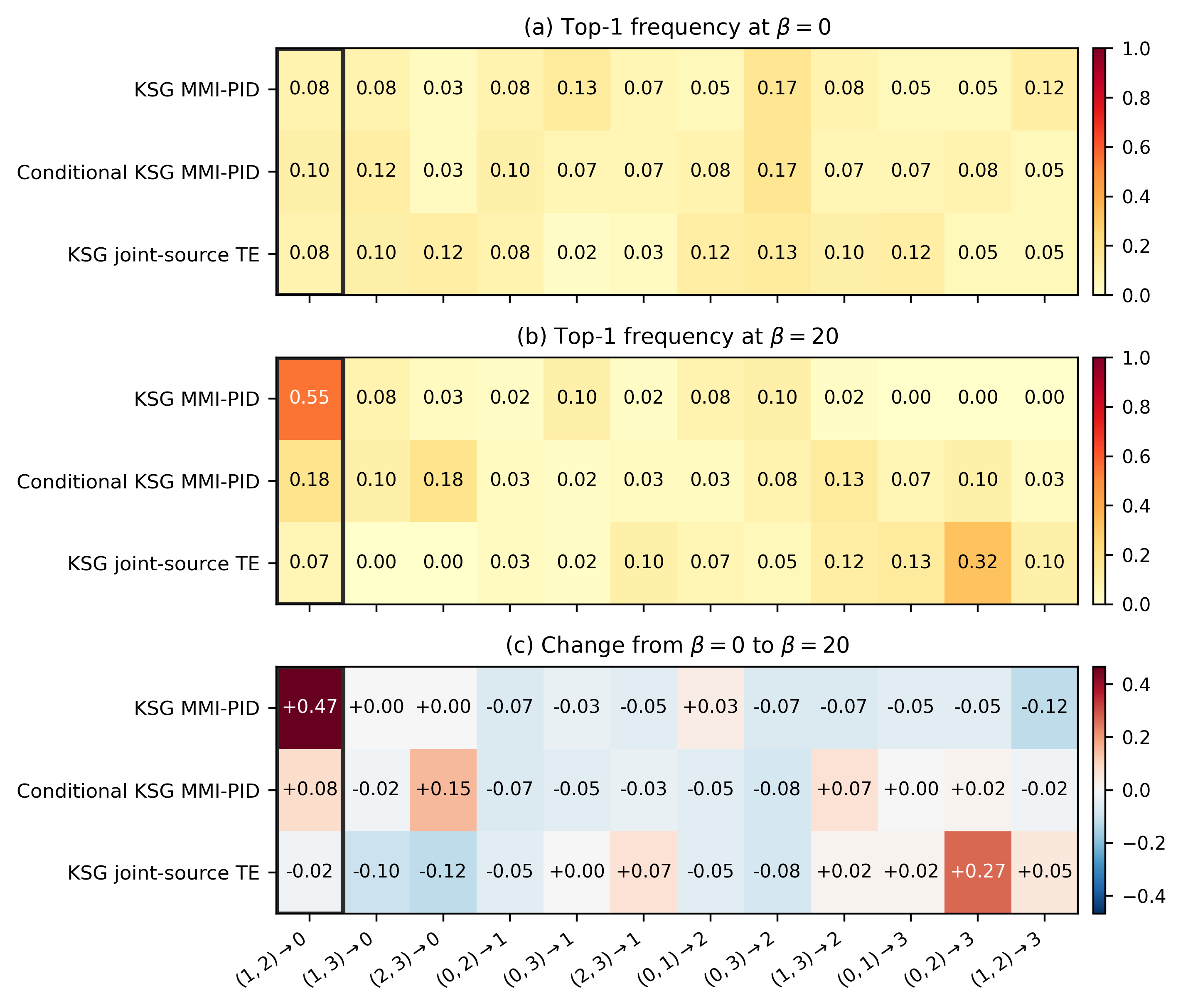}
    \caption{
Top-ranked directed assignments at $\beta=0$ (a), $\beta=20$ (b), and their
difference (c) across 60 replicates. Cells show selection frequencies;
the black outline marks the planted interaction, with nodes 1 and 2 as sources
and node 0 as the target.
}
    \label{fig:base-directed-recovery}
    \par\smallskip
{\small\raggedright
\textbf{Alt text:} Three heatmaps compare directed-assignment recovery
without and with higher-order coupling and the resulting changes.
Unconditional KSG MMI-PID recovers the planted interaction most strongly;
joint-source transfer entropy recovers an incorrect target instead.\par}
\end{figure}

Recovering the directed interaction was more difficult. Unconditional KSG
MMI-PID ranked the exact assignment $(1,2)\rightarrow0$ first in $0.550$ of
replicates ($95\%$ CI $[0.425,0.669]$), compared with the chance-level
frequency of $0.083$ under the $\beta=0$ control (Figure~\ref{fig:base-directed-recovery}). It recovered the
correct source pair in $0.767$ of replicates when target~0 was supplied, but
selected the correct target in $0.667$. Target identification was therefore
the main limitation on exact recovery.

Conditioning MMI-PID on the target's current state reduced exact recovery to
$0.183$. One possible explanation is that the short prediction horizon leaves little additional source information after conditioning on the target's present state. Conditioning also increases the dimension of the estimation problem, which may affect recovery.

Joint-source transfer entropy showed a sharper contrast. It recovered the
correct source pair in $0.833$ of replicates when the target was supplied, but
its exact recovery rate in the blind search was only $0.067$. It frequently
favoured $(0,2)\rightarrow3$, whose proposed sources both connect directly to
node~3 in the pairwise network. Joint-source transfer entropy measures the
total predictive information supplied by the source pair and therefore does
not isolate information that requires both sources jointly. Unconditional
MMI-PID was consequently more selective for the planted directed interaction.

The detection and recovery experiments serve different purposes in the
benchmark. Detection evaluates the known planted triplet to confirm that
activating the higher-order term produces an observable statistical signature
and to compare estimator sensitivity. Recovery withholds the planted identity,
more closely reflecting applications in which the interacting nodes and their
roles are unknown. In the base case, KSG S-information recovered the unordered
participating set reliably, suggesting that it can help identify candidate
groups. Recovering the exact source--target assignment was far less reliable. This distinction illustrates the importance of generative benchmarks that test
both whether higher-order effects are detectable and how reliably their
structure can be recovered.

\subsubsection{Sensitivity to the coupling function}
\label{sec:results-coupling-sensitivity}

We next assessed to what extent the base-case results depended on the
bilinear product coupling. The product was retained as the reference and
compared with the sourcewise tanh, sine, and smooth signed-root couplings
defined in Section~\ref{sec:methods-coupling}. Their calibration and
comparison are described in
Section~\ref{sec:methods-coupling-sensitivity}. The product remains the
coupling used in all other experiments unless otherwise stated.

The purpose of this experiment was to test whether detection generalised
beyond the product coupling $z_jz_k$ while controlling for interaction
strength. We replaced the product with sourcewise tanh, sine, and smooth-root
couplings. Here, ``sourcewise'' means that each source was transformed
separately before the transformed values were multiplied:
$\phi_m(z_j,z_k)=a_m g_m(z_j/s_j)g_m(z_k/s_k)$. This preserves an explicit
dependence on both sources while changing how their individual values
contribute to the interaction. Each alternative was calibrated on source
states from independent $\beta=0$ simulations to have the same output RMS as
the product.

\begin{figure}[!t]
  \centering
  \includegraphics[width=\linewidth]
    {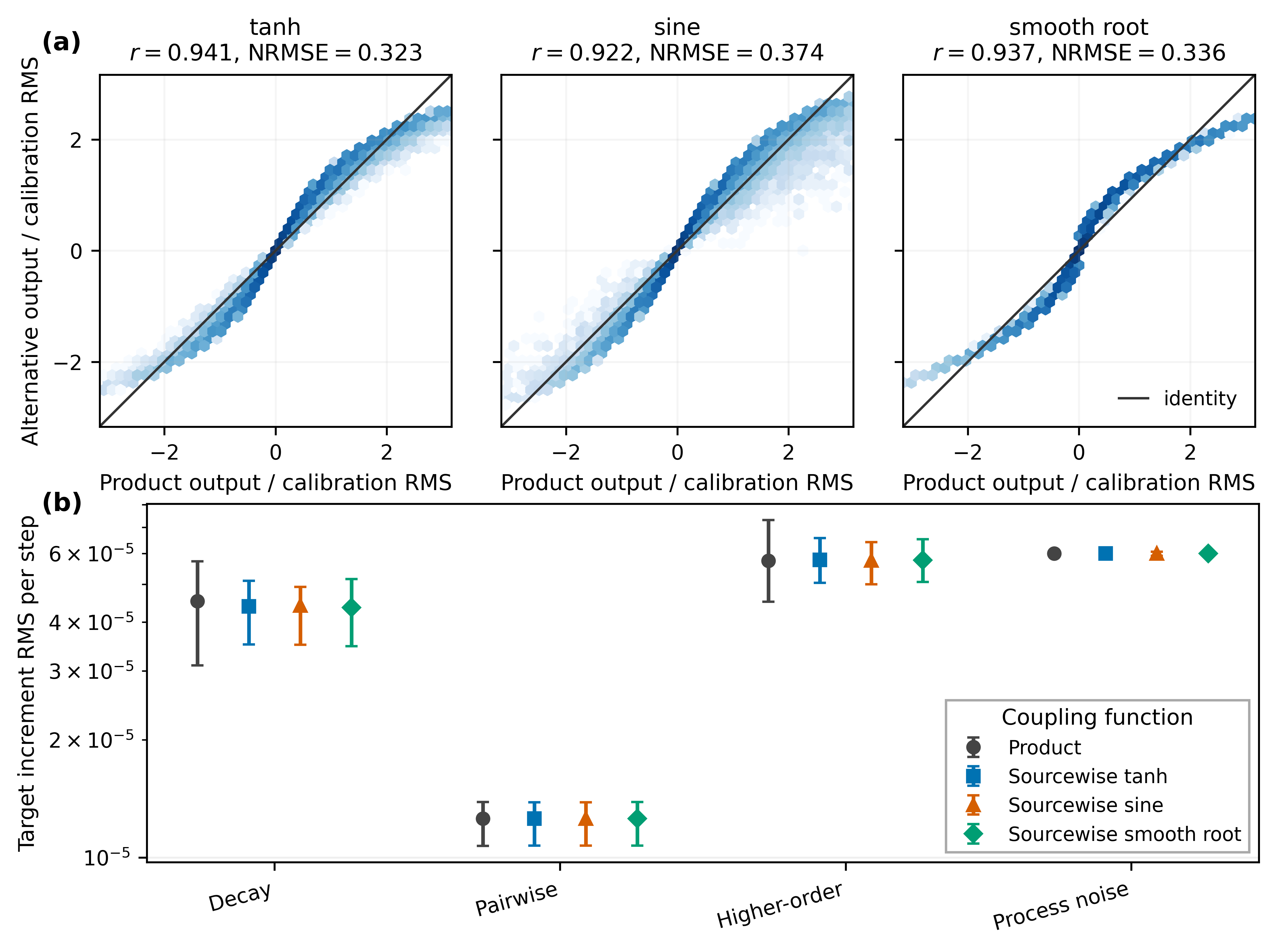}
  \caption{
Calibration and realised coupling scales.
(a) Alternative outputs versus the product for the same source states from
200 independent $\beta=0$ simulations, scaled to a common RMS. Darker
hexagons indicate greater density, and the black line marks equal outputs.
Titles report correlation $r$ and normalised error (NRMSE) relative to the
product, with higher $r$ and lower NRMSE indicating closer agreement. (b) Target-node component RMS per step at $\beta=20$. Points show medians and
bars show the 10th--90th percentiles across 40 replicates. The axis is
logarithmic. No input component is shown because the target receives no
direct input.
}
  \label{fig:coupling-calibration-scale}
  \par\smallskip
{\small\raggedright
\textbf{Alt text:} (a) Alternative coupling outputs closely track
the product but diverge at larger magnitudes.
(b) Target-component magnitudes are similar across coupling functions,
with higher-order and process-noise increments exceeding pairwise
increments.\par}
\end{figure}

The calibrated outputs were related but not equivalent
(Figure~\ref{fig:coupling-calibration-scale}a). All three transformations are
approximately linear near the origin, so their outputs initially follow the
product. At larger source magnitudes, tanh saturates, sine moves beyond its
near-linear region, and the smooth root compresses extreme values. These
effects produce the visible curvature away from the identity line. The
vertical dispersion occurs because the same product $z_jz_k$ can arise from
different individual values of $z_j$ and $z_k$, which the sourcewise
transformations treat differently. Correlations with the product ranged from
$0.922$ to $0.941$, with NRMSE values of $0.323$--$0.374$. The alternatives
therefore retained broad similarity to the product while implementing
meaningfully different coupling functions.

\FloatBarrier
Figure~\ref{fig:coupling-calibration-scale}b confirms that the alternative
couplings affected the target at nearly the same strength once embedded in
the dynamics. At $\beta=20$, their median higher-order increment RMS ranged
from $5.74\times10^{-5}$ to $5.77\times10^{-5}$ per step, a difference of
less than $1\%$. Thus, no coupling function produced a stronger
planted signal than the others. These higher-order increments were also
similar in magnitude to the process-noise increment RMS of
$6.00\times10^{-5}$, so they did not simply overwhelm the stochastic
fluctuations. Differences in detector performance can therefore be
interpreted primarily as sensitivity to the coupling function rather than to
unequal realised signal strengths.

\begin{figure}[H]
    \centering
    \includegraphics[width=0.84\textwidth]
    {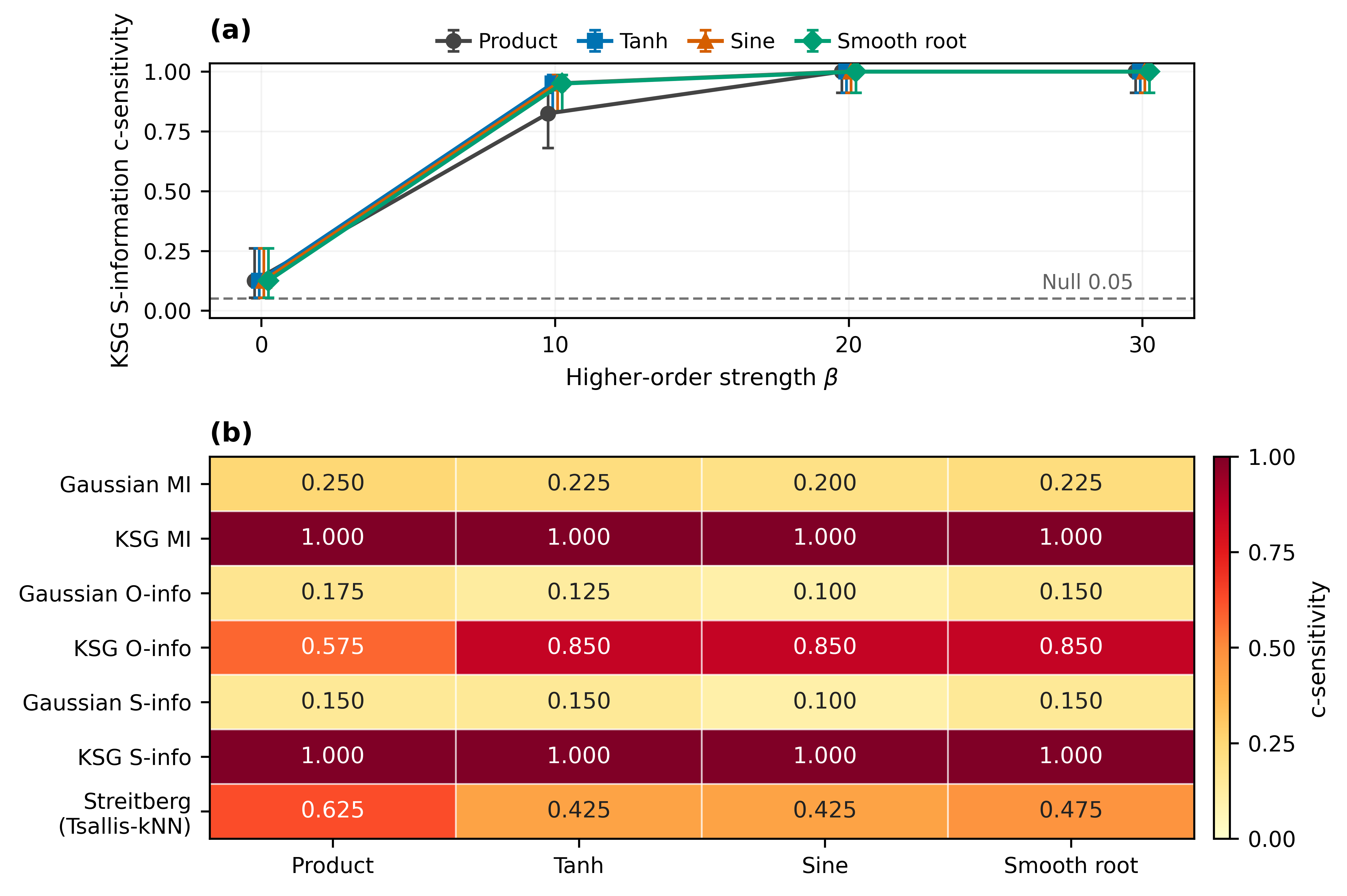}
    \caption{
    Detection across coupling functions. (a) KSG S-information
    c-sensitivity across $\beta$ in 40 replicates, with 95\% Wilson
    intervals and a $5\%$ false-positive rate targeted by the calibration threshold. (b) Detector c-sensitivity scores at
    $\beta=20$, using thresholds from 200 independent $\beta=0$ simulations.
    }
    \label{fig:coupling-detection}
    \par\smallskip
{\small\raggedright
\textbf{Alt text:} (a) KSG S-information reaches full detection
as coupling increases for all four functions.
(b) KSG mutual information and S-information detect all replicates;
other detectors show lower, sometimes function-dependent sensitivity.\par}
\end{figure}

The main result was the consistency of KSG S-information across coupling
functions (Figure~\ref{fig:coupling-detection}a). All four mechanisms showed
a strong increase in c-sensitivity as $\beta$ increased. At $\beta=10$,
c-sensitivity reached $0.825$ for the product and $0.950$ for each sourcewise
alternative. At $\beta=20$ and $\beta=30$, all 40 replicates exceeded the KSG
S-information null threshold for the product, tanh, sine, and smooth-root
couplings. At $\beta=0$, all functions shared the same c-sensitivity of
$0.125$ because the higher-order term was absent and the trajectories
were identical. KSG S-information therefore showed the same overall
transition from the pairwise-only reference under every coupling function.

Several similarities also appeared across the full detector suite at
$\beta=20$ (Figure~\ref{fig:coupling-detection}b). The pairwise KSG
mutual-information baseline and KSG S-information both reached a
c-sensitivity of $1.000$ for all four functions. The Gaussian estimators were
also broadly consistent across functions, but remained weak. Gaussian mutual
information ranged from $0.200$ to $0.250$, Gaussian O-information from
$0.100$ to $0.175$, and Gaussian S-information from $0.100$ to $0.150$.
Within these narrow ranges, the product generally gave the highest values and
sine the lowest.

The clearest differences occurred for KSG O-information and Streitberg
information. KSG O-information increased from $0.575$ for the product to
$0.850$ for each sourcewise alternative. Streitberg information showed the
opposite pattern. Its c-sensitivity was $0.625$ for the product, $0.425$ for
both tanh and sine, and $0.475$ for the smooth root. Thus, tanh and sine
produced identical results for these two measures, while the smooth root
achieved the same KSG O-information performance
but gave slightly stronger Streitberg detection.

Because Figure~\ref{fig:coupling-calibration-scale} established that the four
mechanisms had comparable realised strengths, these differences are best
interpreted as sensitivity to coupling form rather than signal magnitude. The
main group-level result was robust, since KSG S-information detected all four
mechanisms once the coupling was sufficiently strong. However, the relative
performance of KSG O-information and Streitberg information remained
mechanism-dependent. We retain the product in the remaining experiments
because it provides the simplest interpretable coupling, while this analysis
shows that the principal detection result extends to all three sourcewise
alternatives.

All trajectories remained bounded through $\beta=30$. At $\beta=20$,
refining the timestep from $\Delta t=0.02$ to $0.005$ produced median
post-warm-up target-state errors of $2.84$--$2.94\%$ across the four
mechanisms, providing no indication that numerical instability drove the
observed similarities or differences.

\subsubsection{Discriminating non-decomposable and decomposable couplings}
\label{sec:results-discrimination}

\begin{figure}[H]
    \centering
    \includegraphics[width=0.8\linewidth]
        {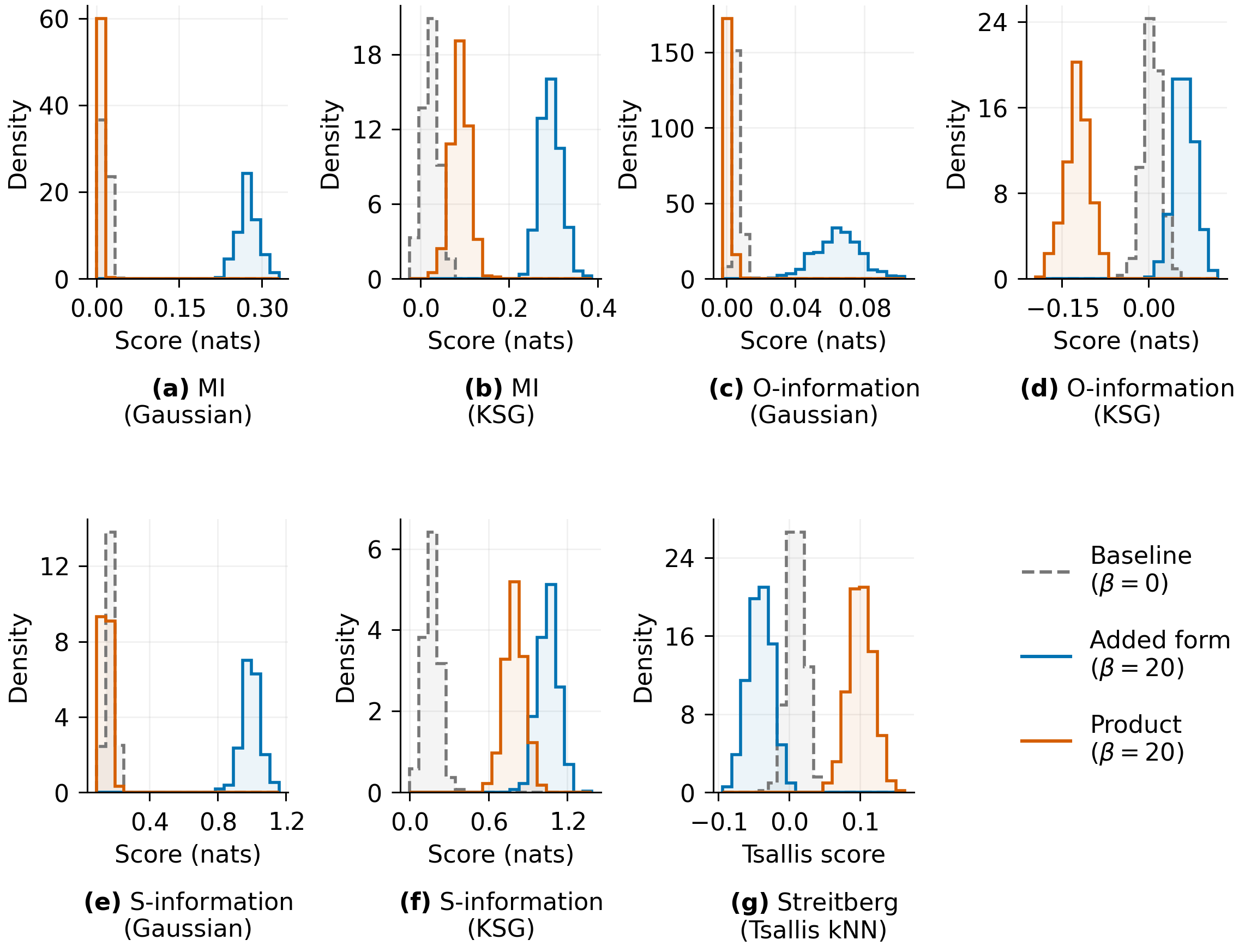}
    \caption{Score distributions for the baseline ($\beta=0$), additive
    control and product coupling ($\beta=20$), shown separately
    for each measure. Each distribution contains 400 held-out
    simulations from the independent validation ensemble.
    Histograms use shared bins within each panel and are
    individually normalised to unit area.
    }
    \label{fig:discrimination-scores}
    \par\smallskip
{\small\raggedright
\textbf{Alt text:} Seven histograms compare baseline, additive-control
and product-coupling scores. Product and additive distributions
are clearly separated for most measures, with greater overlap
for KSG S-information.\par}
\end{figure}

To assess discrimination, we tested whether the measures could
distinguish the product coupling from a specified additive alternative
after scaling the additive output to have the same RMS as the
product on the $\beta=0$ scaling ensemble. Figure~\ref{fig:discrimination-scores} compares the score
distributions for the baseline ($\beta=0$), additive control
and product coupling (both at $\beta=20$).
At $\beta=20$, most measures showed little overlap between the
product and additive distributions, whereas KSG S-information
showed greater overlap. Because calibration is two-sided, either
higher or lower product scores can provide discrimination.

In the independent validation at $\beta=20$, using 1000
calibration simulations, KSG S-information rejected the additive
reference for 356 of 400 product simulations ($89\%$), with a
$4\%$ false-positive rate on additive controls.
The other six measures rejected the reference for all 400
product simulations, with false-positive rates between
$3.25\%$ and $6\%$.

\begin{figure}[H]
    \centering
    \includegraphics[width=0.75\linewidth]
        {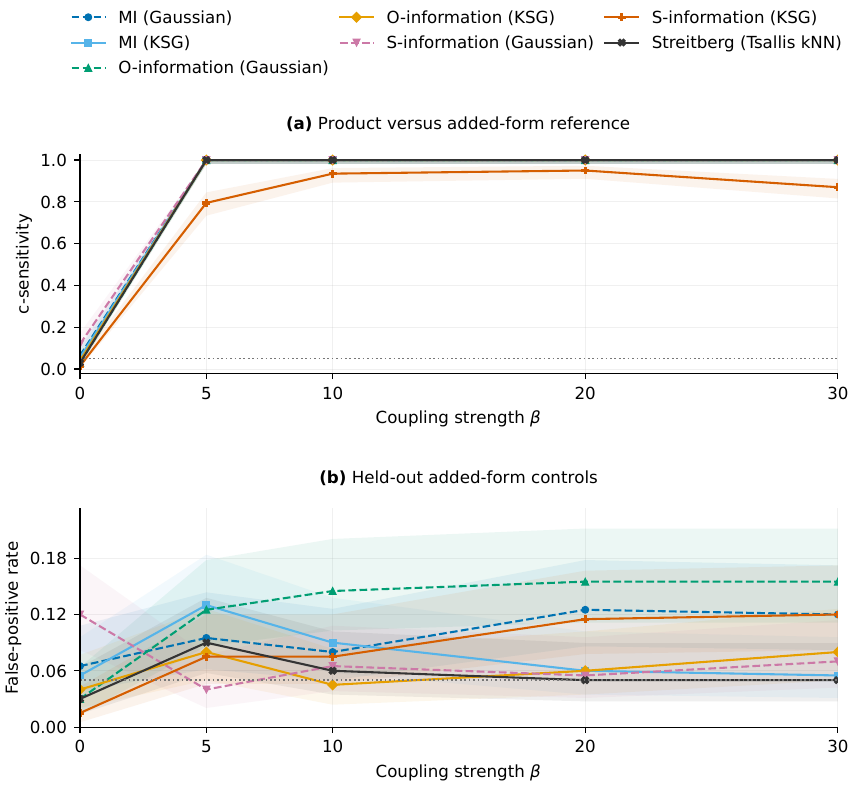}
    \caption{Discrimination across coupling strengths.
\textbf{(a)} Discrimination c-sensitivity: the fraction of product
simulations whose scores fall outside the calibrated additive-reference
interval, using the nonlinear pairwise control rather than the
$\beta=0$ baseline as the reference.
\textbf{(b)} Fraction of additive-control simulations falling
outside the same interval (false-positive rate).
At each $\beta$, thresholds were calibrated using 200 independent
additive simulations and evaluated on 200 held-out simulations
per coupling form. Shading shows pointwise Wilson 95\% intervals
conditional on the fitted thresholds; dotted lines mark the
nominal $5\%$ level. At $\beta=0$, the coupling conditions reduce
to the same baseline.
}
    \label{fig:discrimination-csensitivity}
    \par\smallskip
{\small\raggedright
\textbf{Alt text:} (a) Discrimination is high across all measures
at nonzero coupling strengths, ranging from about eighty to
one hundred percent. (b) Additive-control false-positive rates
vary across measures, often exceeding the nominal five percent.\par}
\end{figure}

The separate across-$\beta$ experiment also showed substantial
separation (Figure~\ref{fig:discrimination-csensitivity}a).
At every tested nonzero strength, all measures except KSG
S-information rejected the additive reference for all 200
product simulations. KSG S-information achieved rejection
rates between $79.5\%$ and $95\%$.
However, additive-control rejection rates ranged from $4\%$
to $15.5\%$ at these strengths
(Figure~\ref{fig:discrimination-csensitivity}b), exceeding the
nominal $5\%$ level for some measures.
Some measures incorrectly flagged additive simulations more often
than others, so their high discrimination rates must be considered
alongside their false-positive rates. In the larger independent
validation at $\beta=20$, discrimination remained strong while
false-positive rates were close to the nominal $5\%$ level.
This larger validation was not repeated at the other coupling
strengths.

Taken together, the results show substantial discrimination between
the product and signed-square additive couplings across all seven
measures, already at the smallest tested nonzero strength,
$\beta=5$. These results must be considered alongside the elevated
false-positive rates for some measures in the across-$\beta$
experiment. The larger independent validation provides stronger
evidence at $\beta=20$. 

These findings complement detection and recovery by addressing
whether higher-order mechanisms can be distinguished from what
pairwise dynamics could also produce. Here, the measures separated
the planted product from a specified nonlinear pairwise alternative
with the same output RMS on the
$\beta=0$ scaling ensemble. This provides evidence
that their discrimination extends beyond detecting a change relative
to $\beta=0$.
The comparison demonstrates distinguishable statistical
signatures of the two tested couplings, but does not establish that
a measure uniquely identifies non-decomposability
or distinguishes it from every possible pairwise alternative.

\subsection{Generalisation across topology regimes}
\label{sec:results-topologies}

We next tested whether the base-case findings generalised as network and
interaction structure became more complex. We varied network size,
interaction multiplicity and arrangement, interaction size, and reciprocity
according to Table~\ref{tab:topology-regimes}. Each layout was calibrated
against its own independently simulated $\beta=0$ null ensemble. Shared-target and separate-target layouts also differ in which nodes receive input, as described in Section~\ref{sec:methods-topology-generalisation}. Their comparison therefore assesses the combined effect of these design changes.

Detection was evaluated across the full coupling sweep, while recovery was
evaluated at $\beta=20$ for the non-reciprocal layouts. These layouts retained
every trajectory. In the reciprocal layouts, numerical validity decreased as
$\beta$ increased. Their high-$\beta$ scores therefore apply only to the
trajectories that remained valid and may not represent the full set of runs.
We consequently treat $\beta=10$, where most reciprocal trajectories remained
valid, as the primary reciprocal result and report higher-$\beta$ scores
alongside their validity counts.

\subsubsection{Higher-order interaction detectability across topologies}
\label{sec:results-topology-detectability}

\begin{figure*}[!b]
  \centering
  \includegraphics[width=0.92\textwidth]
    {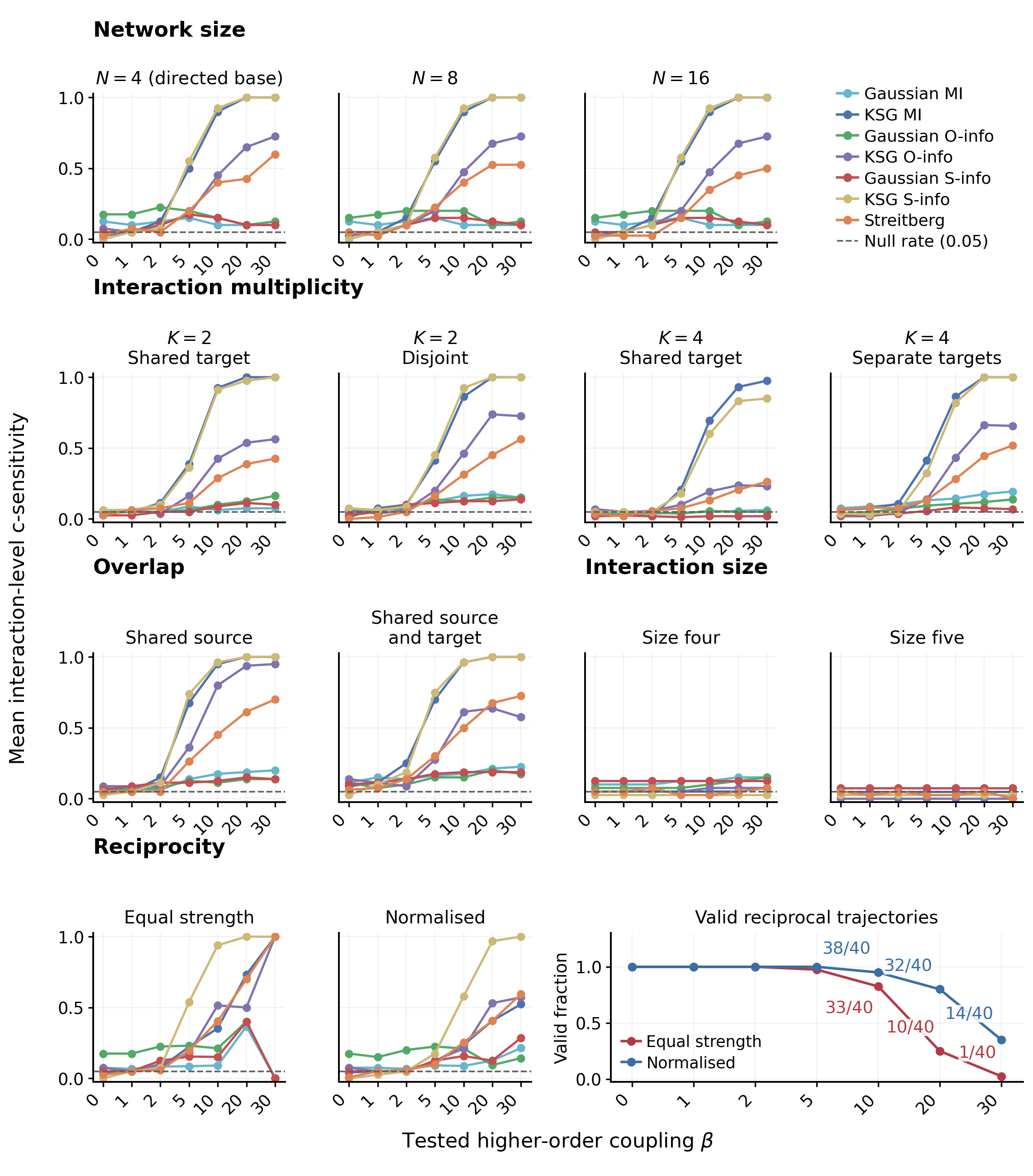}
  \caption{
Null-calibrated detection across topology regimes. Curves show
interaction-level c-sensitivity across $\beta$; for layouts with multiple
planted interactions, scores were calculated separately and then averaged.
The dashed line marks the $5\%$ false-positive rate targeted by the calibration threshold. All trajectories were valid except
in the reciprocal layouts, whose detection scores are conditional on
numerical validity. The final panel reports the corresponding valid/40 counts
at $\beta=10$, $20$, and $30$.
}
  \label{fig:topology-detectability-curves}
  \par\smallskip
{\small\raggedright
\textbf{Alt text:} Detection curves remain similar across network
sizes but vary with interaction multiplicity and overlap.
Four- and five-node interactions show little detection.
Reciprocal interactions are detectable, but fewer trajectories
remain numerically valid at stronger coupling.\par}
\end{figure*}

We first examined how network size and interaction structure affected the
detectability of the planted interactions. Figure~\ref{fig:topology-detectability-curves}
compares null-calibrated c-sensitivity across the tested layouts. Detailed metric-by-$\beta$ values are provided in Appendix
Figure~\ref{fig:topology-detectability-heatmaps-network}
for network size,
Figure~\ref{fig:topology-detectability-heatmaps-multiplicity}
for interaction multiplicity, and
Figure~\ref{fig:topology-detectability-heatmaps-overlap-size}
for interaction overlap and size. Each
interaction was evaluated on its own target and source set using the threshold
from that layout's independent $\beta=0$ calibration ensemble. Values for the $N=4$ directed layout differ slightly from those in
Section~\ref{sec:results-base-detectability} because the topology experiments
use independently simulated calibration and test ensembles. For layouts
with multiple interactions, the curves average the separate interaction-level
scores. Simultaneous detection of all planted interactions is examined
separately below. Numerical validity is also reported separately for the reciprocal conditions.

The topology experiments produced a consistent overall pattern. Detection of
size-three interactions was robust to increasing network size, but shared
targets made simultaneous detection of multiple interactions harder. Source
overlap strengthened the response of KSG O-information, whereas interactions
with more sources became undetectable because the unscaled product coupling
became extremely small. Reciprocal interactions were detectable at
$\beta=10$, but their stronger-coupling results were limited by numerical
validity.

\FloatBarrier
Increasing network size did not reduce detection of the planted size-three
interaction. Because the measures were evaluated on the known planted
triplet, adding nodes did not enlarge the candidate search. Instead, it added
indirect pairwise pathways that could reshape dependence within the triplet
and obscure the higher-order signature. At $\beta=10$, KSG S-information
reached a c-sensitivity of $0.925$ and the pairwise KSG mutual-information
baseline reached $0.900$ for $N=4$, $8$, and $16$; both reached $1.000$ at
$\beta=20$. The broader pairwise network context therefore did not
appreciably weaken detection within the planted triplet.

Interaction multiplicity had a larger effect when several interactions
converged on the same target. Two interactions remained strongly detectable
whether their targets were shared or separate. With four interactions at
$\beta=20$, mean c-sensitivity was $0.931$ for KSG mutual information and
$0.831$ for KSG S-information when the target was shared, compared with
$1.000$ for both measures when the targets were separate. We also counted a
replicate as successful only when all four planted interactions exceeded
their respective thresholds. Under this stricter criterion, the success
fractions were $0.850$ for KSG mutual information and $0.550$ for KSG
S-information in the shared-target layout, but $1.000$ for both measures when
the targets were separate. Shared targets therefore affected simultaneous
detection more strongly than the mean interaction-level scores suggest.

Interaction overlap affected KSG O-information differently. At $\beta=20$,
its mean c-sensitivity was $0.938$ when two interactions shared a source,
compared with $0.738$ when their sources and targets were disjoint. When both
a source and the target were shared, c-sensitivity fell to $0.638$. The
stronger response was therefore associated specifically with source overlap,
rather than overlap in general. This may occur because the two shared-source
products, $z_1z_2$ and $z_1z_3$, both contain fluctuations from $z_1$.
However, O-information is symmetric and does not identify which variable is
the shared source. Streitberg information also varied with interaction
arrangement, whereas the Gaussian estimators remained weak.

Increasing interaction size produced a different failure. For both the
size-four and size-five conditions at $\beta=20$, c-sensitivity was $0.050$
for KSG mutual information and $0.025$ for KSG S-information, with no
consistent increase across the coupling sweep. This resulted from a collapse
in the realised higher-order contribution as more small-amplitude source
states were multiplied together (Appendix
Figure~\ref{fig:topology-component-norms}). At $\beta=20$, the whole-network
higher-order increment RMS decreased by approximately two orders of magnitude
with each additional source, from $1.62\times10^{-5}$ for the size-three
interaction to $1.45\times10^{-9}$ for the size-five interaction, placing the
size-four and size-five contributions approximately $400$ and $41\,000$ times
below the process-noise increment RMS of $6.00\times10^{-5}$. Their failure
therefore does not show that the estimators cannot detect interactions
involving larger sets, but that the fixed-coefficient product coupling does not
preserve realised interaction strength as the number of sources increases.
Section~\ref{sec:coupling-interaction-order} addresses this directly by
comparing the raw product with scale-normalised and alternative coupling
functions.

The reciprocal conditions require more cautious
interpretation. At $\beta=10$, 33/40 trajectories in the equal-strength condition
and 38/40 in the normalised condition remained valid. KSG S-information detected a
collective dependence change on the reciprocal triplet in $31/33$ and $22/38$
of these trajectories, giving c-sensitivities of $0.939$ and $0.579$,
respectively (Appendix
Figure~\ref{fig:topology-detectability-heatmaps-reciprocity}). These are the
primary reciprocal results because most trajectories were represented.

The loss of validity at stronger coupling is consistent with nonlinear
positive feedback. The reciprocal terms form a closed loop in which each node
influences the other two. When $z_0z_1z_2>0$, this feedback can increase the
triplet's magnitude and overcome the stabilising linear decay. Scaling each reciprocal coefficient to $\beta/\sqrt{3}$ weakens
this feedback, explaining the greater numerical validity of the
normalised condition.

At $\beta=20$, c-sensitivity reached 1.000 and 0.969, but only
10/40 trajectories in the equal-strength condition and 32/40
in the normalised condition remained valid; at $\beta=30$,
validity fell to 1/40 and 14/40. Because valid and diverging
trajectories were not compared, these high-coupling scores may be selection
biased and should be interpreted only as detection within the bounded subset. The $\beta=20$
numerical diagnostics are summarised in Appendix
Table~\ref{tab:app-topology-diagnostics}, while the complete reciprocal
c-sensitivity sweep and numerical-validity counts are reported in Appendix
Figure~\ref{fig:topology-detectability-heatmaps-reciprocity}.

\FloatBarrier

\subsubsection{Triplet recovery across topologies}
\label{sec:results-topology-recovery}

Recovery was more difficult than detection because the planted interactions
had to be identified among all admissible candidates. As network size
increased from $N=4$ to $N=16$, the unordered search expanded from 4 to 560
candidate sets and the directed search from 12 to 1680 assignments. The main
result was that recovery declined as the candidate space grew, even though
the planted interaction remained detectable in Section~\ref{sec:results-topology-detectability}. Recovery was also lower when multiple interactions converged on the same target.

\begin{figure}[H]
    \centering
    \includegraphics[width=0.99\textwidth]
    {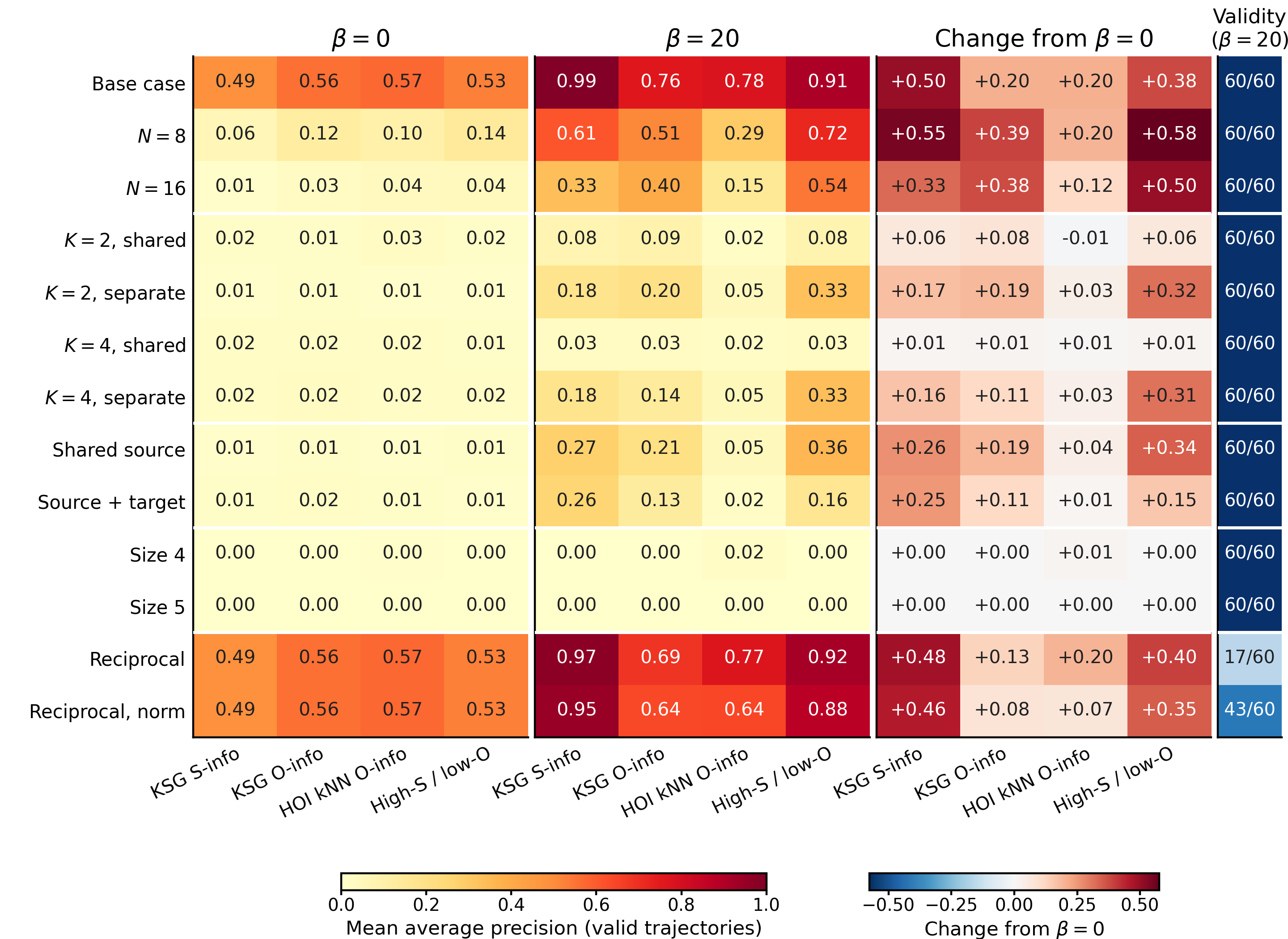}
    \caption{
    Unordered-set recovery across topology regimes. Heatmaps show mean
    average precision at $\beta=0$ and $\beta=20$ and the change between
    them. The right column reports validity at $\beta=20$; scores are
    conditional on validity.
    }
    \label{fig:topology-unordered-recovery}
    \par\smallskip
{\small\raggedright
\textbf{Alt text:} Heatmaps compare unordered-set recovery across
topologies. Recovery improves with coupling for most three-node
interactions, but declines in larger networks and with shared targets.
Larger interactions show little improvement; reciprocal results
use fewer valid trajectories.\par}
\end{figure}

Figure~\ref{fig:topology-unordered-recovery} shows that, across network sizes, KSG S-information unordered mean average precision
decreased from $0.992$ at $N=4$ to $0.333$ at $N=16$. The combined
high-S/low-O score declined less sharply, from $0.908$ to $0.544$, and at
$N=16$ ranked the planted set first in $0.450$ of replicates, compared with
$0.200$ for S-information alone.

As interaction multiplicity increased, recovery depended strongly on whether
the targets were separate. For two interactions, high-S/low-O unordered mean
average precision was $0.325$ with separate targets but $0.078$ with a shared
target. For four interactions, the corresponding values were $0.334$ and
$0.028$. Recovery with a shared source was similar to recovery with disjoint sources and targets. High-S/low-O unordered mean
average precision was $0.358$ when the interactions shared a source and
$0.325$ when they were disjoint, but decreased to $0.165$ when both a source
and the target were shared. Directed MMI-PID similarly reached $0.378$ with
a shared source, compared with $0.363$ for disjoint interactions, but fell
to $0.143$ when both a source and the target were shared. The stronger
O-information detection response under source overlap did not improve its
recovery, which remained similar in the shared-source and disjoint layouts
($0.205$ and $0.200$).

\begin{figure}[H]
    \centering
    \includegraphics[width=0.99\textwidth]
    {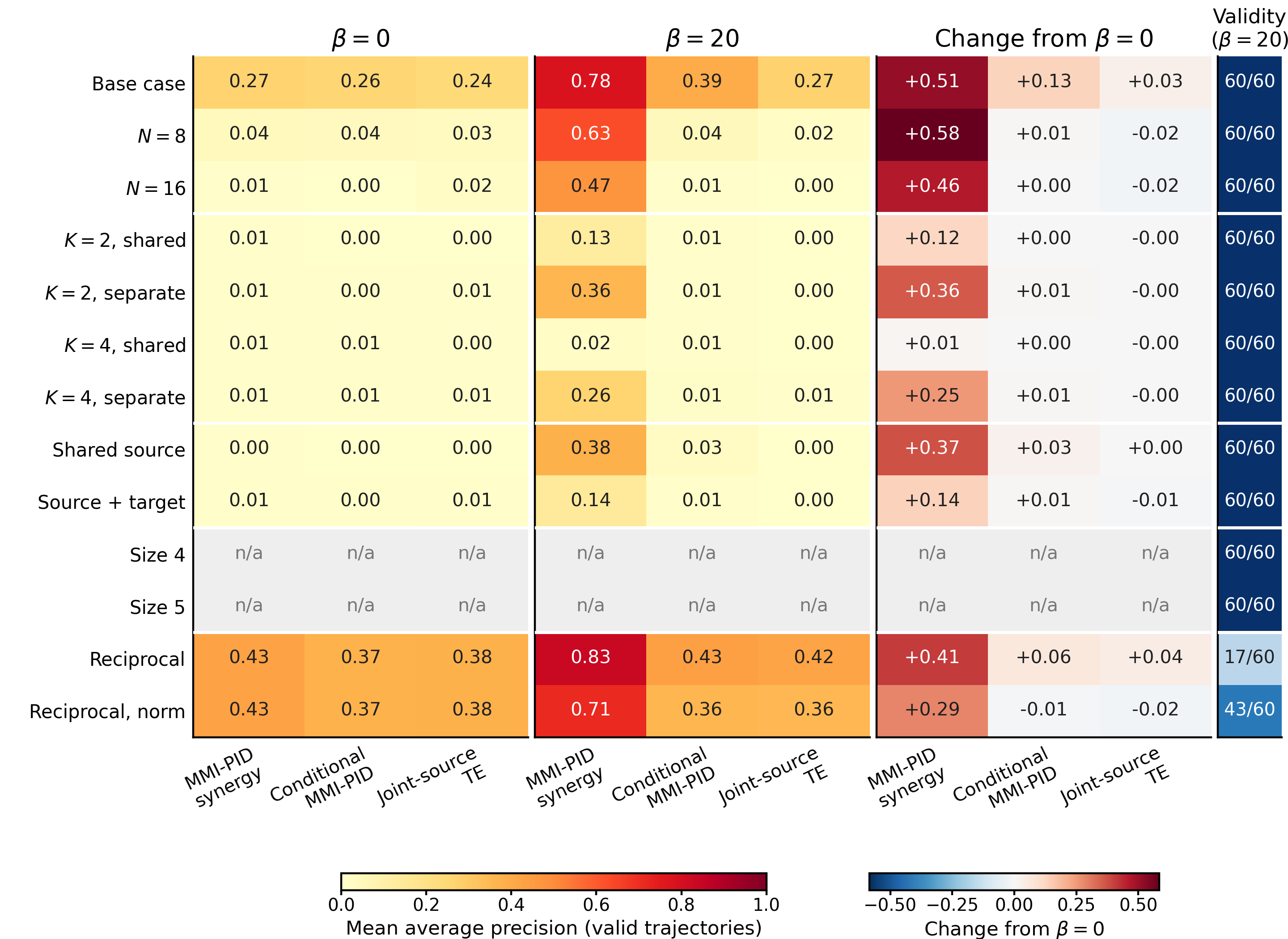}
    \caption{
    Directed recovery across topology regimes. Heatmaps show mean average
    precision at $\beta=0$ and $\beta=20$ and the change between them. The
    right column reports validity at $\beta=20$; grey cells indicate layouts
    for which directed recovery was not evaluated.
    }
    \label{fig:topology-directed-recovery}
    \par\smallskip
{\small\raggedright
\textbf{Alt text:} Heatmaps compare directed recovery across topologies.
Unconditional MMI-PID shows the largest improvements with coupling,
though recovery declines in larger networks and with shared targets.
Reciprocal results use fewer valid trajectories; larger interactions
were not evaluated.\par}
\end{figure}

For the size-four and size-five interactions in Figure~\ref{fig:topology-unordered-recovery}, unordered recovery remained
near its matched-null level, consistent with their weak realised coupling and
absent detectability. Directed recovery was not evaluated because the
two-source MMI-PID decomposition does not extend directly to these larger
source sets.

Finally, unordered recovery in the reciprocal configurations was
assessed primarily at $\beta=10$, where 49/60 trajectories in the
equal-strength condition and all 60 in the normalised condition
remained valid. KSG S-information achieved
mean average precision values of $0.939$ and $0.833$, respectively. This
strong performance reflects the model structure, as the three reciprocal
terms act within the same triplet, $\{0,1,2\}$, reinforcing its multivariate
dependence without requiring the individual directions to be distinguished.
Topology-specific null calibration further separates this higher-order
signature from dependence generated by the pairwise backbone. The stronger equal-strength result is consistent with its larger
combined higher-order contribution relative to the normalised condition.

Figure~\ref{fig:topology-directed-recovery} shows that unconditional MMI-PID
was the only directed measure to achieve meaningful recovery beyond the
base-case and reciprocal layouts. Its mean average precision nevertheless
declined from $0.780$ at $N=4$ to $0.469$ at $N=16$. At $N=16$, conditional
MMI-PID and joint-source transfer entropy reached only $0.006$ and $0.002$,
respectively, and remained at or near their null levels across the other
topology regimes. Unconditional MMI-PID was also consistently more successful
when interactions had separate rather than shared targets.

The $\beta=10$ results provide the primary evidence for reciprocal recovery.
At this coupling strength, 49/60 trajectories in the equal-strength
condition and all 60 in the normalised condition remained valid.
For directed recovery, MMI-PID reached $0.691$ and $0.542$, but all three
planted directions occupied the first three ranks in only $0.265$ and
$0.050$ of valid replicates. These results support reliable identification
of the participating set and partial recovery of direction, but not
consistent reconstruction of the complete reciprocal structure. The unordered and directed scores are not directly comparable because they
evaluate different ranking tasks. Symmetric measures rank participating node
sets, whereas target-specific measures rank directed source--target
assignments over a larger candidate space.

Although the reciprocal scores increased at $\beta=20$, they were
based on only 17/60 trajectories in the equal-strength condition
and 43/60 in the normalised condition. We did not
establish that these valid trajectories were representative of those that
diverged, so the apparent improvement may reflect selection bias. The
$\beta=20$ values are therefore reported in the figures as conditional
sensitivity results and are not interpreted as evidence that stronger
coupling improved recovery. The $\beta=10$ result for the normalised condition, for which
all trajectories remained valid, provides the strongest reciprocal evidence;
the equal-strength estimate remains conditional on the $49/60$ valid
trajectories.

Overall, detection remained robust across network sizes, whereas both
unordered and directed recovery became less reliable as the candidate space
grew and when multiple interactions shared the same target.

\subsubsection{Sensitivity to coupling function across interaction sizes}
\label{sec:coupling-interaction-order}

The product coupling used in the main experiments multiplies the states of
all sources participating in an interaction. When this coupling is applied
without an amplitude adjustment, its realised magnitude decreases as
interaction size increases because it multiplies an increasing number of
small-amplitude source states. We refer to this original implementation as
the \emph{unscaled product}. We repeated the detection and recovery
experiments in the $N=16$ network for interactions containing three, four,
or five nodes. The unscaled product was compared with an
\emph{RMS-normalised product} control and three other couplings: sourcewise tanh,
sine, and smooth root, as defined in
Section~\ref{sec:methods-coupling-sensitivity}. These alternatives were also scaled to the RMS magnitude of the
size-three unscaled product for each interaction size.

\begin{figure}[H]
    \centering
    \includegraphics[width=\textwidth]
        {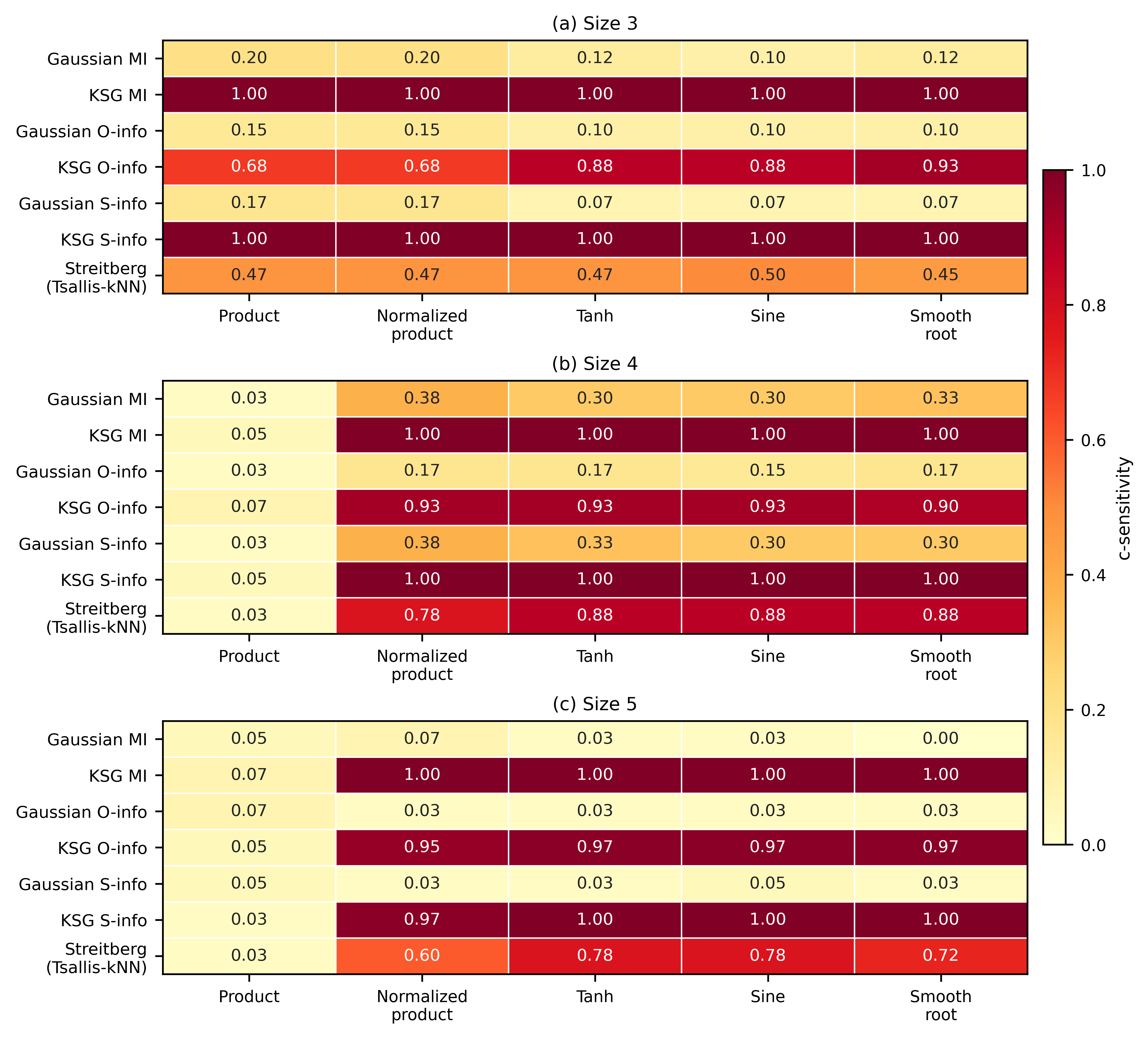}
    \caption{
    Detection across coupling functions in the $N=16$ network at
    $\beta=20$. Panels (a)--(c) correspond to interactions containing three,
    four, and five nodes. Rows identify the information-theoretic measure and
    columns identify the coupling function. Cells show c-sensitivity across
    40 test replicates using thresholds estimated from 200 independent
    $\beta=0$ simulations. The unscaled product retains its natural
    interaction-size-dependent amplitude. For each interaction size, the
    RMS-normalised product and the three other couplings use source scales and
    fixed output-amplitude constants calibrated to the RMS magnitude of the
    size-three unscaled product, as described in
    Section~\ref{sec:methods-coupling-sensitivity}.
    }
    \label{fig:coupling-order-detection}
    \par\smallskip
{\small\raggedright
\textbf{Alt text:} Three heatmaps compare detection across coupling
functions and interaction sizes. KSG mutual information and
S-information detect three-node interactions reliably. For four-
and five-node interactions, amplitude rescaling restores strong
detection lost with the unscaled product.\par}
\end{figure}

Here, RMS normalisation refers only to an offline calibration used to place the
coupling-function outputs on a common amplitude scale. For each interaction
size, source scales were first estimated from an independent $\beta=0$
ensemble to place the source states on comparable input scales. A separate output-amplitude constant was then estimated for each
coupling function and interaction size so that its coupling-output
RMS was the same as that of the size-three unscaled product. These constants were fixed before the evaluation
simulations and were not adjusted across trajectories or time. The
higher-order term was therefore not dynamically normalised in the state
equation: $\beta$, the planted interaction structure, and all other model
terms remained unchanged. The unscaled product retained its original form
and natural interaction-size-dependent amplitude. This design separates
sensitivity to coupling form from the attenuation produced by multiplying
additional small-amplitude sources.

At size three, KSG mutual information and KSG S-information detected every
coupling condition in all replicates
(Figure~\ref{fig:coupling-order-detection}(a)). The unscaled and RMS-normalised products were identical at this
size because the size-three unscaled product defined the common
amplitude reference. KSG O-information was more sensitive
to coupling form, increasing from $0.68$ for the two product conditions to
$0.88$--$0.93$ for the other couplings.

\begin{figure}[H]
    \centering
    \includegraphics[width=\textwidth]
        {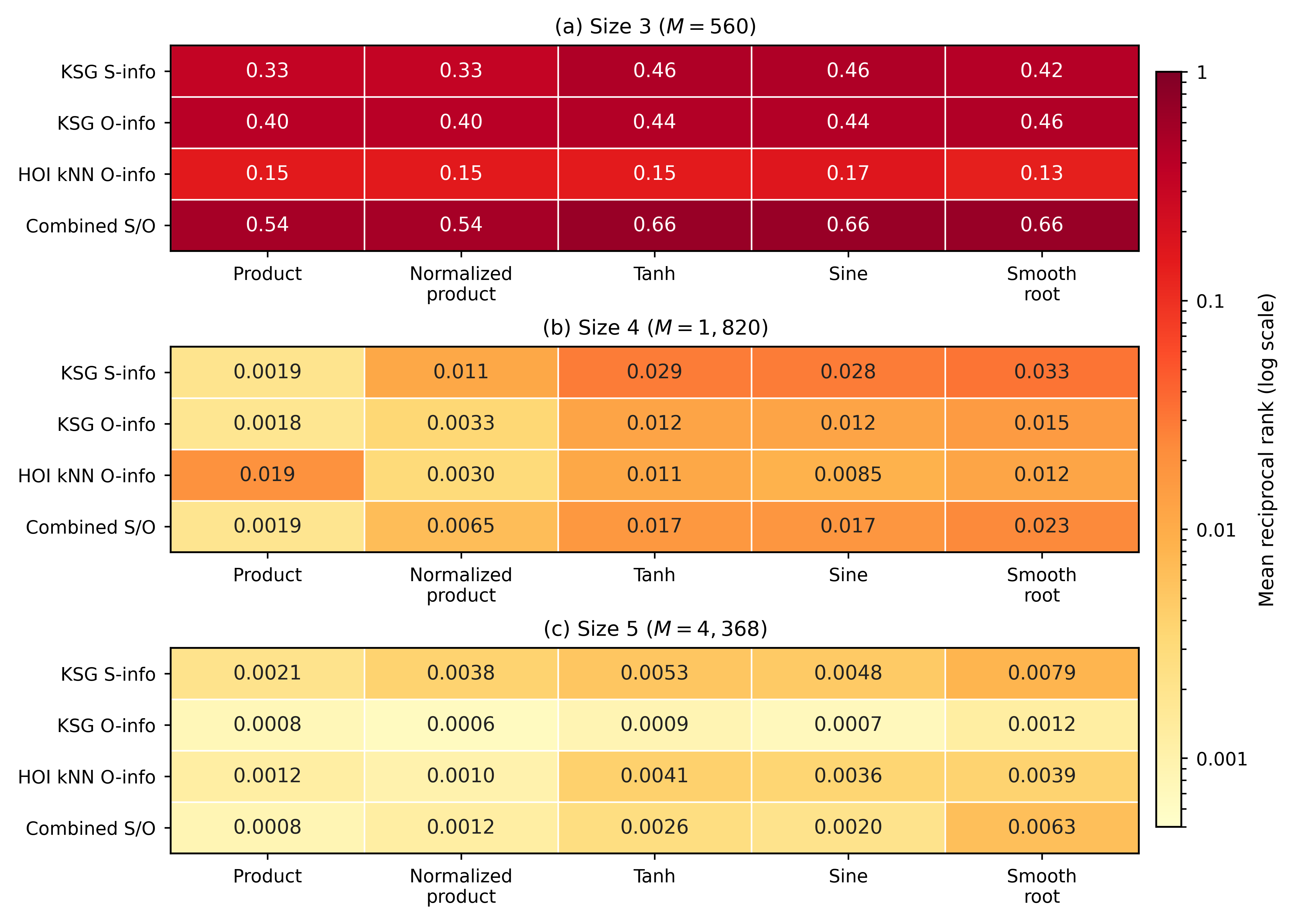}
    \caption{
    Recovery of the planted unordered interaction in the $N=16$ network at
    $\beta=20$. Panels (a)--(c) correspond to interaction sizes three, four,
    and five; $M$ is the number of candidate sets. Rows identify the ranking
    measure and columns identify the coupling function. Cells show mean
    reciprocal rank across 60 replicates using candidate-specific
    $\beta=0$ calibrations. Higher values indicate better localisation, and
    the common colour scale is logarithmic. Because each condition contains
    one planted interaction, mean reciprocal rank equals mean average
    precision.
    }
    \label{fig:coupling-order-recovery}
    \par\smallskip
{\small\raggedright
\textbf{Alt text:} Three heatmaps compare recovery across coupling
functions. The combined score performs best for three-node
interactions. Recovery scores are much lower for four- and
five-node interactions, although alternative couplings improve
many rankings.\par}
\end{figure}

\clearpage
The effect of RMS normalisation became clear at larger interaction sizes in Figure~\ref{fig:coupling-order-detection}. For the
unscaled product, KSG S-information c-sensitivity fell to $0.05$ at size four
and $0.03$ at size five. By contrast, it reached $1.00$ at size four and
$0.97$--$1.00$ at size five for the RMS-normalised product and the three other
couplings. KSG mutual information also reached $1.00$ for each of these four conditions at both sizes. These results show that the
apparent loss of detectability for larger product interactions arose
primarily from the interaction-size-dependent attenuation of the unscaled
product rather than from an intrinsic inability of the measures to detect
interactions involving larger sets.

Recovery remained more difficult than detection
(Figure~\ref{fig:coupling-order-recovery}). At size three, the combined
high-S/low-O score achieved a mean reciprocal rank of $0.544$ for both
product conditions, whereas the other couplings increased it to
$0.662$--$0.664$. Additional summaries of the same candidate rankings, which
are not displayed in the figure, gave top-1 recovery rates of $0.450$ for the
product conditions and $0.583$ for each of the other couplings.

At size four, RMS normalisation increased the mean reciprocal rank of the combined
score from $0.0019$ for the unscaled product to $0.0065$ for the RMS-normalised
product. The other couplings produced larger improvements, reaching
$0.017$--$0.023$. Exact top-1 recovery was nevertheless absent apart from one
unscaled-product replicate under HOI kNN O-information. With the smooth-root
coupling, the combined score placed the planted set at a median rank of
$104.5$ among $1{,}820$ candidates. At size five, no method achieved top-1
recovery; the highest mean reciprocal rank was $0.0079$, obtained by KSG
S-information with the smooth-root coupling.

Scaling the coupling outputs to a common RMS on the $\beta=0$ ensemble restored high detection rates for KSG mutual information and KSG S-information at interaction sizes four and five. Alternative coupling functions also improved some recovery rankings. First-rank recovery remained rare or absent for these larger interactions. The common RMS was imposed on the scaling ensemble, and realised contributions at non-zero coupling may differ. Interaction size and the number of candidate sets increased together, so these comparisons reflect their combined effects on recovery. Across the benchmark, detecting statistical changes induced by specified higher-order network interactions was more successful than recovering their participating nodes and source-target roles.

\section{Conclusion}
\label{sec:conclusion}

We introduced a generative model of stochastic network dynamics and a framework for evaluating information-theoretic measures against higher-order interactions. Across coupling functions, network topologies and interaction sizes, the experiments assess detection of changes relative to a pairwise-only reference and recovery of the participating nodes. Directed recovery additionally evaluates source-target roles for three-node interactions. The framework also tests whether the planted
higher-order coupling can be distinguished from a specified decomposable
nonlinear pairwise alternative. Used independently, the model can generate continuous multivariate time series
with controlled pairwise and higher-order structure. The framework allows new measures, mechanisms and network configurations to be tested using the same experimental setup.

Detection and recovery experiments produced different results. At sufficiently strong coupling, KSG mutual information and KSG S-information achieved high detection rates across the tested coupling functions and network sizes. Scaling coupling outputs to a common RMS on the $\beta=0$ ensemble also supported detection of interactions with more sources. Recovery was less successful in larger candidate spaces and in shared-target layouts. Identifying source--target assignments was harder than identifying the participating nodes. In the base case, KSG O-information had lower detection rates than KSG S-information. In the discrimination experiments, the measures separated the product coupling from the specified nonlinear additive control. False-positive rates were close to the $5\%$ level in the independent validation at $\beta=20$, and exceeded this level for some measures in the across-$\beta$ experiment. Together, these results show that statistical signatures can respond to a
planted higher-order mechanism and distinguish it from a specified nonlinear
pairwise alternative, while recovering its participating nodes and direction
remains more challenging, particularly as the candidate space grows or
multiple interactions share a target.

These conclusions are limited to the conditions tested. All model states were observed, with process noise and no observation model. Detection and recovery used a known pairwise-only reference, and discrimination used the specified nonlinear additive control. Performance summaries for reciprocal layouts were conditional on numerical validity. The higher-order mechanisms
examined were built by multiplying separate functions of individual source
states, which leaves out interactions defined through a non-factorised joint
function or a collective threshold rule, and the estimators were used with
fixed, prespecified settings. The results should therefore be read as
outcomes for the mechanisms and settings examined here rather than as a
general ranking of higher-order detection methods. The modular construction of the
framework supports extensions in each of these directions, including partial
observation, measurement noise, learned rather than known pairwise and higher-order
references, further coupling functions, and larger networks with structured
candidate searches. Because the planted structure and the corresponding pairwise-only reference
are retained, such extensions can test which aspects of a higher-order
mechanism remain detectable and recoverable as the setting becomes more
realistic.

For empirical time-series analysis, our results highlight the distinction between detecting statistical signatures and identifying the interactions that generated them. Validation frameworks with known generating mechanisms are therefore essential for assessing whether inferred interactions correspond to those actually present in the simulated system. Our results suggest that KSG S-information may help identify candidate groups of interacting nodes, while the combined S/O score may improve recovery in larger candidate spaces. To guide inference for a particular type of data, the model should be adapted to reflect the relevant properties of the system and its recordings, allowing measures to be validated under conditions appropriate to that application. Future work could examine whether, and under what conditions, other statistical measures or inference models improve recovery across different types of synthetic data. Adding measurement models and developing model-inversion methods could provide a route to estimating interaction structures and their parameters from observed signals. Together, these developments could help establish when statistical signatures support reliable conclusions about the mechanisms underlying different types of empirical time series data.

\section*{Funding}
\begin{sloppypar}
This work was supported by the Digital Sciences for Society of Tilburg University [DSFS 2023ICON006]. 
KVH gratefully acknowledges funding from the Dutch Research Council
[NWO; VIDI grant no. 09150171910043].
\end{sloppypar}

\section*{Acknowledgements}

OpenAI Codex, ChatGPT, and Claude were used as supporting tools for manuscript revision, code development, numerical consistency checks, and submission preparation. After using these
tools, the authors reviewed and edited the content as needed and take full responsibility for the content of the article.

\section*{Data and code availability}

This study uses synthetic time series data generated by the model
described in this article. The code for the model and experimental
pipeline is maintained at
\url{https://github.com/ValentinaSanchezMelchor/higher-order-network-dynamics}.

\section*{Conflict of interest}
The authors declare no conflicts of interest.

\clearpage
\appendix
% Keep appendix PDF destinations distinct from the main-text float counters.
\renewcommand{\theHfigure}{appendix.\arabic{figure}}
\renewcommand{\theHtable}{appendix.\arabic{table}}
\setcounter{figure}{0}
\setcounter{table}{0}
\section{Background on the model and information-theoretic measures}
\label{app:background}

\subsection{DCM neural state equation}
\label{app:dcm}

Dynamic causal modelling (DCM) was introduced by Friston et al.~\cite{friston2003dcm} as a Bayesian framework for inferring effective connectivity between brain regions from neuroimaging data. It combines a model of neural dynamics with an observation model linking neural activity to measured signals. Smith et al.~\cite{smith2011network} used a DCM neural state equation coupled to a haemodynamic model to generate synthetic fMRI time series with known connectivity. We use their neural state formulation as the starting point for our stochastic network model.

In the formulation used by Smith et al.~\cite{smith2011network}, the neural state equation is
\begin{equation}
\label{eq:dcm}
\dot z(t)
= \sigma A z(t)+Cu(t),
\end{equation}
where $z(t)\in\mathbb R^N$ contains the neural states of $N$ regions and $\dot z(t)$ is its time derivative. The matrix $A \in\mathbb R^{N\times N}$ has diagonal entries $A_{ii}=-1$, representing decay within each node, and off-diagonal entries $A_{ij}$ that specify the coupling from source $j$ to target $i$. The coefficient $\sigma>0$ scales the decay and pairwise coupling rates. Without pairwise coupling or input, a node decays with time constant $1/\sigma$. The vector $u(t)\in\mathbb R^M$ contains $M$ inputs, and $C\in\mathbb R^{N\times M}$ determines how these inputs affect each node. In their simulations, the inputs switched between two levels according to a Poisson process, with added neural noise.

We analyse the simulated node states $z(t)$, without a haemodynamic observation model. We specify the generating parameters so that we can focus on evaluating the information-theoretic measures under known interaction mechanisms, without introducing the additional statistical inference problem of estimating those parameters using DCM. In our model, self-decay is represented by $\lambda>0$ and pairwise coupling by a matrix $A$ with $A_{ii}=0$. The symbol $A$ therefore has a different meaning from the matrix in Equation~\ref{eq:dcm}, which includes self-decay. The linear state-dependent terms agree when $\lambda=\sigma$ and the off-diagonal entries of our coupling matrix equal $\sigma$ times the corresponding off-diagonal entries of the DCM matrix. We allow $\lambda$ and the pairwise coupling strengths to be specified separately, add higher-order couplings and additive process noise, and use Ornstein-Uhlenbeck inputs in the simulations. The stochastic state equation is given in Section~\ref{sec:methods-model}, and the input processes are described in Section~\ref{sec:base-case}.

\subsection{Higher-order mechanisms and directed representations}
\label{app:higher-order}

Higher-order descriptions of a system can refer to its network structure, its coupling functions, or statistical dependence in its outputs \cite{bick2023higher,rosas2022disentangling}. A hyperedge identifies a group of participating nodes, but even when it joins three or more nodes, the associated coupling can be a sum of pairwise contributions \cite{neuhaeuser2024effective}. The number of participating nodes therefore does not determine whether the coupling requires a higher-order description. In this paper, we use the non-additivity criterion in Equation~\ref{eq:decomposability} to identify higher-order mechanisms in the original node states. Directed hyperedges represent the participating sources and target, while information-theoretic measures characterise the statistical dependence in the resulting time series.

The functions $g_{ij}$ and $g_{ik}$ in Equation~\ref{eq:decomposability} may be non-linear, so the criterion concerns additivity \cite{lim2015interactions}. When that decomposition exists, $g_{ij}(z_j)$ and $g_{ik}(z_k)$ describe separate pairwise contributions from sources $j$ and $k$ to target $i$. For example, the coupling $z_j^2+z_k^2$ is non-linear but additive in the source states and can be represented by two non-linear pairwise terms. For the product coupling $z_jz_k$, a non-zero increment $\delta$ in $z_j$ changes the coupling by $\delta z_k$. The effect of changing one source therefore depends on the state of the other, which prevents an additive decomposition into separate source contributions.

To specify where a joint source contribution acts, we use the 1-directed hypergraph convention of Gallo et al.~\cite{gallo2022directed}, in which each hyperedge has a single target. Let $V$ denote the node set. A directed triplet is represented by
\begin{equation}
\label{eq:directed-hyperedge}
e=(i,\{j,k\}),
~i\in V,
~\{j,k\}\subseteq V\setminus\{i\},
~j \neq k,
\end{equation}
where $i$ is the target and $\{j,k\}$ is the unordered source pair. Exchanging the source labels leaves the hyperedge unchanged, while choosing a different target defines a different directed interaction. This extends the convention for $A_{ij}$, which represents a contribution from source $j$ to target $i$. The direction identifies the state equation receiving the joint contribution. The same interaction can also be represented as a focal interaction with support $\{i,j,k\}$ and focal node $i$ in the framework of Moutuou~\cite{moutuou2026}.

Information-theoretic measures describe properties of the joint distribution of the variables being analysed. This distribution reflects the combined effects of the network topology, coupling functions, interaction strengths, inputs, and process noise. Pairwise mechanisms can produce multivariate dependence beyond that captured by pairwise summaries \cite{rosas2022disentangling}. A statistical signature therefore does not uniquely identify a higher-order mechanism, and a weak response from a measure does not establish that such a mechanism is absent. In our approach, comparisons with the pairwise-only reference assess changes induced by the added higher-order term. The separate comparison with a non-linear additive control tests whether a measure can discriminate between those two coupling mechanisms.

\subsection{Information-theoretic measures}
\label{app:information-measures}

We characterise statistical dependence in the simulated node states using the information-theoretic measures summarised in Table~\ref{tab:metric-suite}. The table gives their definitions, interpretations, labels used in the figures, and roles in the experiments. It also includes the combined S/O score, a ranking statistic for candidate node sets. Estimator settings and implementation details are given in Section~\ref{sec:estimators-implementation}.

\begin{table}[!t]
  \centering
  \small
  \setlength{\tabcolsep}{3pt}
  \renewcommand{\arraystretch}{1.15}
  \caption{Definitions, interpretations, figure labels, and experimental roles of the information-theoretic measures and combined S/O ranking score. Interpretations of the measures refer to population quantities. Italic text gives the figure labels. Gaussian labels refer to covariance-based estimates, KSG to Kraskov-St\"ogbauer-Grassberger estimation, and kNN to $k$-nearest neighbours. HOI identifies the software package. In the ``Use'' column, $\uparrow$ and $\downarrow$ indicate the upper and lower reference tails used for recovery. Detection and discrimination use two-sided comparisons.}
  \label{tab:metric-suite}
  \begin{tabularx}{\linewidth}{@{}
      >{\raggedright\arraybackslash}p{3.1cm}
      >{\raggedright\arraybackslash}p{4.9cm}
      >{\raggedright\arraybackslash}X
      >{\raggedright\arraybackslash}p{2.0cm}@{}}
    \toprule
    \textbf{Quantity / figure labels}
      & \textbf{Definition}
      & \textbf{Interpretation}
      & \textbf{Use} \\
    \midrule

    \textbf{Mutual information}
    \citep{shannon1948,cover2006}
    \newline
    \emph{Gaussian MI; KSG MI}
      & $I(X;Y)=H(X)+H(Y)-H(X,Y)$
      & Larger values indicate greater pairwise dependence
      & Detection; discrimination \\
    \addlinespace

    \textbf{O-information}
    \citep{rosas2019oinfo}
    \newline
    \emph{Gaussian O-info; KSG O-info; HOI kNN O-info}
      & $\Omega(\mathbf z)=\mathrm{TC}(\mathbf z)-\mathrm{DTC}(\mathbf z)$
      & $\Omega>0$: redundancy-dominated dependence;
        $\Omega<0$: synergy-dominated dependence
      & Detection; discrimination; set recovery $\downarrow$ \\
    \addlinespace

    \textbf{S-information}
    \citep{rosas2019oinfo}
    \newline
    \emph{Gaussian S-info; KSG S-info}
      & $\Sigma(\mathbf z)=\mathrm{TC}(\mathbf z)+\mathrm{DTC}(\mathbf z)$
      & Larger values indicate greater overall dependence,
        including pairwise contributions
      & Detection; discrimination; set recovery $\uparrow$ \\
    \addlinespace

    \textbf{Streitberg information}
    \citep{liu2024streitberg}
    \newline
    \emph{Streitberg (Tsallis-kNN)}
      & $\mathrm{SI}^{(d)}=
          \sum_{\pi\in\Pi_d}\mu(\pi)
          D_\alpha(P_\pi\|\prod_{r=1}^{d}P_r)$
      & Zero if the variables can be partitioned into two or more
        mutually independent groups
      & Detection; discrimination \\
    \addlinespace

    \textbf{MMI-PID synergy}
    \citep{williams2010nonnegative,barrett2015exploration}
    \newline
    \emph{KSG MMI-PID; Conditional KSG MMI-PID}
      & $\begin{aligned}[t]
          \mathrm{Syn}_{\mathrm{MMI}}
            ={}& I([X_1,X_2];Y) \\
               & {}-\max_{m\in\{1,2\}}I(X_m;Y)
        \end{aligned}$
        \newline
        Conditional form: replace each $I(\cdot;\cdot)$
        by $I(\cdot;\cdot\mid Z)$
      & Additional target information from the joint sources
        relative to the more informative source
      & Directed recovery $\uparrow$ \\
    \addlinespace

    \textbf{Joint-source transfer entropy}
    \citep{schreiber2000,novelli2019}
    \newline
    \emph{KSG joint-source TE}
      & $I([X_1,X_2];Y\mid Z)$
      & Information about the target's next state from the source
        pair, conditional on the target's current state
      & Directed recovery $\uparrow$ \\
    \addlinespace

    \textbf{Combined S/O score}
    \newline
    \emph{KSG high-S / low-O}
      & $Q=-\log p_{S,\mathrm{upper}}
           -\log p_{O,\mathrm{lower}}$
      & Combines upper-tail S-information and lower-tail
        O-information departures from the pairwise-only reference
      & Set recovery $\uparrow$ \\

    \bottomrule
  \end{tabularx}
\end{table}

Let $\mathbf z=(z_1,\ldots,z_d)$ denote a random vector of $d$ node states, and let $\mathbf z_{-r}$ be the vector obtained by removing component $r$. For continuous random variables or vectors $X$, $Y$, and $Z$, $H(X)$ and $H(X\mid Y)$ denote differential entropy and conditional differential entropy, respectively \citep{cover2006}. The entropy expressions below assume that the relevant densities exist and that all entropy terms are finite. Mutual information is $I(X;Y) = H(X)+H(Y)-H(X,Y)$, and conditional mutual information is $I(X;Y\mid Z) = H(X\mid Z)+H(Y\mid Z)-H(X,Y\mid Z)$. Total correlation is $\mathrm{TC}(\mathbf z)=\sum_{r=1}^{d}H(z_r)-H(\mathbf z)$ \citep{watanabe1960}, and dual total correlation is $\mathrm{DTC}(\mathbf z) = H(\mathbf z)-\sum_{r=1}^{d}H(z_r\mid\mathbf z_{-r})$ \citep{han1978}.

O-information is defined as $\Omega(\mathbf z) = \mathrm{TC}(\mathbf z)-\mathrm{DTC}(\mathbf z)$ \citep{rosas2019oinfo}. Positive values indicate redundancy-dominated dependence, and negative values indicate synergy-dominated dependence. Dependent variables can also have $\Omega(\mathbf z) = 0$, so a zero value does not imply mutual independence. S-information is defined as $\Sigma(\mathbf z) = \mathrm{TC}(\mathbf z)+\mathrm{DTC}(\mathbf z)$ and summarises overall statistical dependence, including pairwise contributions. It is non-negative and equals zero if and only if the components of $\mathbf z$ are mutually independent.

Streitberg information is constructed using all partitions of the analysed variables \citep{liu2024streitberg}. Let $\Pi_d$ denote the set of partitions of $\{1,\ldots,d\}$, where each partition $\pi$ consists of non-empty, disjoint blocks whose union is $\{1,\ldots,d\}$. For a block $B$, let $P_B$ denote the joint marginal distribution of the states indexed by $B$, and write $P_\pi = \prod_{B\in\pi}P_B$ and $P_r = P_{\{r\}}$. Streitberg information is defined as $\mathrm{SI}^{(d)} = \sum_{\pi\in\Pi_d}\mu(\pi)D_\alpha(P_\pi\|\prod_{r=1}^{d}P_r)$, where $\mu(\pi) = (|\pi|-1)!(-1)^{|\pi|-1}$ and $|\pi|$ is the number of blocks. The Tsallis divergence is $D_\alpha(P\|Q) = (\int p(x)^\alpha q(x)^{1-\alpha}\,\mathrm{d}x-1)/(\alpha-1)$, where $p$ and $q$ are the densities of $P$ and $Q$; we use $\alpha=0.5$. If the variables can be partitioned into at least two mutually independent groups, their joint distribution factorises as $P_\pi$ and Streitberg information is zero at the population level.

Partial information decomposition (PID) separates the mutual information between a joint source vector $[X_1,X_2]$ and a target $Y$ into redundant information, information unique to each source, and synergy \citep{williams2010nonnegative}. We use the minimum mutual information (MMI) definition of redundancy, $\min_{m\in\{1,2\}}I(X_m;Y)$ \citep{barrett2015exploration}. The resulting synergy is $\mathrm{Syn}_{\mathrm{MMI}} = I([X_1,X_2];Y)-\max_{m\in\{1,2\}}I(X_m;Y)$. It quantifies the additional information about the target obtained by considering both sources together, relative to the more informative source. For directed recovery, we use $X_1 = z_j(t)$, $X_2 = z_k(t)$, and $Y = z_i(t+\Delta t)$, where $\Delta t$ is the integration step. The conditional version uses the current target state $Z = z_i(t)$ and is defined by $\mathrm{Syn}_{\mathrm{MMI}\mid Z} = I([X_1,X_2];Y\mid Z)-\max_{m\in\{1,2\}}I(X_m;Y\mid Z)$.

Using the same source and target variables, the one-step joint-source transfer entropy (TE) is defined as $I([X_1,X_2];Y\mid Z)$ \citep{schreiber2000,novelli2019}. It quantifies the information about the target's next state available from the current source pair, conditional on the target's current state. This quantity includes predictive information available from single sources as well as conditional synergy. Its direction refers to temporal prediction, and a non-zero value does not establish a causal interaction.

For unordered-set recovery, we combine estimates of S- and O-information using the ranking score $Q = -\log p_{S,\mathrm{upper}}-\log p_{O,\mathrm{lower}}$. The quantities $p_{S,\mathrm{upper}}$ and $p_{O,\mathrm{lower}}$ are empirical upper-tail probabilities for S-information and lower-tail probabilities for O-information, respectively, obtained from pairwise-only reference simulations for each node set. The score increases as either tail probability decreases, combining departures towards higher S-information and lower O-information. For each node set, we calculate an empirical null probability for $Q$ using its reference distribution, which incorporates dependence between the S- and O-information estimates. We rank the sets by these probabilities, following the procedure in Section~\ref{sec:triplet-recovery}.

% ============================================================
\section{Base-model validation}
\label{app:model-validation}

This appendix provides the numerical results supporting the model-validation
checks reported in Section~\ref{sec:model-validation}. The convergence
analysis evaluates the selected integration timestep, while the sensitivity
analysis tests whether the base model remains numerically finite and retains a
dynamically substantial higher-order contribution under perturbations of its
pairwise and input parameters and of its process-noise
intensity.

% ------------------------------------------------------------
\subsection{Timestep convergence}
\label{app:timestep-convergence}

\begin{figure}[!htbp]
  \centering
  \includegraphics[width=0.99\linewidth]
    {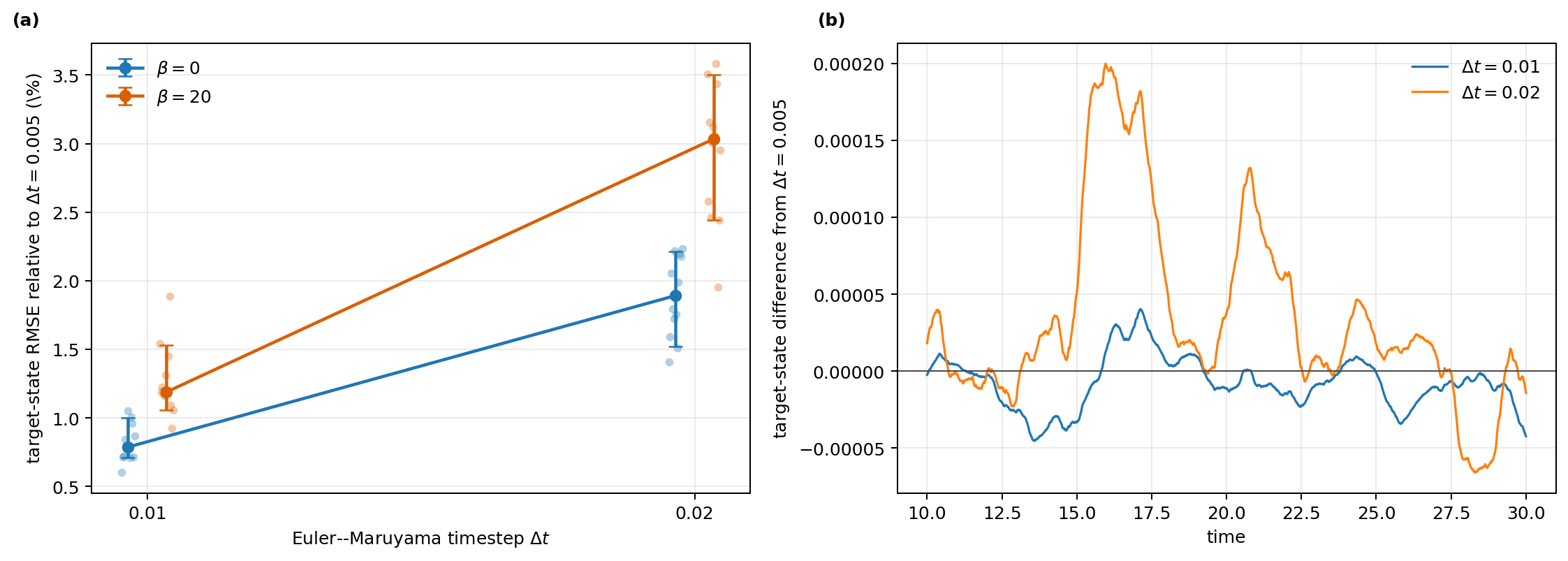}
  \caption{Timestep convergence of the base model. Panel~(a) shows the
  target-state RMSE at $\Delta t=0.01$ and $\Delta t=0.02$, relative to the
  reference solution at $\Delta t=0.005$. Small points show the 12 individual
  replicates, large points show the medians, and error bars span the 10th to
  90th percentiles. The reference step is omitted because its error is zero by
  construction. Panel~(b) shows the target-state difference from the reference
  solution for one representative $\beta=20$ replicate after warm-up. Initial
  states, OU paths, and Wiener paths were reused across timesteps, with
  fine-scale Wiener increments aggregated to construct the corresponding
  coarse increments.}
  \label{fig:validation-convergence}
  \par\smallskip
{\small\raggedright
\textbf{Alt text:} (a) Target-state errors decrease with a smaller
integration timestep, both with and without higher-order coupling.
(b) The smaller timestep generally produces smaller deviations
from the reference trajectory over time.\par}
\end{figure}

\begin{table}[!htbp]
  \centering
  \caption{Timestep-convergence results. RMSE values are expressed as percentages of the corresponding reference RMS at $\Delta t=0.005$ and are reported as the median with the 10th-90th percentile interval across 12 replicates.}
  \label{tab:validation-convergence}
  \small
  \begin{tabular}{ccccc}
    \toprule
    $\beta$
      & $\Delta t$
      & Finite runs
      & All-state RMSE (\%)
      & Target-state RMSE (\%) \\
    \midrule
    $0$
      & $0.02$
      & $12/12$
      & $2.631\;[2.532,\,2.689]$
      & $1.892\;[1.519,\,2.215]$ \\
    $0$
      & $0.01$
      & $12/12$
      & $1.087\;[1.053,\,1.147]$
      & $0.788\;[0.711,\,1.000]$ \\
    $20$
      & $0.02$
      & $12/12$
      & $2.636\;[2.538,\,2.697]$
      & $3.034\;[2.445,\,3.501]$ \\
    $20$
      & $0.01$
      & $12/12$
      & $1.092\;[1.057,\,1.149]$
      & $1.189\;[1.059,\,1.532]$ \\
    \bottomrule
  \end{tabular}
\end{table}

% ------------------------------------------------------------
\newpage
\subsection{Parameter sensitivity}
\label{app:parameter-sensitivity}

\begin{figure}[!htbp]
  \centering
  \includegraphics[width=\linewidth]
    {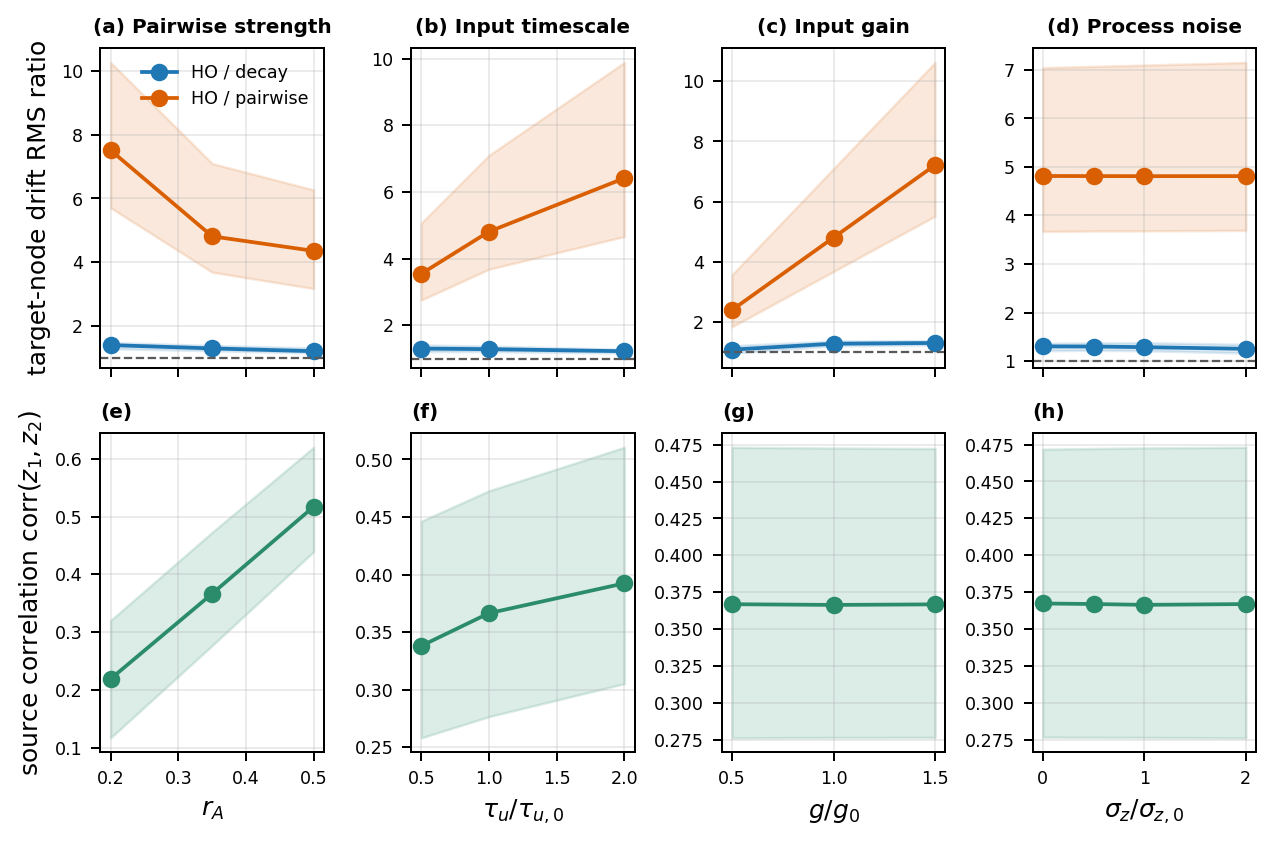}
  \caption{Sensitivity of the base model at $\beta=20$. The columns vary the
    pairwise spectral fraction $r_A$, the input correlation time relative to
    its base value $\tau_{u,0}$, the common input gain relative to its base
    value $g_0$, and the process-noise intensity relative to its base value
    $\sigma_{z,0}$. Panels~(a)--(d) show the median target-node
    higher-order-to-decay and higher-order-to-pairwise drift RMS ratios. The
    dashed line marks a ratio of one. Panels~(e)--(h) show the median zero-lag
    Pearson correlation between the two source states. Shaded regions span the
    10th to 90th percentiles across 30 replicates.}
  \label{fig:validation-sensitivity}
  \par\smallskip
{\small\raggedright
\textbf{Alt text:} Top panels show higher-order drift exceeding
pairwise and decay contributions across parameter variations.
Bottom panels show source correlation increasing with pairwise
strength and input timescale, while remaining nearly unchanged
with input gain and process noise.\par}
\end{figure}

\begin{table}[!htbp]
  \centering
  \caption{Condition-wise parameter-sensitivity results. Finite runs are
  reported separately for $\beta=0$ and $\beta=20$. A run was considered
  finite when all states were finite and
  $\max_{i,t}|z_i(t)|<10^{6}$. The linear margin is $\lambda-\alpha(A)$. In panel~(b), $\operatorname{corr}(z_1,z_2)$ is the zero-lag Pearson correlation between the source states. HO/decay and HO/pairwise are the ratios of the higher-order drift RMS to the decay and pairwise drift RMS, respectively, at target node~0. These three quantities are evaluated at $\beta=20$ and reported as the median with the 10th-90th percentile interval across 30 replicates. Here, $\tau_{u,0}=0.2756$, $g_0=0.02285$, and
  $\sigma_{z,0}=0.003\sqrt{0.02}=4.24\times10^{-4}$.}
  \label{tab:validation-sensitivity}

  \small
  \setlength{\tabcolsep}{4pt}
  \renewcommand{\arraystretch}{1.08}

  \textit{(a) Numerical validity}

  \smallskip
  \begin{tabular*}{\linewidth}{@{\extracolsep{\fill}}llccc@{}}
    \toprule
    Condition
      & Setting
      & Linear margin
      & \shortstack{Finite runs\\$\beta=0$}
      & \shortstack{Finite runs\\$\beta=20$} \\
    \midrule
    Baseline
      & \shortstack[l]{$r_A=0.35,\ \tau_u=\tau_{u,0},$\\
                       $g=g_0,\ \sigma_z=\sigma_{z,0}$}
      & $0.65$
      & $30/30$
      & $30/30$ \\
    Pairwise strength
      & $r_A=0.20$
      & $0.80$
      & $30/30$
      & $30/30$ \\
    Pairwise strength
      & $r_A=0.50$
      & $0.50$
      & $30/30$
      & $30/30$ \\
    Input correlation time
      & $\tau_u=0.5\tau_{u,0}$
      & $0.65$
      & $30/30$
      & $30/30$ \\
    Input correlation time
      & $\tau_u=2\tau_{u,0}$
      & $0.65$
      & $30/30$
      & $30/30$ \\
    Input gain
      & $g=0.5g_0$
      & $0.65$
      & $30/30$
      & $30/30$ \\
    Input gain
      & $g=1.5g_0$
      & $0.65$
      & $30/30$
      & $30/30$ \\
    Process noise
      & $\sigma_z=0$
      & $0.65$
      & $30/30$
      & $30/30$ \\
    Process noise
      & $\sigma_z=0.5\sigma_{z,0}$
      & $0.65$
      & $30/30$
      & $30/30$ \\
    Process noise
      & $\sigma_z=2\sigma_{z,0}$
      & $0.65$
      & $30/30$
      & $30/30$ \\
    \bottomrule
  \end{tabular*}

  \medskip
  \textit{(b) Dynamics at $\beta=20$}

  \smallskip
  \begin{tabular*}{\linewidth}{@{\extracolsep{\fill}}llccc@{}}
    \toprule
    Condition
      & Setting
      & $\operatorname{corr}(z_1,z_2)$
      & HO/decay
      & HO/pairwise \\
    \midrule
    Baseline
      & \shortstack[l]{$r_A=0.35,\ \tau_u=\tau_{u,0},$\\
                       $g=g_0,\ \sigma_z=\sigma_{z,0}$}
      & \shortstack{$0.366$\\$[0.277,\,0.473]$}
      & \shortstack{$1.287$\\$[1.215,\,1.378]$}
      & \shortstack{$4.810$\\$[3.683,\,7.099]$} \\
    Pairwise strength
      & $r_A=0.20$
      & \shortstack{$0.219$\\$[0.117,\,0.320]$}
      & \shortstack{$1.391$\\$[1.279,\,1.475]$}
      & \shortstack{$7.516$\\$[5.706,\,10.275]$} \\
    Pairwise strength
      & $r_A=0.50$
      & \shortstack{$0.517$\\$[0.439,\,0.620]$}
      & \shortstack{$1.199$\\$[1.141,\,1.303]$}
      & \shortstack{$4.352$\\$[3.167,\,6.270]$} \\
    Input correlation time
      & $\tau_u=0.5\tau_{u,0}$
      & \shortstack{$0.338$\\$[0.258,\,0.446]$}
      & \shortstack{$1.303$\\$[1.214,\,1.413]$}
      & \shortstack{$3.548$\\$[2.759,\,5.070]$} \\
    Input correlation time
      & $\tau_u=2\tau_{u,0}$
      & \shortstack{$0.392$\\$[0.305,\,0.511]$}
      & \shortstack{$1.226$\\$[1.170,\,1.284]$}
      & \shortstack{$6.413$\\$[4.663,\,9.883]$} \\
    Input gain
      & $g=0.5g_0$
      & \shortstack{$0.367$\\$[0.276,\,0.473]$}
      & \shortstack{$1.089$\\$[0.977,\,1.221]$}
      & \shortstack{$2.405$\\$[1.846,\,3.576]$} \\
    Input gain
      & $g=1.5g_0$
      & \shortstack{$0.367$\\$[0.277,\,0.472]$}
      & \shortstack{$1.311$\\$[1.244,\,1.386]$}
      & \shortstack{$7.218$\\$[5.521,\,10.624]$} \\
    Process noise
      & $\sigma_z=0$
      & \shortstack{$0.367$\\$[0.277,\,0.472]$}
      & \shortstack{$1.303$\\$[1.215,\,1.367]$}
      & \shortstack{$4.814$\\$[3.671,\,7.046]$} \\
    Process noise
      & $\sigma_z=0.5\sigma_{z,0}$
      & \shortstack{$0.367$\\$[0.277,\,0.472]$}
      & \shortstack{$1.300$\\$[1.222,\,1.376]$}
      & \shortstack{$4.812$\\$[3.680,\,7.072]$} \\
    Process noise
      & $\sigma_z=2\sigma_{z,0}$
      & \shortstack{$0.367$\\$[0.276,\,0.473]$}
      & \shortstack{$1.253$\\$[1.166,\,1.345]$}
      & \shortstack{$4.812$\\$[3.693,\,7.154]$} \\
    \bottomrule
  \end{tabular*}
\end{table}

\FloatBarrier

% ============================================================
\section{Supplementary figures and tables}
\label{app:supplementary-results}

\begin{figure}[!htbp]
    \centering
    \includegraphics[width=0.96\linewidth]
        {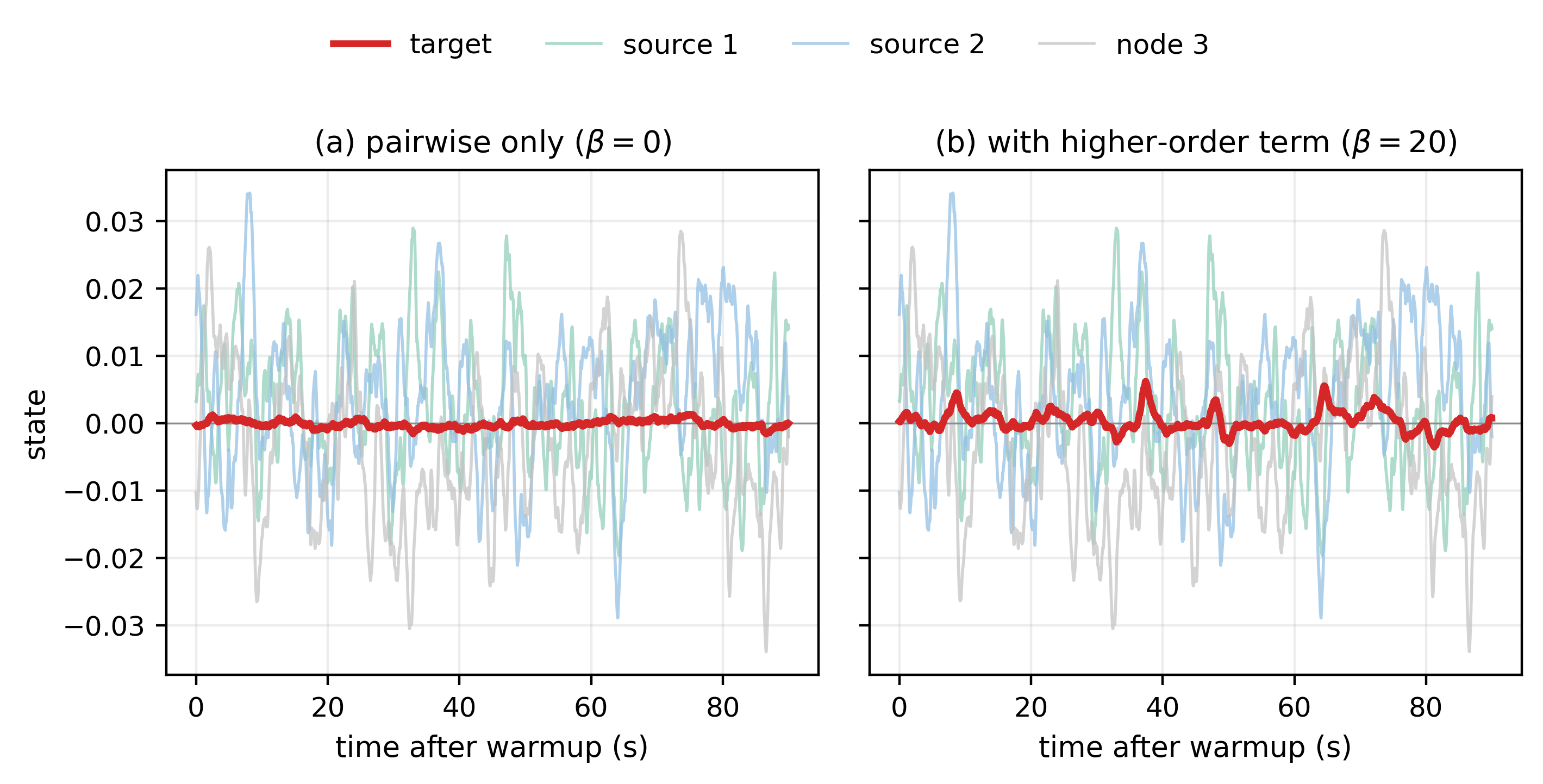}
    \caption{
Post-warm-up trajectories at $\beta=0$ (a) and $\beta=20$ (b) for one
replicate. Initial states, OU inputs, and Wiener paths are identical across
conditions. The higher-order term enters only the target equation directly,
producing larger target excursions; indirect changes in the other nodes are
negligible at the plotted scale.
}
    \label{fig:base-node-timeseries}
    \par\smallskip
{\small\raggedright
\textbf{Alt text:} Two time-series panels compare the same replicate
without and with higher-order coupling. The target shows larger
excursions with coupling.\par}
\end{figure}

\begin{figure}[!htbp]
  \centering
  \includegraphics[width=0.80\linewidth]
    {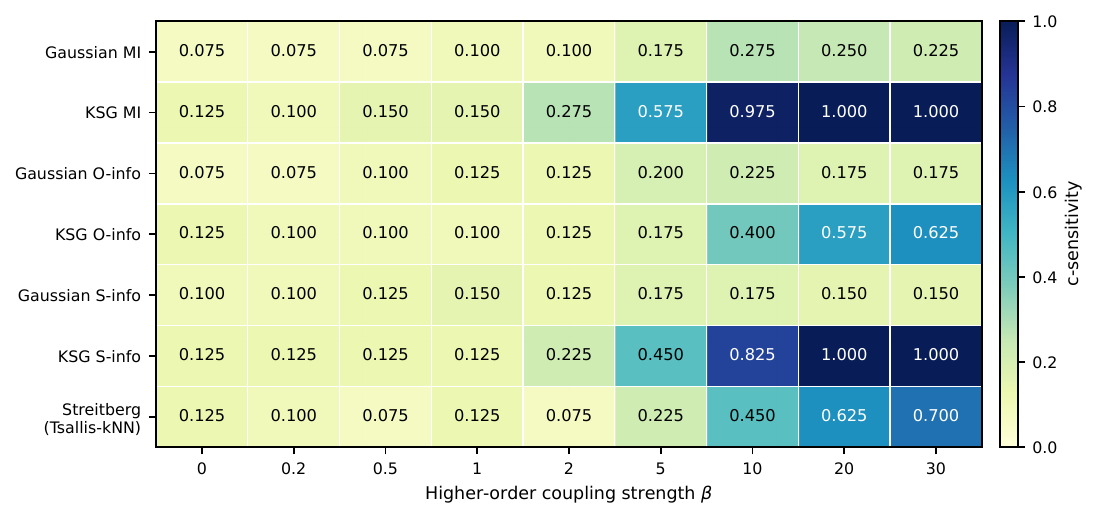}
  \caption{C-sensitivity for each detection metric and higher-order coupling
  strength in the four-node base case. The heatmap presents the same values as
  Figure~\ref{fig:base-detectability} while making the
  metric-by-$\beta$ comparisons explicit.}
  \label{fig:app-csensitivity-heatmap}
  \par\smallskip
{\small\raggedright
\textbf{Alt text:} Detection heatmap showing the strongest increases
for KSG mutual information and S-information as coupling grows.\par}
\end{figure}

\begin{table}[H]
  \centering
  \caption{KSG O-information of the planted triplet across the coupling sweep
  in the four-node base case. Values are taken over the 40 test
  replicates at each $\beta$. For replicate $r$, the paired difference is $\Delta_r(\beta)=\widehat{\Omega}_r(\beta)-\widehat{\Omega}_r(0)$, where $\widehat{\Omega}_r$ denotes the KSG estimate of O-information. Each pair shares the same initial state, input paths and Wiener paths. The \emph{More negative} column reports the number of replicates with $\Delta_r(\beta)<0$ out of the 40 test replicates. The final
  column reproduces the null-calibrated c-sensitivity of
  Figure~\ref{fig:base-detectability}. Every value is negative at every
  coupling strength, as are all 200 simulations of the independent $\beta=0$
  calibration ensemble.}
  \label{tab:app-oinformation}
  \small
  \begin{tabular}{rrrrrr}
    \toprule
    $\beta$
      & Mean
      & Range
      & Median $\Delta$
      & More negative
      & c-sensitivity \\
    \midrule
    $0$
      & $-0.631$
      & $[-0.761,-0.505]$
      & $+0.000$
      & $0/40$
      & $0.125$ \\
    $0.2$
      & $-0.631$
      & $[-0.763,-0.509]$
      & $-0.000$
      & $20/40$
      & $0.100$ \\
    $0.5$
      & $-0.632$
      & $[-0.778,-0.504]$
      & $+0.002$
      & $18/40$
      & $0.100$ \\
    $1$
      & $-0.635$
      & $[-0.769,-0.514]$
      & $-0.004$
      & $23/40$
      & $0.100$ \\
    $2$
      & $-0.639$
      & $[-0.791,-0.514]$
      & $-0.011$
      & $25/40$
      & $0.125$ \\
    $5$
      & $-0.662$
      & $[-0.789,-0.516]$
      & $-0.035$
      & $30/40$
      & $0.175$ \\
    $10$
      & $-0.709$
      & $[-0.847,-0.560]$
      & $-0.089$
      & $37/40$
      & $0.400$ \\
    $20$
      & $-0.753$
      & $[-0.885,-0.576]$
      & $-0.118$
      & $38/40$
      & $0.575$ \\
    $30$
      & $-0.761$
      & $[-0.931,-0.555]$
      & $-0.136$
      & $38/40$
      & $0.625$ \\
    \bottomrule
  \end{tabular}
\end{table}

\begin{table}[H]
  \centering
  \caption{Search-space dimensions and numerical diagnostics
  for the topology-generalisation regimes at $\beta=20$. $K$ is the number of directed interactions, $q$ is the total number of nodes in each interaction including its target, $M_{\mathrm U}$ is the number of unordered candidate node sets, and $M_{\mathrm D}$ is the number of directed source-set--target candidates. In each reciprocal configuration, the three directed interactions correspond to one unordered node set. Detection and recovery validity are
  reported as valid/requested trajectories. The final column gives the mean, across valid detection trajectories,
of the higher-order-to-process-noise increment RMS ratio, with both
RMS quantities evaluated over the planted target nodes. Directed
  recovery was not evaluated for the size-four and size-five interactions.
  The planted interactions and coefficient normalisations are defined in
  Table~\ref{tab:topology-regimes}.}
  \label{tab:app-topology-diagnostics}

  \begingroup
  \small
  \setlength{\tabcolsep}{3.2pt}
  \renewcommand{\arraystretch}{1.10}
  \begin{tabular*}{\textwidth}
    {@{\extracolsep{\fill}}lrrrrrrrr@{}}
    \toprule
    Regime
      & $N$
      & $K$
      & $q$
      & $M_{\mathrm U}$
      & $M_{\mathrm D}$
      & \shortstack{Detection\\valid}
      & \shortstack{Recovery\\valid}
      & \shortstack{HO/noise\\RMS} \\
    \midrule

    \multicolumn{9}{l}{\emph{Network size}} \\
    Directed base case
      & 4 & 1 & 3 & 4 & 12 & $40/40$ & $60/60$ & $0.996$ \\
    Eight-node network
      & 8 & 1 & 3 & 56 & 168 & $40/40$ & $60/60$ & $0.994$ \\
    Sixteen-node network
      & 16 & 1 & 3 & 560 & 1680 & $40/40$ & $60/60$ & $0.994$ \\

    \addlinespace
    \multicolumn{9}{l}{\emph{Interaction multiplicity}} \\
    Two interactions, shared target
      & 16 & 2 & 3 & 560 & 1680 & $40/40$ & $60/60$ & $1.45$ \\
    Two interactions, disjoint targets
      & 16 & 2 & 3 & 560 & 1680 & $40/40$ & $60/60$ & $0.969$ \\
    Four interactions, shared target
      & 16 & 4 & 3 & 560 & 1680 & $40/40$ & $60/60$ & $2.17$ \\
    Four interactions, separate targets
      & 16 & 4 & 3 & 560 & 1680 & $40/40$ & $60/60$ & $0.941$ \\

    \addlinespace
    \multicolumn{9}{l}{\emph{Overlap and interaction size}} \\
    Shared source
      & 16 & 2 & 3 & 560 & 1680 & $40/40$ & $60/60$ & $0.934$ \\
    Shared source and target
      & 16 & 2 & 3 & 560 & 1680 & $40/40$ & $60/60$ & $1.45$ \\
    Size-four interaction
      & 16 & 1 & 4 & 1820 & --- & $40/40$ & $60/60$ & $0.0104$ \\
    Size-five interaction
      & 16 & 1 & 5 & 4368 & --- & $40/40$ & $60/60$
      & $1.08{\times}10^{-4}$ \\

    \addlinespace
    \multicolumn{9}{l}{\emph{Reciprocal configurations}} \\
    Reciprocal, equal strength
      & 4 & 3 & 3 & 4 & 12 & $10/40$ & $17/60$ & $0.839$ \\
    Reciprocal, normalised
      & 4 & 3 & 3 & 4 & 12 & $32/40$ & $43/60$ & $0.365$ \\

    \bottomrule
  \end{tabular*}
  \endgroup
\end{table}

\clearpage
\subsection{Sampling and calibration checks for discrimination}
\label{app:discrimination-validation}

\begin{figure}[H]
    \centering
    \includegraphics[width=0.7\linewidth]
        {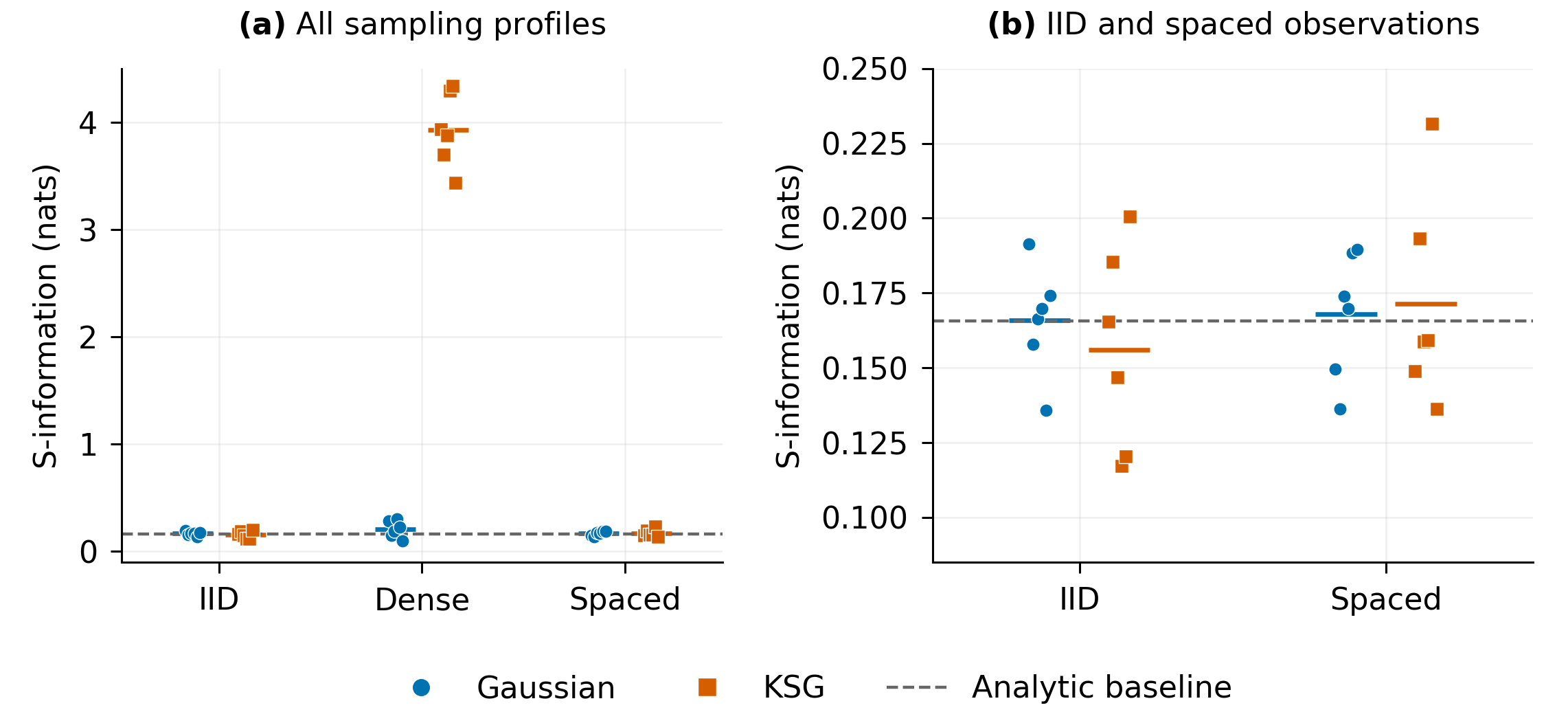}
    \caption{Baseline S-information estimates across six replicates per condition. \textbf{(a)} Independent multivariate Gaussian observations (4501 observations), dense trajectories ($T=100$, 4501 observations), and widely spaced trajectories ($T=5000$, 999 observations, spacing 5 time units). \textbf{(b)} Expanded view of the independent and widely spaced conditions. Points show individual estimates and short horizontal lines their means. The dashed line marks the analytic stationary S-information of the discretised $\beta=0$ model, $0.1657$ nats (natural units of information). Agreement under wider spacing supports the discrimination protocol, without establishing unbiased estimation under nonzero coupling.}
    \label{fig:discrimination-sampling}
    \par\smallskip
{\small\raggedright
\textbf{Alt text:} Dense sampling inflates KSG S-information;
independent and widely spaced observations give estimates
near the analytic baseline.\par}
\end{figure}

\begin{figure}[H]
    \centering
    \includegraphics[width=0.7\linewidth]
        {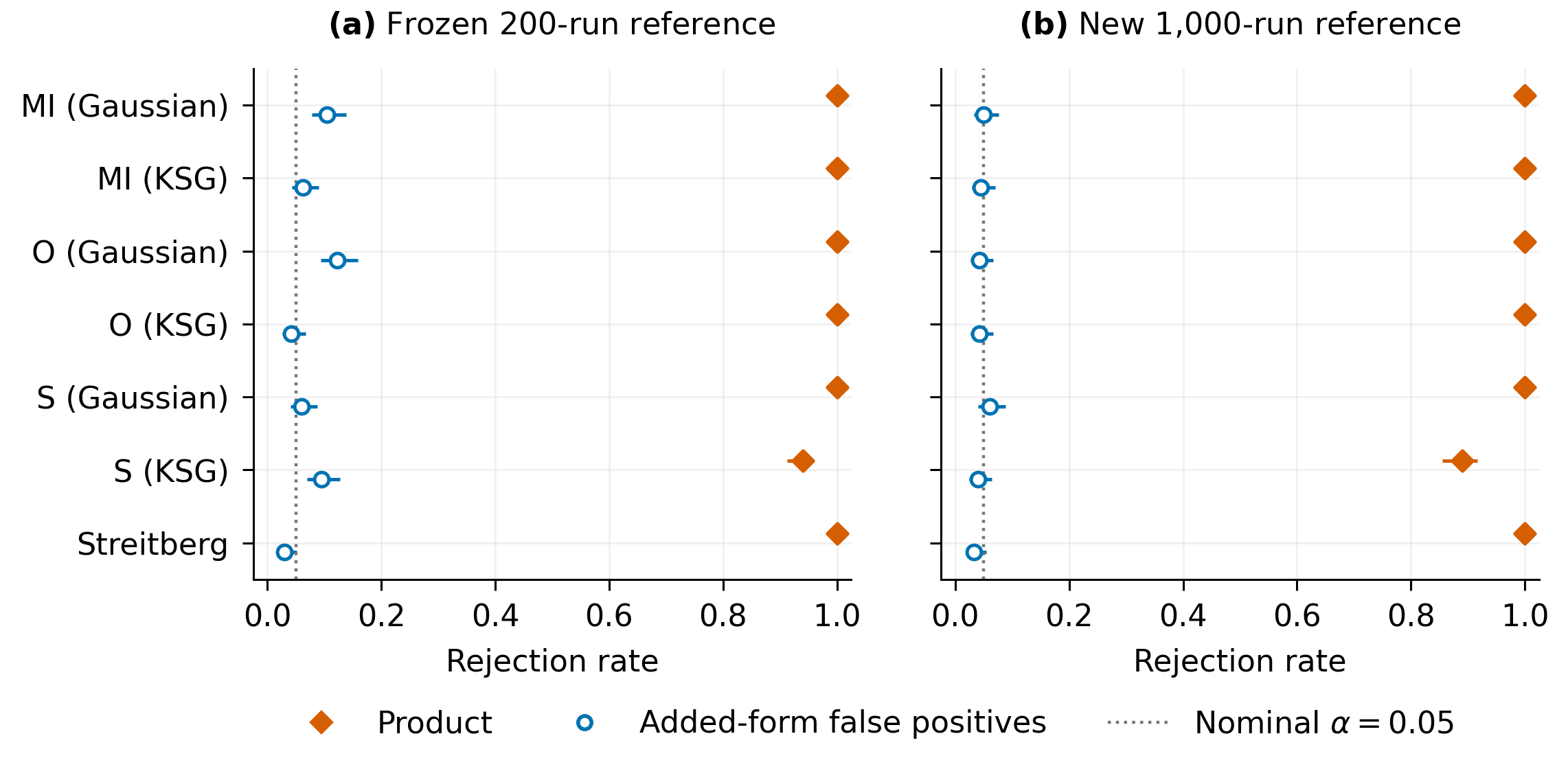}
    \caption{
    Calibration comparison at $\beta=20$ using
    \textbf{(a)} the original 200-run additive reference and
    \textbf{(b)} an independent 1000-run additive reference.
    Both are evaluated on the same 400 held-out simulations per
    coupling form. Orange diamonds show the fraction of product
    simulations outside the reference interval; blue circles
    show false-positive rates on additive controls.
    Bars show Wilson 95\% intervals conditional on the fitted
    thresholds; dotted lines mark the nominal $5\%$ level.
    The comparison concerns these two reference ensembles,
    rather than a systematic assessment of calibration size.
    }
    \label{fig:discrimination-calibration}
    \par\smallskip
{\small\raggedright
\textbf{Alt text:} Both reference ensembles yield high discrimination;
false-positive rates are closer to five percent with the
independent reference.\par}
\end{figure}

Figures~\ref{fig:discrimination-sampling}
and~\ref{fig:discrimination-calibration} provide the sampling
and calibration checks supporting Section~\ref{sec:discrimination}.
The sampling comparison changes both recording duration and
observation spacing; the calibration comparison evaluates two
reference ensembles on the same held-out simulations.

\FloatBarrier
\clearpage

\begin{figure}[!htbp]
    \centering
    \includegraphics[width=0.94\linewidth]
    {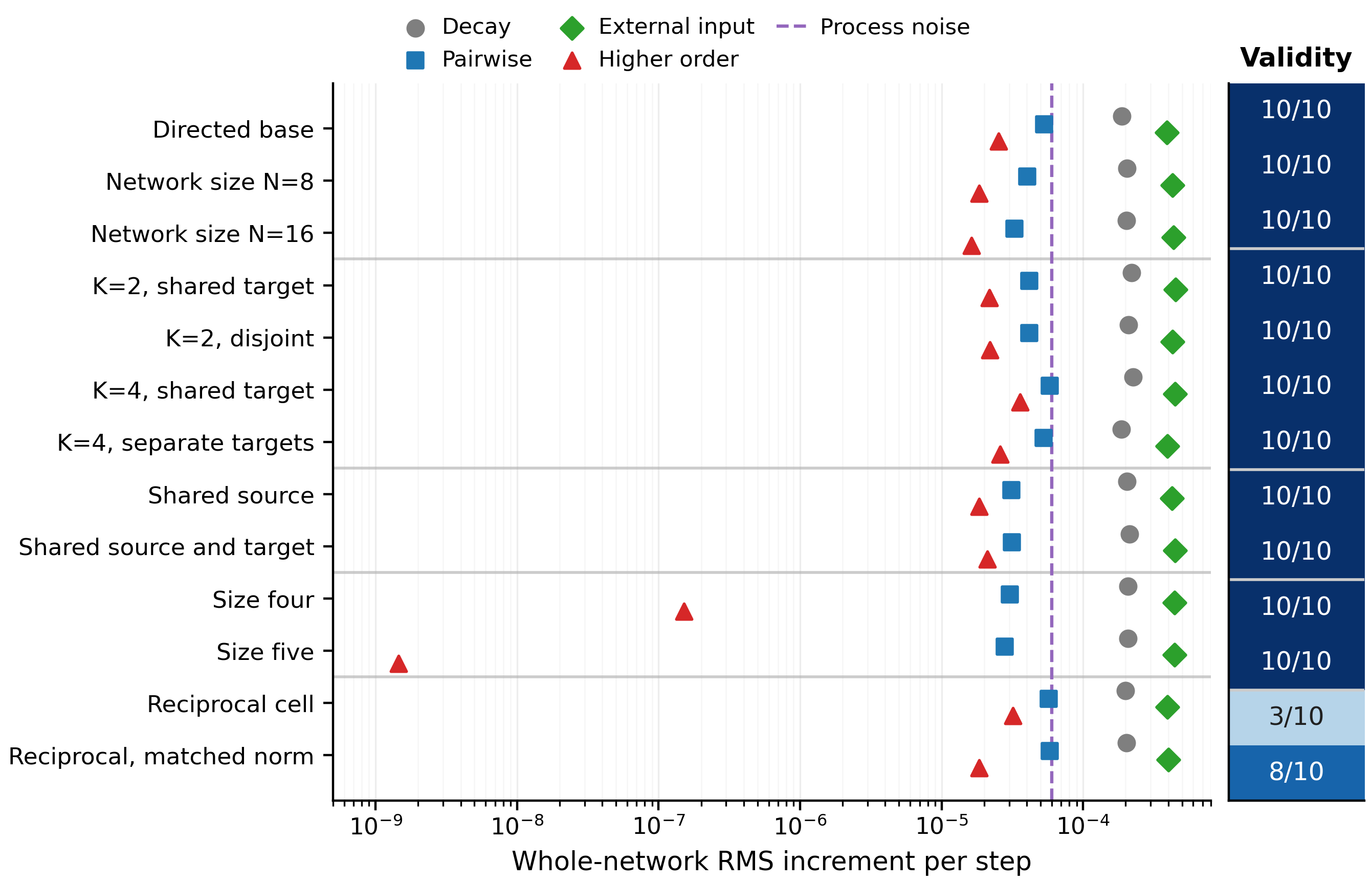}
    \caption{
Whole-network component RMS increments and numerical validity at $\beta=20$,
evaluated using 10 diagnostic simulations per layout, separate from the
detection and recovery ensembles. Markers show means across numerically valid
diagnostic trajectories, the dashed line marks the process-noise RMS, and the
right panel reports valid/requested diagnostic trajectory counts.
}
  \label{fig:topology-component-norms}
  \par\smallskip
{\small\raggedright
\textbf{Alt text:} Higher-order increments shrink sharply for larger
interactions; only reciprocal layouts have numerically invalid
diagnostic trajectories.\par}
\end{figure}

\begin{figure*}[p]
  \centering
  \includegraphics[
    width=\textwidth,
    height=0.58\textheight,
    keepaspectratio
  ]{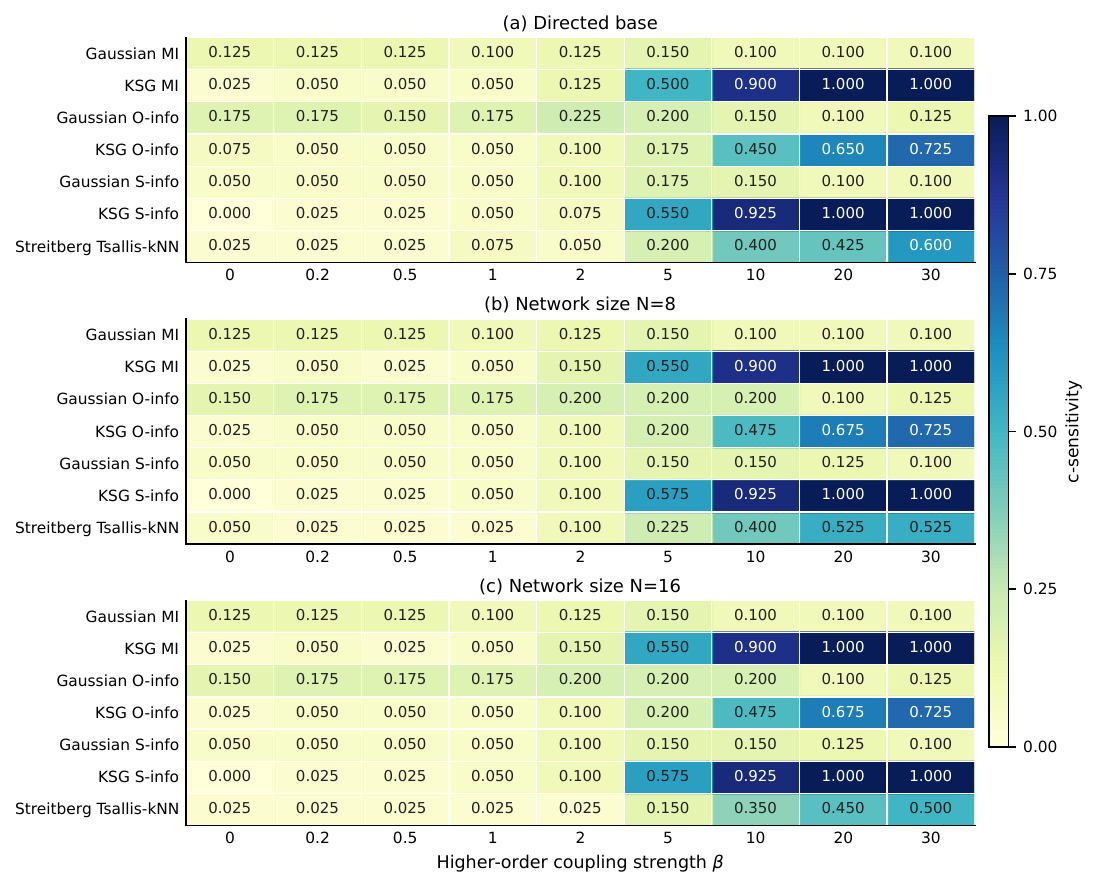}
  \caption{Null-calibrated c-sensitivity across the
  network-size regimes. Panels show the four-node directed base case and the
  corresponding eight- and sixteen-node configurations. Rows denote the seven
  detection metrics, columns denote the higher-order coupling strength
  $\beta$, and each cell reports the fraction of valid test trajectories that
  exceeded the corresponding topology-specific null threshold. All $40$
  requested trajectories were numerically valid at every coupling strength
  in these regimes.}
  \label{fig:topology-detectability-heatmaps-network}
  \par\smallskip
{\small\raggedright
\textbf{Alt text:} Three heatmaps show similar detection patterns
across network sizes, with KSG mutual information and S-information
reaching full detection at strong coupling.\par}
\end{figure*}

\begin{figure*}[p]
  \centering
  \includegraphics[
    width=\textwidth,
    height=0.88\textheight,
    keepaspectratio
  ]{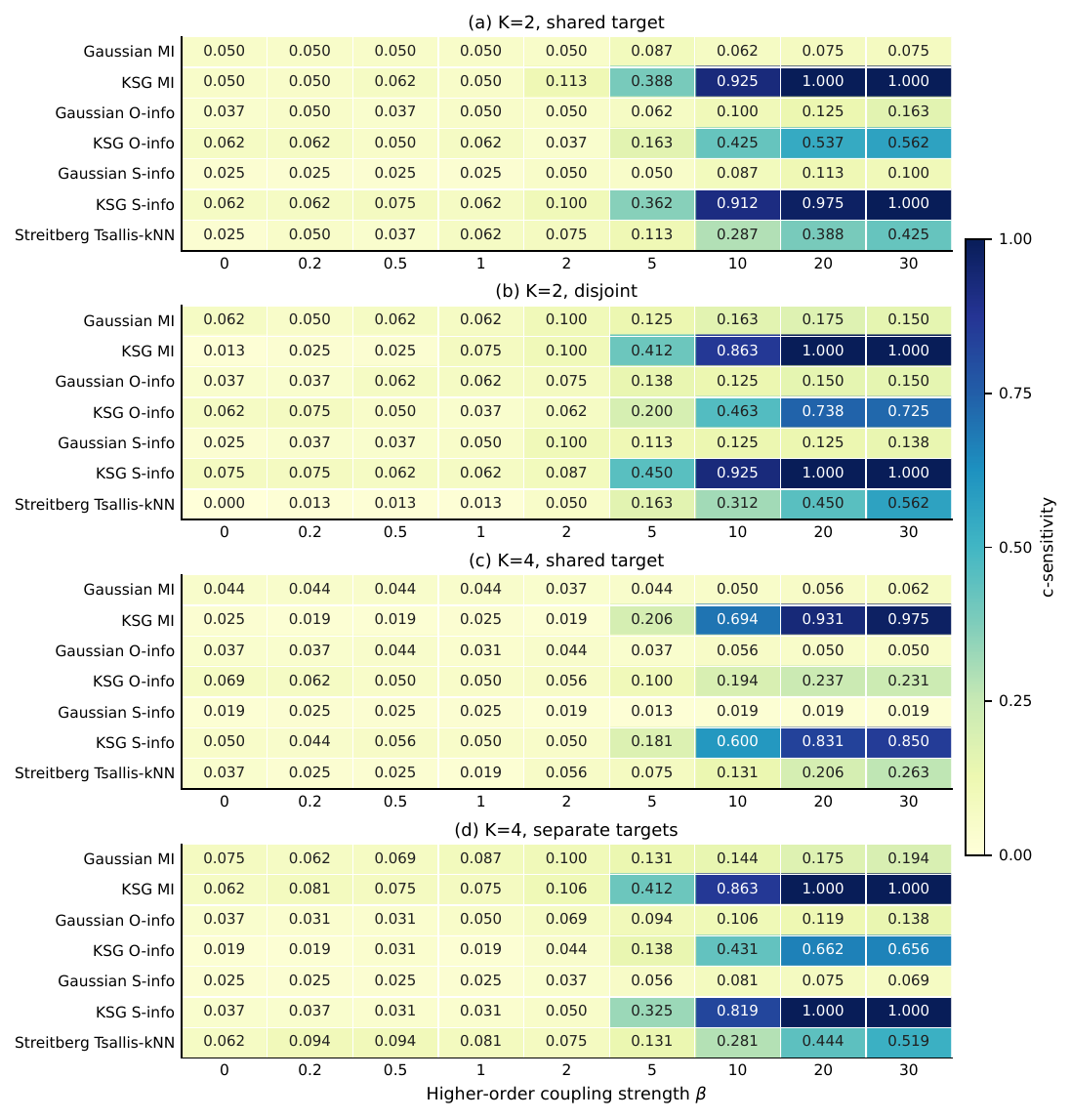}
  \caption{Null-calibrated c-sensitivity across the
  interaction-multiplicity regimes. The panels compare two and four planted
  interactions under shared-target and separate-target configurations. Rows
  denote detection metrics and columns denote $\beta$. For each metric and
  coupling strength, c-sensitivity was calculated separately for each planted
  interaction and then averaged across the $K$ interactions. All $40$
  requested trajectories were numerically valid in each condition.}
  \label{fig:topology-detectability-heatmaps-multiplicity}
  \par\smallskip
{\small\raggedright
\textbf{Alt text:} Four heatmaps compare detection of multiple
interactions, showing weaker detection when four interactions
share a target.\par}
\end{figure*}

\begin{figure*}[p]
  \centering
  \includegraphics[
    width=\textwidth,
    height=0.88\textheight,
    keepaspectratio
  ]{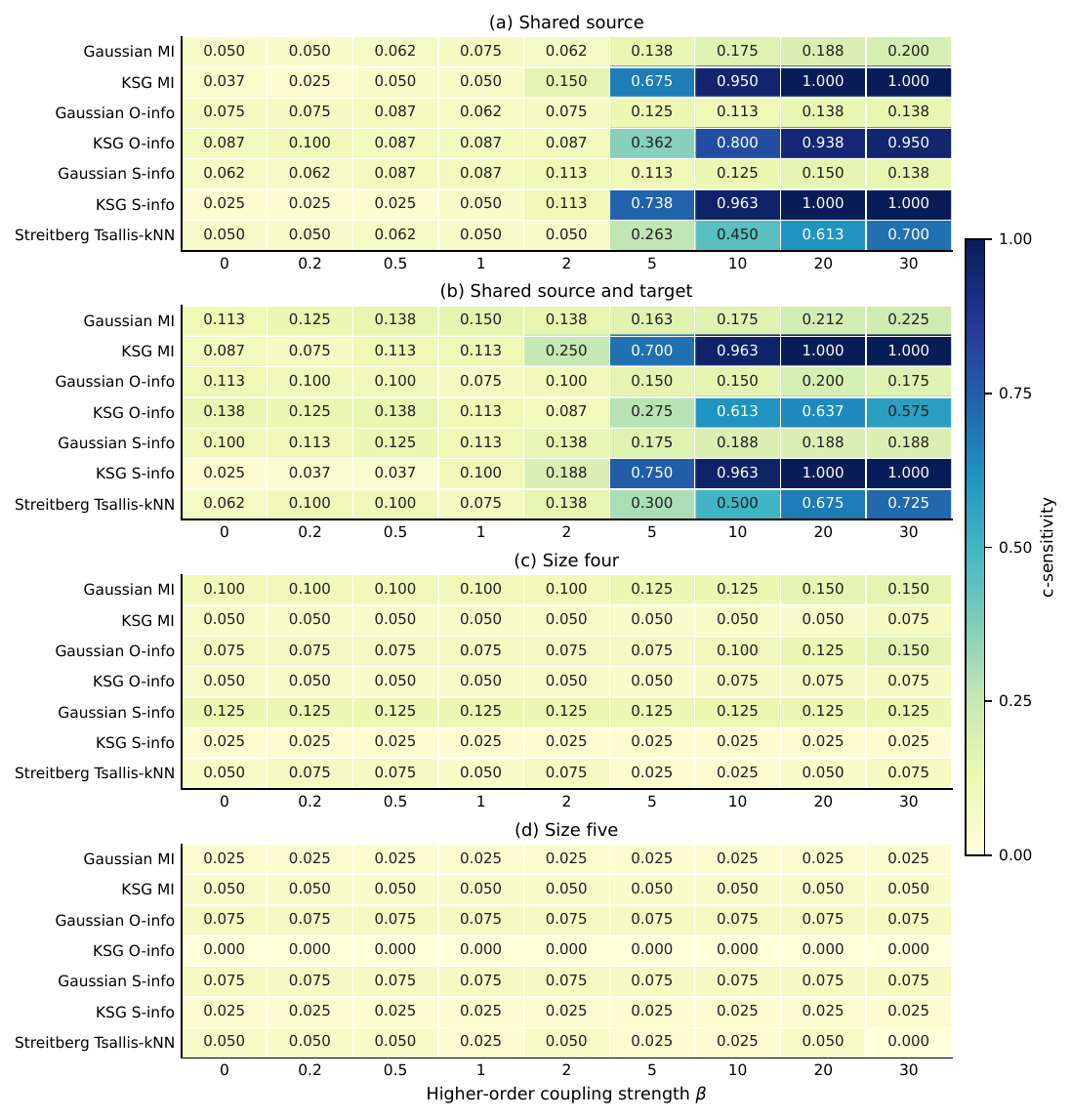}
  \caption{Null-calibrated c-sensitivity across the
  interaction-overlap and interaction-size regimes. Panels (a) and (b)
  compare interactions sharing a source with interactions sharing both a
  source and their target. Panels (c) and (d) show single raw-product
  interactions containing four and five nodes, respectively, including the
  target. Rows denote detection metrics and columns denote $\beta$. Values in
  the multi-interaction conditions are means across the separately retained
  interaction-level results. All $40$ requested trajectories were numerically
  valid in each condition.}
  \label{fig:topology-detectability-heatmaps-overlap-size}
  \par\smallskip
{\small\raggedright
\textbf{Alt text:} Heatmaps show detection increasing with coupling
for overlapping interactions, but remaining low for four-
and five-node unscaled-product interactions.\par}
\end{figure*}

\begin{figure*}[p]
  \centering
  \includegraphics[
    width=\textwidth,
    height=0.78\textheight,
    keepaspectratio
  ]{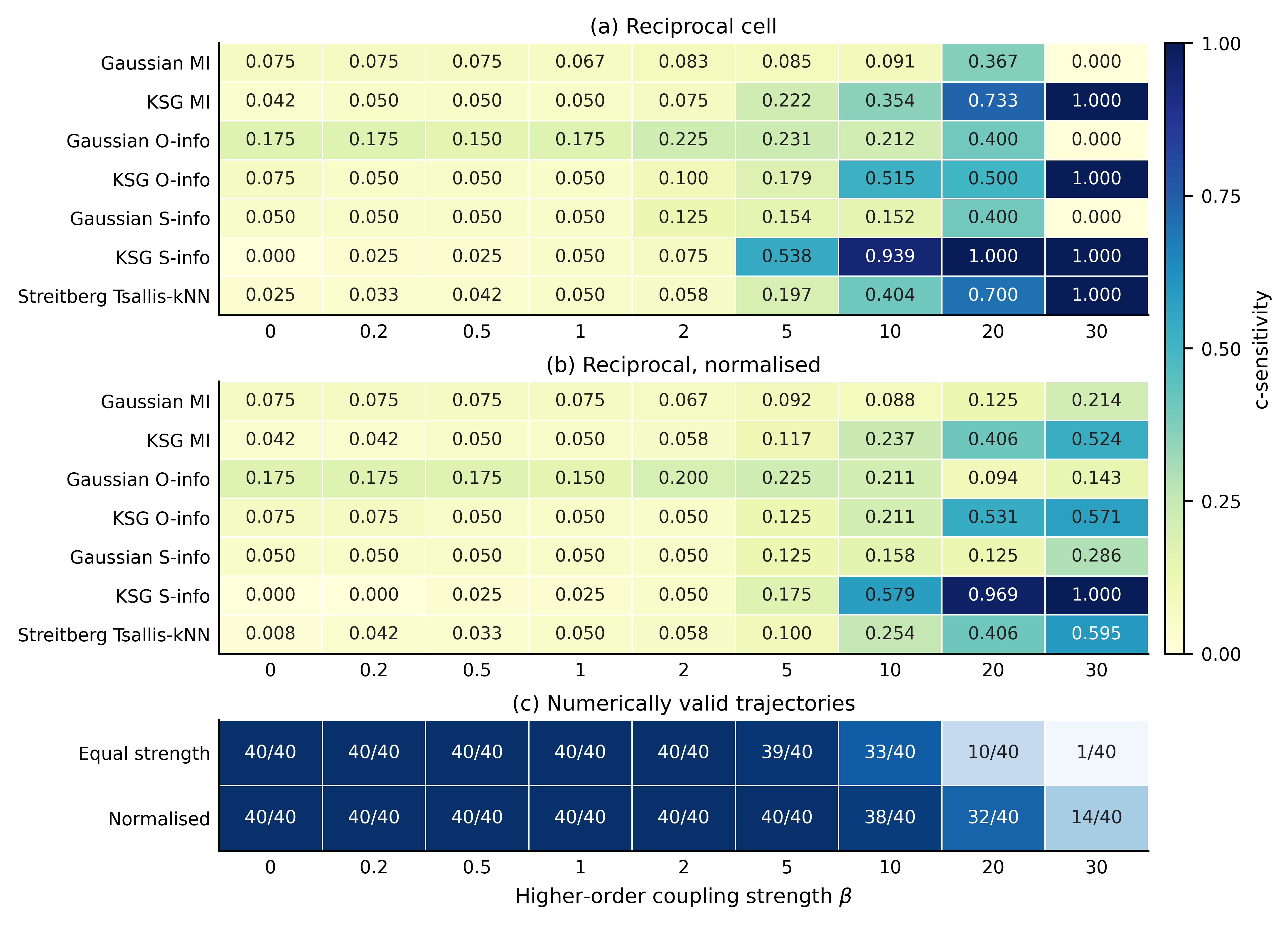}
  \caption{Null-calibrated c-sensitivity and numerical
  validity in the reciprocal configurations. Panels (a) and (b) show mean interaction-level c-sensitivity under
equal local interaction strength and a total coefficient norm of
$\beta$, respectively. Rows denote detection metrics
  and columns denote $\beta$. Panel (c) reports the number of numerically valid
  trajectories out of the $40$ requested trajectories. Because c-sensitivity
  is calculated conditional on numerical validity, values at the largest
  coupling strengths are based on progressively smaller retained subsets,
  particularly in the equal-strength condition.}
  \label{fig:topology-detectability-heatmaps-reciprocity}
  \par\smallskip
{\small\raggedright
\textbf{Alt text:} Heatmaps show stronger detection among valid
simulations as coupling increases, but fewer simulations remain
numerically valid, especially in the equal-strength configuration.\par}
\end{figure*}

\FloatBarrier

% Place all remaining appendix floats before the bibliography.
\clearpage

% \input{comnet/to_do_list}
%\input{acknowledgements.tex}
%\input{funding.tex}
%\input{data_availability.tex}


\bibliographystyle{comnet}
\bibliography{references}

% To get the unnumbered reference style the author should use [unnumbib]
% as an option in the document class. For example: \documentclass[unnumbib]{comnet}

\end{document}